\documentclass[letterpaper]{article} 
\usepackage{aaai2026}  
\usepackage{times}  
\usepackage{helvet}  
\usepackage{courier}  
\usepackage[hyphens]{url}  
\usepackage{graphicx} 
\usepackage{natbib}  
\usepackage{algorithm}
\usepackage{algorithmic}

\usepackage{booktabs} 
\usepackage{amsmath}     
\usepackage{amssymb}      
\usepackage{colortbl}
\usepackage{multirow}
\nocopyright 

\title{Shrinking the Teacher: An Adaptive Teaching Paradigm for Asymmetric EEG-Vision Alignment}
\author{
    Lukun Wu\textsuperscript{\rm 1},
    Jie Li\textsuperscript{\rm 1}\footnotemark[1],
    Ziqi Ren\textsuperscript{\rm 2},
    Kaifan Zhang\textsuperscript{\rm 1},
    Xinbo Gao\textsuperscript{\rm 1}\thanks{Corresponding author.}
}
\affiliations{
    \textsuperscript{\rm 1}School of Electronic Engineering, Xidian University, Xi'an, China\\
    \textsuperscript{\rm 2}School of Life Science and Technology, Xidian University, Xi'an, China\\
    lkwu@stu.xidian.edu.cn, \{leejie, xbgao\}@mail.xidian.edu.cn
}

\usepackage{bibentry}

\begin{document}

\maketitle

\begin{abstract}
Decoding visual features from EEG signals is a central challenge in neuroscience, with cross-modal alignment as the dominant approach. We argue that the relationship between visual and brain modalities is fundamentally asymmetric, characterized by two critical gaps: a Fidelity Gap (stemming from EEG's inherent noise and signal degradation, vs. vision's high-fidelity features) and a Semantic Gap (arising from EEG's shallow conceptual representation, vs. vision's rich semantic depth). Previous methods often overlook this asymmetry, forcing alignment between the two modalities as if they were equal partners and thereby leading to poor generalization. To address this, we propose the adaptive teaching paradigm. This paradigm empowers the ``teacher" modality (vision) to dynamically shrink and adjust its knowledge structure under task guidance, tailoring its semantically dense features to match the ``student" modality (EEG)'s capacity. We implement this paradigm with the ShrinkAdapter, a simple yet effective module featuring a residual-free design and a bottleneck structure. Through extensive experiments, we validate the underlying rationale and effectiveness of our paradigm. Our method achieves a top-1 accuracy of 60.2\% on the zero-shot brain-to-image retrieval task, surpassing previous state-of-the-art methods by a margin of 9.8\%. Our work introduces a new perspective for asymmetric alignment: the teacher must shrink and adapt to bridge the vision-brain gap.
\end{abstract}

\begin{links}
    \link{Code}{https://github.com/LukunWuXDU/ATS}
\end{links}

\section{Introduction}
\label{sec:introduction}
Visual neural decoding aims to interpret visual content from brain activity, serving as a bridge between human cognition and artificial intelligence while deepening our understanding of the human visual mechanism. Among various neuroimaging techniques, electroencephalography (EEG) has attracted significant attention due to its non-invasive nature, high temporal resolution, and portability, endowing it with greater potential for brain-computer interface (BCI) applications.
\begin{figure}[t]
  \centering
    \includegraphics[width=1.0\linewidth]{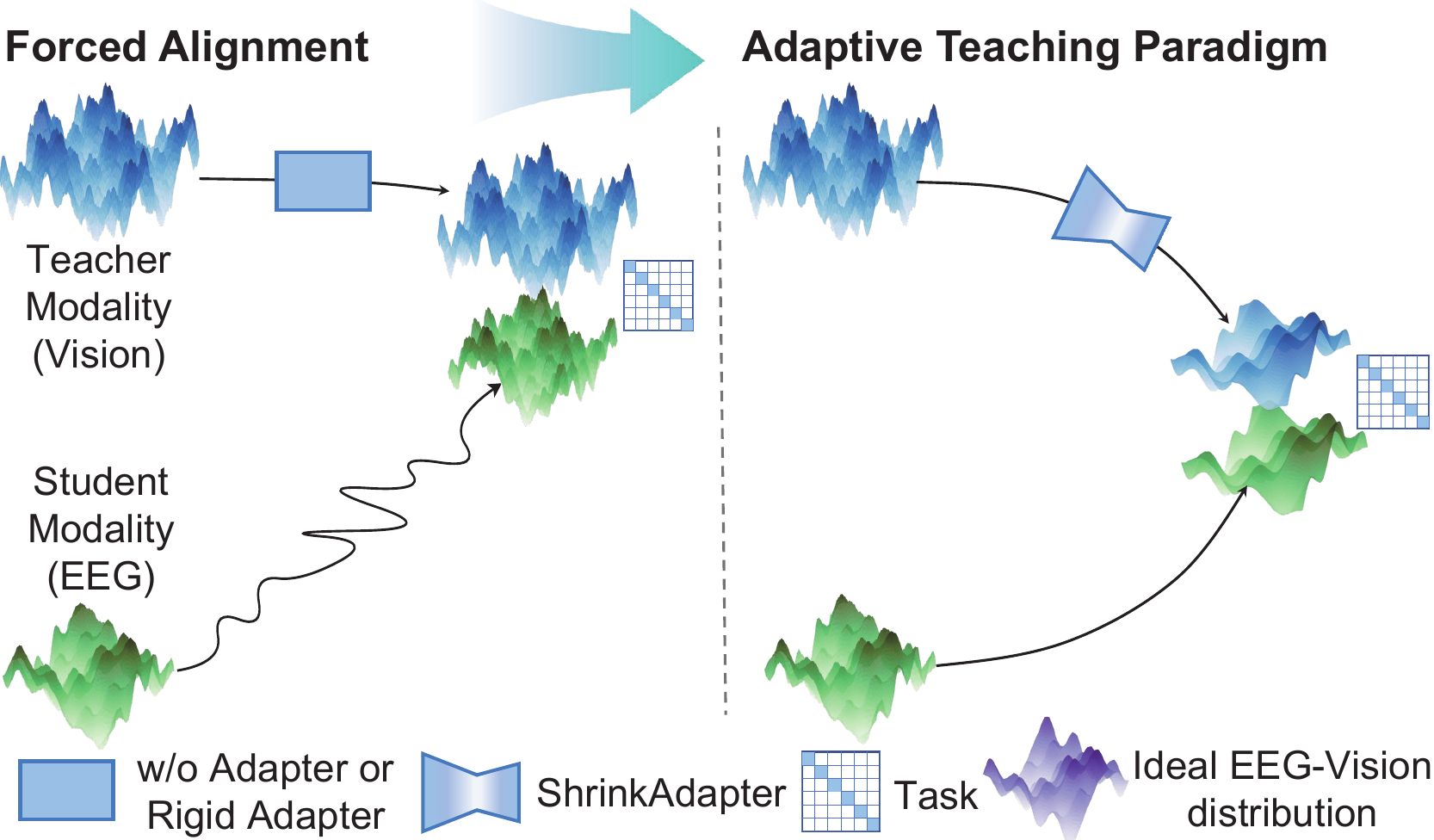}
    \caption{From Forced Alignment to Adaptive Teaching: A paradigm shift for asymmetric modality alignment.}
    \label{fig:paradigm_comparison}
\end{figure}

The dominant approach currently decodes visual content by aligning EEG signals with pretrained visual features. While most previous methods acknowledge the differences between the EEG and visual modalities, they still treat the alignment task as a symmetric problem, implicitly assuming that the two modalities have comparable fidelity and capacity. In contrast, we argue that such modality differences are inherently \textbf{asymmetric}.

\begin{figure*}[t]
  \centering
    \includegraphics[width=0.94\linewidth]{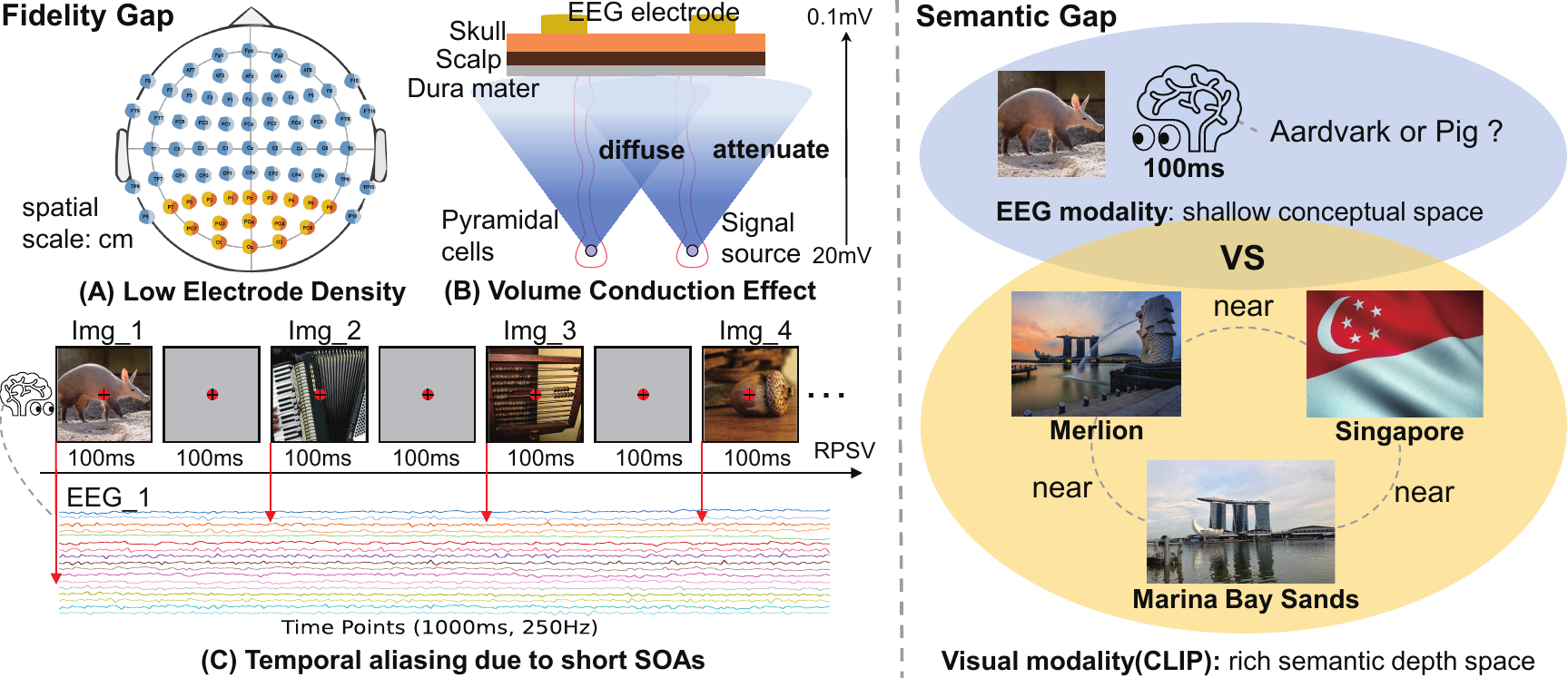}
    \caption{The physiological basis for our motivation: Deconstructing the asymmetric modality gap between vision and EEG into Fidelity Gap and Semantic Gap.}
    \label{fig:motivation}
\end{figure*}

We deconstruct this asymmetry into two core components. The first is a \textbf{Fidelity Gap}, arising from the physical limitations of EEG acquisition. As illustrated in Figure~\ref{fig:motivation}, spatially, the sparse distribution of electrodes and the inherent \textbf{volume conduction effect}~\cite{michel2012towards} lead to significant spatial blurring, as neural signals attenuate and diffuse while propagating through the head. This issue is compounded temporally by the \textbf{temporal aliasing} in Rapid Serial Visual Presentation (RSVP) paradigms~\cite{grootswagers2019representational,keysers2001speed}, leading to significant cross-stimulus interference. These factors degrade the signal quality, resulting in a low-fidelity representation that starkly contrasts with the clean, detailed features from vision models. The second component is a more subtle \textbf{Semantic Gap}. It is questionable whether the human brain, within a fleeting $100-200$\,ms exposure, can form a neural representation as semantically rich and nuanced as that of a large vision model trained on billions of images. The resulting brain signal, therefore, occupies a much smaller and less structured semantic subspace.

Given the profound asymmetry between the ``teacher" modality (the visual features from a pretrained vision model) and the ``student" modality (the EEG signal), forcing the student to directly learn from the fixed teacher—a process we term \textbf{Forced Alignment}—is an ill-posed strategy that often leads to overfitting to noise. This motivates a conceptual shift to our \textbf{Adaptive Teaching Paradigm}, as shown in Figure~\ref{fig:paradigm_comparison}. Inspired by the pedagogical principle of teaching according to aptitude, we argue the teacher modality must dynamically shrink and adjust its knowledge structure under \textbf{task guidance}, tailoring its semantically dense features to match the student modality (EEG)'s capacity, thereby achieving a more robust and generalizable alignment.

To realize this paradigm, we introduce the \textbf{Adaptive Teaching System (ATS)}, a framework designed to better bridge the vision-brain gap, thereby improving neural decoding performance. For the teacher, we develop the \textbf{ShrinkAdapter}, a simple yet efficient module that provides the conditions for the teacher modality to unconstrainedly shrink and adjust its knowledge structure. For the student, we design the \textbf{Shared Temporal Attention Encoder (STAE)} to effectively extract salient features from temporally noisy EEG data.

Our contributions can be summarized as follows:
\begin{enumerate}
    \item To our knowledge, we are the first to deconstruct the vision-brain modality gap into an asymmetric problem, comprising a Fidelity Gap and a Semantic Gap.
    \item We introduce a framework called Adaptive Teaching System (ATS), which implements our Adaptive Teaching Paradigm through the ShrinkAdapter. Additionally, the Shared Temporal Attention Encoder (STAE) enhances feature extraction from EEG signals.
    \item Through extensive experiments, we validate the underlying rationale and effectiveness of our paradigm. Our work achieves SOTA performance and provides new insights into asymmetric alignment tasks in neuroscience.
\end{enumerate}


\section{Related Work}
\label{sec:related_work}

\subsection{Visual Neural Decoding}

Visual neural decoding aims to interpret brain activity to retrieve, recognize, or reconstruct visual stimuli. While early efforts often utilized fMRI for its high spatial resolution~\cite{ren2021reconstructing,takagi2023high, scotti2024reconstructing}, EEG has recently gained significant attention due to its high temporal resolution, low cost, and portability~\cite{spampinato2017deep,ijcai2019p192}. The advent of large-scale datasets like THINGS-EEG~\cite{gifford2022large}, coupled with pioneering frameworks like BraVL~\cite{du2023decoding} and NICE~\cite{song2024decoding}, laid the groundwork for the current research landscape. Consequently, zero-shot EEG-to-image retrieval based on self-supervised contrastive learning has been cemented as the dominant paradigm, making the pursuit of a more effective and robust vision-brain alignment the next core challenge for the field.

\begin{figure*}[ht]
  \centering
    \includegraphics[width=0.95\linewidth]{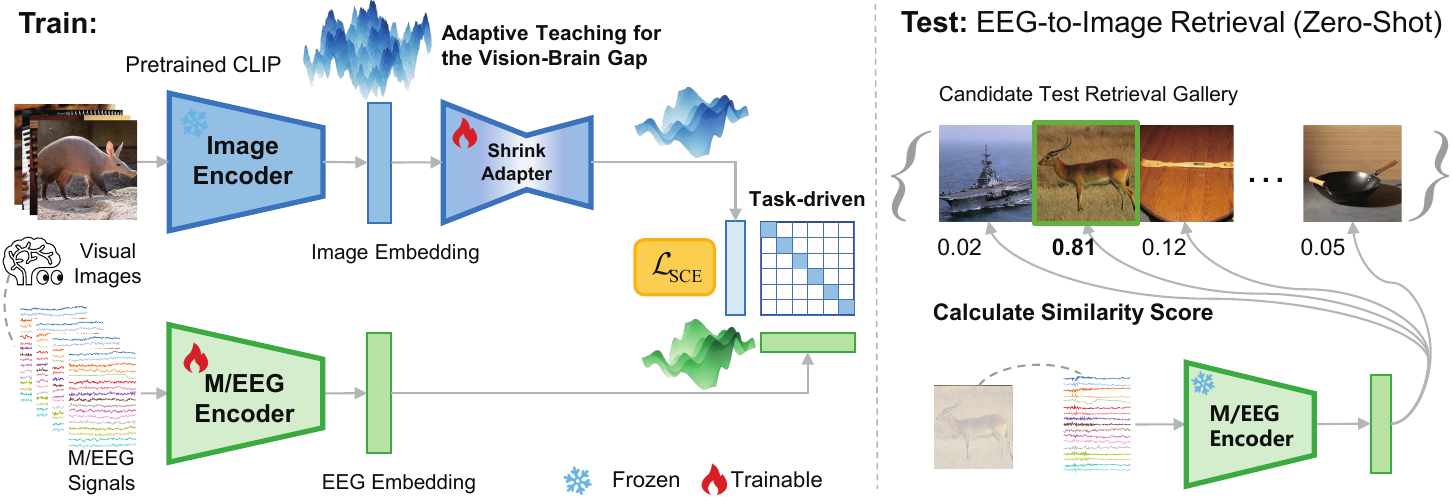}
    \caption{Overview of the Adaptive Teaching System for zero-shot EEG-to-Image retrieval. Training: A ShrinkAdapter enables the visual ``teacher" to adapt its features for alignment with the EEG ``student" via InfoNCE loss. Testing: The trained student encoder performs zero-shot image retrieval from a candidate set.}
    \label{fig:pipeline}
\end{figure*}

\subsection{Multi-Modal Alignment in EEG Retrieval}

The prevailing approach for EEG retrieval is contrastive learning, which aligns EEG features with pretrained visual embeddings in a shared latent space. Previous works have generally improved upon this paradigm from two main directions. The first is \textbf{information enhancement}. Based on the premise that human cognition is inherently multi-modal, these methods enrich the learning process with auxiliary data to better interpret the rich information within EEG signals. For instance, BraVL and NICE++~\cite{song2025recognizing} incorporate an additional text modality, while CognitionCapturer~\cite{zhang2025cognitioncapturer} even introduces depth map information. The second direction is \textbf{constraint reinforcement}. These methods acknowledge the existence of the modality differences and design more sophisticated alignment strategies to mitigate it. For example, MB2C~\cite{wei2024mb2c} introduces a cycle-consistency loss, and VE-SDN~\cite{chen2024visualneuraldecodingimproved} constructs a joint semantic space to improve alignment.

While these approaches have advanced the field significantly, they largely ignore the inherent asymmetry between the visual and EEG modalities, leading to suboptimal generalization performance. A notable exception is the UBP~\cite{wu2025bridging}, which, for the first time, approached the problem from the biological characteristics of EEG signals. It introduced a dynamically adjusted blur prior to bridge what it termed the ``System Gap" and ``Random Gap". This hand-crafted prior proved highly effective. However, being manually designed, it is neither comprehensive nor flexible enough to capture the full complexity of the modality gap. This limitation motivates us to re-examine this gap and propose the \textbf{Adaptive Teaching Paradigm} as a more fundamental and flexible solution.

\subsection{Feature Adaptation for Pre-trained Models}

A projection layer is typically used when adapting large pretrained models for downstream tasks, primarily for light feature fine-tuning.~\cite{kumar2022finetuningdistortpretrainedfeatures} Yet, when applied to the vision-EEG field, this approach is often adopted without critical examination, leaving its true potential and strategic importance critically overlooked. Consequently, prior works either omit this projection entirely (e.g., UBP), placing the full learning burden on the ``student," or treat it as a generic stabilizer (e.g., NICE, MB2C), failing to harness its full potential to address the asymmetric modality gap.


\section{Methodology}
\label{sec:methodology}

\subsection{Problem Formulation}
\label{sec:formulation}
Our ultimate goal is to decode the visual features of stimuli directly from brain signals, thereby exploring the informational capacity of EEG signals and advancing our understanding of the human visual mechanism. The zero-shot retrieval task serves as an effective paradigm for this objective. Following the standard setup~\cite{song2024decoding}, we are given a training dataset $D_{train} = \{ (x_v, x_b, y) \}$, where $x_v$ is an image, $x_b$ is the corresponding brain signal (EEG/MEG), and $y$ is the class label of $x_v$ from a set of training classes, $Y_{seen}$. The model is trained only on $D_{train}$.

As illustrated in Figure~\ref{fig:pipeline}, during the test phase, the model is evaluated on a test set $D_{test} = \{ (x_v^u, x_b^u, y^u) \}$, where the labels $y^u$ belong to a set of testing classes $Y_{unseen}$, ($Y_{seen} \cap Y_{unseen} = \emptyset$). For a given test brain signal $x_b^u$, the objective is to retrieve its corresponding image $x_v^u$ from a candidate set $X_v^u$ containing all images from $Y_{unseen}$. This is achieved by identifying the image feature with the highest similarity score to the brain feature in the learned shared latent space.

\subsection{The Adaptive Teaching System (ATS)}
\label{sec:framework}
To address the asymmetric modality gap, we introduce the Adaptive Teaching System (ATS), illustrated in Figure~\ref{fig:pipeline}. The framework comprises two main branches that map inputs into a shared latent space.

The \textbf{visual branch} (the ``teacher") first uses a powerful pretrained vision encoder $f_V$ (e.g., from CLIP)~\cite{radford2021learning} to extract a high-dimensional feature vector $h_v = f_V(x_v)$. Crucially, we then introduce a trainable ShrinkAdapter, denoted as $f_A$, which dynamically adapts this feature into a representation $z_v = f_A(h_v)$ that is more accessible for the student modality.

The \textbf{brain branch} (the ``student") employs a trainable encoder $f_B$, which learns to map a brain signal $x_b$ into an embedding $z_b = f_B(x_b)$ in the shared latent space.

The alignment between the adapted teacher and student is driven by a symmetric contrastive loss, whose objective is to pull positive pairs closer together while pushing negative pairs apart. Following the standard in-batch negative sampling setup, within a mini-batch of size $N$, a corresponding image-brain pair constitutes a positive pair, while all other non-corresponding pairings serve as negative pairs.
We employ the Symmetric Cross-Entropy (SCE) loss, an objective derived from the InfoNCE loss~\cite{wang2021combating,oord2019representationlearningcontrastivepredictive}, to jointly optimize the ShrinkAdapter~$f_A$ and the brain encoder~$f_B$:
\begin{multline}
\label{eq:align_loss}
\mathcal{L}_{\text{SCE}} = -\frac{1}{2N} \sum_{i=1}^{N} \Biggl[ \log \frac{\exp(z_{v,i}^\top z_{b,i}/\tau)}{\sum_{k=1}^{N} \exp(z_{v,i}^\top z_{b,k}/\tau)} \\
+ \log \frac{\exp(z_{b,i}^\top z_{v,i}/\tau)}{\sum_{k=1}^{N} \exp(z_{b,i}^\top z_{v,k}/\tau)} \Biggr],
\end{multline}
where $(z_{v,i}, z_{b,i})$ represents a positive pair of L2-normalized embeddings, $(\cdot)^\top(\cdot)$ denotes cosine similarity, and $\tau$ is a learnable temperature parameter. This objective not only encourages the student to align with the teacher but, critically, compels the teacher (via the trainable ShrinkAdapter) to adapt its representation $z_v$ to be more accessible to the student, thus realizing our adaptive teaching paradigm. The algorithmic flow is detailed in Appendix~A.4.

\subsection{ShrinkAdapter}
\label{sec:shrink_adapter}
\begin{figure}
    \centering
    \includegraphics[width=1.0\linewidth]{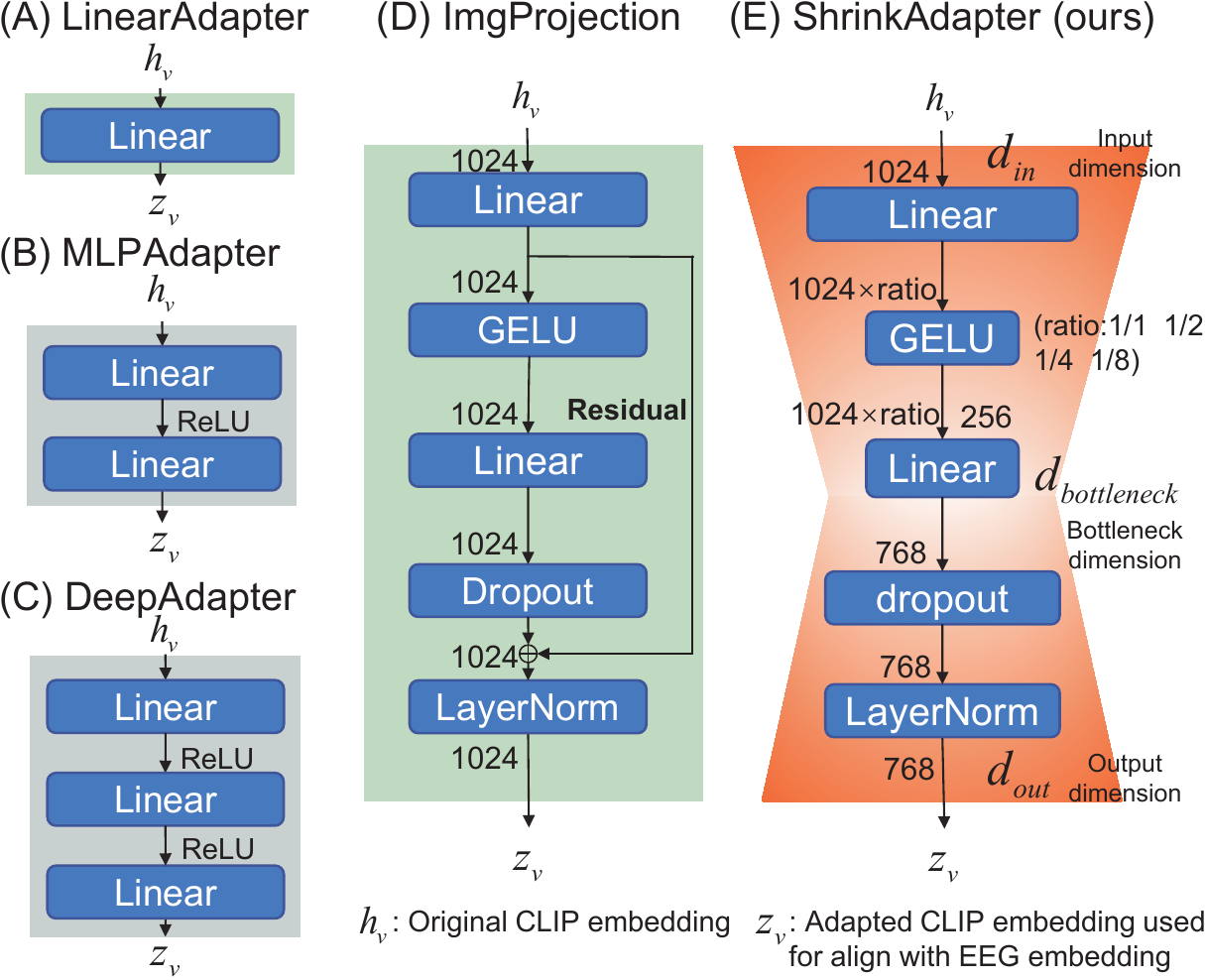}
    \caption{Structure of different adapters.}
    \label{fig:adapter_structure}
\end{figure}

Based on our analysis of the asymmetric modality gap, we argue that the pre-trained ``teacher" modality contains knowledge that is complex and redundant relative to the ``student's" capacity. Therefore, the core of adaptive teaching is to enable the teacher to adapt and shrink its knowledge to match the student's potential. We realize this through our ShrinkAdapter, a module guided by two core principles: adaptive freedom via a residual-free design, and information shrinking via a bottleneck.

First, to provide the teacher with the freedom to fully adapt, we deliberately \textbf{eliminate residual connections}. A residual connection~\cite{he2016deep}, while commonly used in prior work to preserve the original feature distribution (Figure~\ref{fig:adapter_structure}(D)), is fundamentally in conflict with our design philosophy. By removing this constraint, the adapter is free to learn an entirely new representation, driven by the task.

Second, to shrink the teacher's redundant information, we employ a \textbf{bottleneck architecture}. As illustrated in Figure~\ref{fig:adapter_structure}(E), this structure forces the visual features through a low-dimensional bottleneck, thereby filtering out irrelevant information. The architecture is defined as:
\begin{equation}
z_v = f_A(h_v) = W_{up} \text{GELU}(W_{down} h_v),
\label{eq:adapter_arch}
\end{equation}
where $W_{down}$ and $W_{up}$ are linear layers, and we define the bottleneck compression ratio as $d_{\text{bottleneck}} / d_{\text{in}}$, with the bottleneck dimension satisfying $d_{\text{bottleneck}} \ll d_{\text{in}}$.

The design of the ShrinkAdapter is theoretically motivated by the \textbf{Information Bottleneck (IB) principle}~\cite{tishby2000informationbottleneckmethod}. The IB principle seeks to learn a maximally compressed representation $z_v$ of an input $h_v$ that preserves the most relevant information about a target $z_b$. This trade-off is formalized by minimizing the Lagrangian:
\begin{equation}
\mathcal{L}_{\text{IB}} = I(h_v; z_v) - \beta I(z_v; z_b),
\label{eq:ib_loss}
\end{equation}
where $I(\cdot;\cdot)$ is the mutual information and $\beta$ is a Lagrange multiplier. The ShrinkAdapter implements this principle through its core mechanisms. The \textbf{bottleneck architecture} is the primary mechanism for minimizing the compression term, $I(h_v; z_v)$, and this is enabled by the \textbf{residual-free design}, which removes a major obstacle to this goal. Meanwhile, our \textbf{alignment loss} ($\mathcal{L}_{\text{SCE}}$) serves as a proxy to maximize $I(z_v; z_b)$, the term representing the task-relevant information for alignment. Through this synergistic design, our ATS effectively realizes the IB principle.


\subsection{Shared Temporal Attention Encoder (STAE)}
\label{sec:stae_encoder}

To enhance the ``student's" ability to learn from the adapted ``teacher," we propose the \textbf{Shared Temporal Attention Encoder (STAE)}. Its objective is to mitigate the temporal aliasing effects inherent in the RSVP paradigm, as illustrated in Figure~\ref{fig:motivation}(C), by learning to apply greater attention to the most informative temporal segments of the EEG signal.

Our STAE builds upon the EEGProject~\cite{wu2025bridging}, a simple yet effective encoder. To reduce the risk of overfitting, STAE learns a single, parameter-efficient, \textbf{shared} temporal attention vector $\alpha \in \mathbb{R}^{T}$---instead of a complex 2D spatio-temporal map---which is then applied uniformly across all channels to reweight the input brain signal $x_b \in \mathbb{R}^{C \times T}$:
\begin{equation}
x'_b = x_b \odot \text{softmax}(\alpha),
\label{eq:stae}
\end{equation}
where $C$ is the number of channels, $T$ is the number of time steps, and $\odot$ denotes element-wise multiplication with broadcasting. This principled design allows the model to suppress noise from aliasing and amplify critical temporal features. Further architectural details of STAE are provided in Appendix~A.3.

\section{Experiments and Analysis}
\label{sec:experiments}
\subsection{Experimental Setup and Main Results}
\subsubsection{Datasets and Preprocessing.} Our experiments are conducted on two large-scale public datasets: \textbf{THINGS-EEG}~\citep{gifford2022large} and \textbf{THINGS-MEG}~\citep{hebart2023things}, with the details shown in Table~\ref{tab:datasets}.
The EEG dataset contains data from ten participants
who underwent a time-efficient RSVP. The training set includes $1654$ concepts $\times$ $10$ images $\times$ $4$ repetitions. The test set includes $200$ concepts $\times$ $1$ image $\times$ $80$ repetitions. The stimulus onset asynchronies (SOAs) is $200$\,ms, including a stimulus of $100$\,ms followed by a $100$\,ms blank screen. For preprocessing, we epoched EEG data into trials ranging from $0$ to $1000$\,ms after stimulus onset. All EEG repetitions of each image were averaged to ensure a high signal-noise-ratio (SNR). The MEG dataset, including its configuration and preprocessing, follows a similar protocol~\cite{song2025recognizing,wu2025bridging}.

\begin{table}[ht]
\centering
\small
\setlength{\tabcolsep}{1.1mm}

\begin{tabular}{lccccc}    
\toprule    
Type & Sub & Chan & Train$^*$ & Test$^*$ & SOA(ms)  \\
\midrule 
EEG  & 10 & 63 & 1654 $\mid$ 10 $\mid$ 4 & 200 $\mid$ 1 $\mid$ 80 & 200 (100)\\ 
MEG  & 4 & 271 & 1854 $\mid$ 12 $\mid$ 1 & 200 $\mid$ 1 $\mid$ 12 & 1500$\pm$200 (500)\\


\bottomrule 
\multicolumn{6}{l}{$^*$ concepts (classes) $\mid$ conditions (images) $\mid$ repetitions (times).} \\  
\end{tabular}
\caption{Configuration details of the THINGS-EEG and THINGS-MEG datasets (Sub: Subjects, Chan: Channels).}   
\label{tab:datasets} 

\end{table}

In our work, THINGS-EEG served as the primary dataset for deriving main results and analyses, while THINGS-MEG was employed to validate our findings and test the generalizability of our approach on a different neuroimaging modality.

\subsubsection{Vision Encoders.} Our research employs the visual branches of models from OpenCLIP~\cite{ilharco2021openclip} and DINOv2~\cite{oquab2024dinov2learningrobustvisual} as the vision encoder. We conducted extensive experiments across 10 different models (7 from CLIP and 3 from DINOv2). For the detailed comparative analysis within this paper, we focus on two representative encoders from CLIP: RN50 and ViT-L/14. These models were selected to represent different underlying architectures (CNN vs. Transformer) and pre-training data scales (400M vs. 2B pairs), respectively. Unless otherwise specified, RN50 is employed as the default model.

\subsubsection{Brain Encoders.} For the M/EEG branch, we adopt the Shared Temporal Attention Encoder (STAE) as proposed in Section~\ref{sec:stae_encoder}. To further evaluate the generalizability of our method, we additionally conducted experiments with alternative architectures, including ShallowNet~\cite{schirrmeister2017deep}, EEGNet~\cite{lawhern2018eegnet}, TSConv~\cite{song2024decoding} and EEGProject~\cite{wu2025bridging}.

\subsubsection{Baselines.} We compare our proposed ATS framework against a wide range of recent methods, including BraVL~\citep{du2023decoding}, NICE~\citep{song2024decoding}, ATM-S~\citep{li2024visual}, CognitionCapturer~\cite{zhang2025cognitioncapturer}, VE-SDN~\citep{chen2024visualneuraldecodingimproved} and the previous best-performing method UBP~\citep{wu2025bridging}. The positioning of most methods within the broader research landscape is discussed in Section~\ref{sec:related_work}.

Further details on data preprocessing, hyperparameters, hardware configurations and introduction to comparative methods are provided in Appendix~A.

\subsubsection{Main Results.}

As shown in Table~\ref{tab:things_eeg_results}, ATS establishes state-of-the-art performance on the primary THINGS-EEG benchmark, achieving a Top-1 accuracy of \textbf{60.2\%} and surpassing the previous SOTA (UBP) by a significant margin of \textbf{9.8\%}. The effectiveness and generalizability of our ATS framework is further validated on the THINGS-MEG dataset, where it achieves a significant \textbf{3.0\%} Top-1 accuracy improvement. The complete EEG and MEG results for both intra-subject and inter-subject settings are provided in Appendix~B.1.

Our method also substantially improves retrieval accuracy when we use the ViT-L/14 model as the vision encoder. However, the overall performance is lower than that achieved with RN50 by a margin of \textbf{10.0\%}. This finding suggests that an overly powerful ``teacher" model may inadvertently widen the asymmetric modality gap, making the adaptive teaching more challenging and thus leading to a significant decrease in performance. This trend is consistently observed across all 10 tested encoders on THINGS-EEG, as detailed in Appendix~B.2.

\begin{table*}[ht]
  \centering
  \small
  \setlength{\tabcolsep}{1.2pt} 
  \begin{tabular}{lcccccccccccccccccccccc}
  \toprule
  \multirow{2}{*}{Method} & \multicolumn{2}{c}{Subject 1} & \multicolumn{2}{c}{Subject 2} & \multicolumn{2}{c}{Subject 3} & \multicolumn{2}{c}{Subject 4} & \multicolumn{2}{c}{Subject 5} & \multicolumn{2}{c}{Subject 6} & \multicolumn{2}{c}{Subject 7} & \multicolumn{2}{c}{Subject 8} & \multicolumn{2}{c}{Subject 9} & \multicolumn{2}{c}{Subject 10} & \multicolumn{2}{c}{Avg} \\
  \cmidrule(lr){2-3} \cmidrule(lr){4-5} \cmidrule(lr){6-7} \cmidrule(lr){8-9} \cmidrule(lr){10-11} \cmidrule(lr){12-13} \cmidrule(lr){14-15} \cmidrule(lr){16-17} \cmidrule(lr){18-19} \cmidrule(lr){20-21} \cmidrule(lr){22-23}
  & top-1 & top-5 & top-1 & top-5 & top-1 & top-5 & top-1 & top-5 & top-1 & top-5 & top-1 & top-5 & top-1 & top-5 & top-1 & top-5 & top-1 & top-5 & top-1 & top-5 & top-1 & top-5 \\
  \midrule
  BraVL & 6.1 & 17.9 & 4.9 & 14.9 & 5.6 & 17.4 & 5.0 & 15.1 & 4.0 & 13.4 & 6.0 & 18.2 & 6.5 & 20.4 & 8.8 & 23.7 & 4.3 & 14.0 & 7.0 & 19.7 & 5.8 & 17.5 \\
  NICE & 13.2 & 39.5 & 13.5 & 40.3 & 14.5 & 42.7 & 20.6 & 52.7 & 10.1 & 31.5 & 16.5 & 44.0 & 17.0 & 42.1 & 22.9 & 56.1 & 15.4 & 41.6 & 17.4 & 45.8 & 16.1 & 43.6 \\
  NICE-S & 13.3 & 40.2 & 12.1 & 36.1 & 15.3 & 39.6 & 15.9 & 49.0 & 9.8 & 34.4 & 14.2 & 42.4 & 17.9 & 43.6 & 18.2 & 50.2 & 14.4 & 38.7 & 16.0 & 42.8 & 14.7 & 41.7 \\
  NICE-G & 15.2 & 40.1 & 13.9 & 40.1 & 15.7 & 42.7 & 17.6 & 48.9 & 9.0 & 29.7 & 16.3 & 44.4 & 14.9 & 43.1 & 20.3 & 52.1 & 14.1 & 39.7 & 19.6 & 46.7 & 15.6 & 42.8 \\
  MB2C & 23.6  & 56.3 & 22.6 & 50.5 & 26.3 & 60.1 & 34.8 & 67.0 & 21.3 & 53.0 & 31.0 & 62.3 & 25.0 & 54.8 & 39.0 & 69.3 & 27.5 & 59.3 & 33.1 &70.8 & 28.4 & 60.3 \\
  ATM-S & 25.6 & 60.4 & 22.0 & 54.5 & 24.0 & 62.4 & 31.4 & 60.9 & 12.9 & 43.0 & 21.4 & 51.1 & 30.5 & 61.5 & 38.8 & 72.0 & 34.4 & 51.5 & 29.1 & 63.5 & 28.5 & 60.4 \\
  CogCap & 31.4 & 79.6 & 31.4 & 77.8 & 38.1 & 85.6 & 40.3 & 85.8 & 24.4 & 66.3 & 34.8 & 78.7 & 34.6 & 80.9 & 48.1 & 88.6 & 37.4 & 79.3 & 35.5 & 79.2 & 35.6 & 80.2 \\
  VE-SDN & 32.6 & 63.7 & 34.4 & 69.9 & 38.7 & 73.5 & 39.8 & 72.0 & 29.4 & 58.6 & 34.5 & 68.8 & 34.5 & 68.3 & 49.3 & 79.8 & 39.0 & 69.6 & 39.8 & 75.3 & 37.2 & 69.9 \\
  UBP & 40.5 & 71.0 & 49.5 & 82.5 & 49.5 & 82.0 & 49.5 & 76.0 & 45.0 & 73.0 & 56.5 & 83.0 & 48.5 & 80.0 & 57.0 & 86.0 & 44.0 & 76.0 & 64.0 & 87.5 & 50.4 & 79.7 \\
  \midrule 
  \textbf{ATS} & \textbf{53.0} & \textbf{79.0} & \textbf{62.0} & \textbf{87.5} & \textbf{61.5} & \textbf{89.0} & \textbf{57.0} & \textbf{86.5} & \textbf{55.0} & \textbf{84.0} & \textbf{68.0} & \textbf{90.5} & \textbf{53.0} & \textbf{84.0} & \textbf{66.5} & \textbf{91.0} & \textbf{58.5} & \textbf{86.0} & \textbf{67.5} & \textbf{89.0} & \textbf{60.2} & \textbf{86.7} \\

  \bottomrule
  \end{tabular}
  \caption{Top-1 and Top-5 accuracy (\%) for 200-way zero-shot retrieval on THINGS-EEG.}
  \label{tab:things_eeg_results}
  \end{table*}

\subsection{Validating the Adaptive Teaching Paradigm}
Building on the significant overall performance gains of ATS, we now systematically dissect and validate our core contribution: the adaptive teaching paradigm and its implementation via the ShrinkAdapter.

\begin{figure}[ht]
  \centering
  \includegraphics[width=0.98\linewidth]{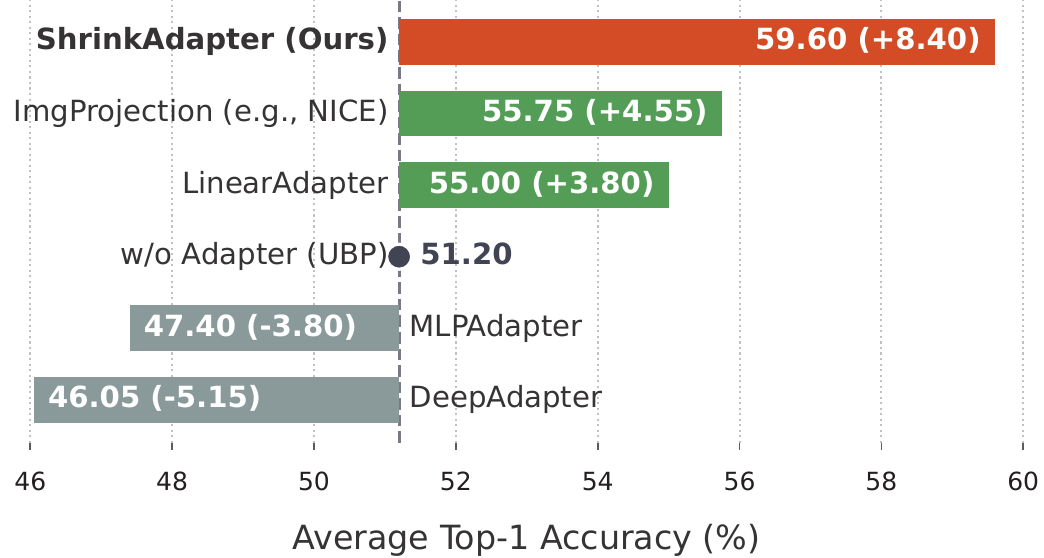}
  \caption{Performance comparison of Adapter architectures on the THINGS-EEG Dataset.}
  \label{fig:adapter_comparison}
\end{figure}

As shown in Figure~\ref{fig:adapter_comparison}, our proposed \textbf{ShrinkAdapter} significantly outperforms both the w/o Adapter baseline and other conventional adapter designs. This overall superior performance highlights that the design of the ShrinkAdapter itself is crucial.

\subsubsection{The Necessity of Freedom of Adaptation.}

Based on our theory of adaptive teaching, we argue that any design choice intended to preserve the teacher's original knowledge is in fact harmful. For an effective alignment, the teacher must be granted the freedom to fully adapt its knowledge structure. We validate this principle from two perspectives: architectural constraints and loss-based constraints.

First, we investigate architectural constraints by ablating the residual connection, a common operation in projection layers (ImgProjection) used to enforce feature preservation. As shown in Table~\ref{tab:residual_ablation_final}, removing the residual connection leads to a consistent and significant improvement across all ShrinkAdapter configurations. 

\begin{figure*}[ht]
  \centering
  \includegraphics[width=1.0\linewidth]{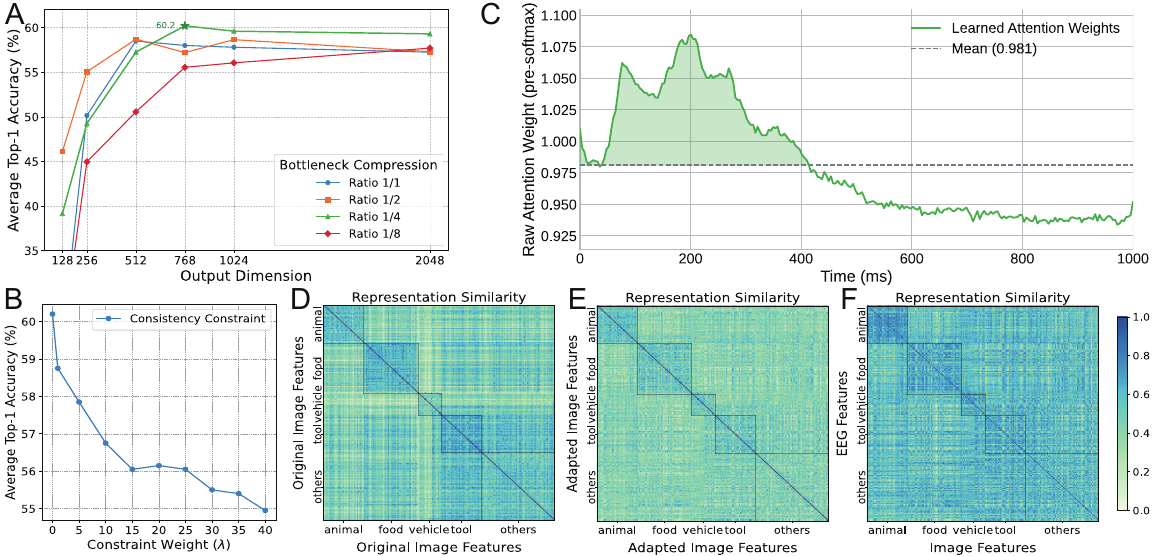}
  \caption{(A) Optimal ShrinkAdapter structure search. (B) Negative impact of consistency constraints. (C) Learned temporal attention weights.
  (D-F) Representation Similarity Matrices (RSMs) showing: (D) self-similarity of original visual features, (E) self-similarity of adapted visual features, and (F) the cross-modal similarity between EEG and adapted visual features.}
  \label{fig:multiplot}
\end{figure*}

Next, we examine the effect of an explicit loss-based constraint. Inspired by the multimodal similarity-keeping strategy proposed in~\cite{chen2024mindseyeimagerecognition}, we introduce a semantic distribution consistency loss to our main objective. The loss is defined as:
\begin{equation}
\label{eq:consistency_loss}
\mathcal{L}_{\text{total}} = \mathcal{L}_{\text{SCE}} + \lambda \cdot \left[ 1 - \text{sim}_{\text{cos}}(\mathbf{M}_{h_v}, \mathbf{M}_{z_v}) \right],
\end{equation}
where $\mathbf{M}_{h_v} = \text{sim}_{\text{cos}}(h_v, h_v)$ is the self-similarity matrix of the original CLIP features, representing their intrinsic semantic distribution. $\mathbf{M}_{z_v}$ is the equivalent for the adapted features. The final cosine similarity term thus measures the consistency of these semantic distributions before and after adaptation, while $\lambda$ controls the strength of this constraint. Figure~\ref{fig:multiplot}(B)  shows a clear trend: as $\lambda$ increases (strengthening semantic preservation), Top-1 accuracy drops steadily.

Taken together, these experiments—both architectural and loss-based—strongly support our main claim: for effective asymmetric alignment, the teacher modality must have the freedom to adjust its knowledge structure fundamentally. This validates the core of our Adaptive Teaching Paradigm.

\begin{table}[h]
\centering

\small 
\setlength{\tabcolsep}{1.3mm} 
\begin{tabular}{l ccc ccc}
\toprule
\multirow{2}{*}{\textbf{Ratio}} & \multicolumn{3}{c}{\textbf{Avg. Top-1 Acc (\%)}} & \multicolumn{3}{c}{\textbf{Avg. Top-5 Acc (\%)}} \\
\cmidrule(lr){2-4} \cmidrule(lr){5-7}
& w/ Res & w/o Res & Improv. & w/ Res & w/o Res & Improv. \\
\midrule
1:1 & 54.35 & \textbf{57.80} & +3.45 & 83.85 & \textbf{85.90} & +2.05 \\
1:2 & 54.45 & \textbf{58.65} & +4.20 & 83.45 & \textbf{86.25} & +2.80 \\
1:4 & 54.05 & \textbf{59.60} & +5.55 & 83.25 & \textbf{87.55} & +4.30 \\
1:8 & 53.60 & \textbf{56.05} & +2.45 & 82.75 & \textbf{86.70} & +3.95 \\
\bottomrule
\end{tabular}
\caption{Ablation study of residual connection in the ShrinkAdapter. The first column denotes the bottleneck compression ratio (e.g., 1:4).}
\label{tab:residual_ablation_final}
\end{table}

\subsubsection{The Necessity of Filtering Redundant Information.}

Based on our analysis of the asymmetric modality gap, we believe the visual modality is information-redundant relative to the EEG modality. To address this, we introduce a bottleneck structure inspired by the IB principle. We further hypothesize that alignment performance is sensitive to the choice of latent space dimension. To validate these hypotheses, we evaluate two key hyperparameters of our ShrinkAdapter: the bottleneck compression ratio and its output dimension, which determines the size of the latent space.

The results in Figure~\ref{fig:multiplot}(A) provide strong empirical support for our hypotheses. We observe a clear optimal configuration at a \textbf{1/4 bottleneck ratio} and a \textbf{768-dimensional latent space}, which achieves the best performance. Notably, excessive compression (e.g., a 1/8 ratio) leads to a sharp drop in accuracy due to over-filtering of essential information, while omitting compression entirely (1/1 ratio) also underperforms the optimal setting. Furthermore, simply increasing the latent space dimension to 2048 does not yield additional gains, indicating that increased capacity does not necessarily translate to better performance. These findings highlight the importance of a well-calibrated bottleneck and confirm the thoroughness of our hyperparameter search.

In summary, the experimental results validate the rationale of the adaptive teaching paradigm and demonstrate the effectiveness of our proposed design.

\subsection{Analysis of the Shared Temporal Attention Encoder (STAE)}
\label{sec:stae_analysis}

We validate the effectiveness of STAE by comparing it with established EEG encoders. As shown in Table~\ref{tab:eegencoder_comparison}, our proposed encoder consistently outperforms both classic and recent methods.

\begin{table}[h]
\centering
\small

\begin{tabular}{lcc}
\toprule
\textbf{EEG Encoder} & {\textbf{Avg. Top-1}} & {\textbf{Avg. Top-5}} \\
\midrule
EEGNet       & 25.65 & 57.70 \\
ShallowNet   & 31.30 & 65.25 \\
TSConv (NICE)       & 44.85 & 76.75 \\
EEGProject (UBP)   & 56.75 & 84.30 \\
\midrule 
\textbf{STAE (ours)} & \textbf{60.20} & \textbf{86.65} \\
\bottomrule
\end{tabular}
\caption{Comparison of different EEG encoder architectures.}
\label{tab:eegencoder_comparison}
\end{table}

To provide insight into the mechanism behind this success, we visualized the learned attention weights in Figure~\ref{fig:multiplot}(C). The visualization reveals that STAE automatically learns to focus its processing on the $50-400$\,ms window after stimulus onset. This learned temporal focus is both significant and neuroscientifically plausible. The start of this window at $\sim$50\,ms aligns with the latency for the initial feedforward sweep of visual information to travel from the retina to the primary visual cortex~\cite{thorpe1996speed}. Furthermore, the sustained attention across this extended period is highly consistent with the temporal dynamics of visual object recognition reported in previous studies~\cite{gifford2022large}. Ultimately, these qualitative results reveal the mechanism underlying our performance gains and demonstrate the neuroscientific plausibility of our model through consistency with established findings.

\subsection{Qualitative Analysis of the ATS}
To dissect the mechanisms underlying adaptive teaching and successful retrieval, we employ Representational Similarity Analysis (RSA)~\cite{cichy2020eeg} on the test set, yielding the representational similarity matrices (RSMs) by averaging all subjects. Additionally, we grouped the $200$ test concepts into five superordinate categories: animal, food, vehicle, tool, and others, following~\cite{song2024decoding}.

As shown in Figure~\ref{fig:multiplot}(D), the original image feature $h_v$, acting as the ``teacher," exhibits strong intra-category aggregation (bright diagonal blocks), confirming its highly semantic nature. Notably, it also contains nuanced inter-category similarity, which we consider redundant semantic knowledge that exceeds the ``student" modality's capacity. After being refined by our adaptive teaching paradigm, as shown in Figure~\ref{fig:multiplot}(E), this redundant information is effectively filtered out, while the core conceptual semantics are preserved. This result visually confirms that our adaptive teaching paradigm operates as intended.

A central question remains: is the discrimination we obtain from EEG driven by high-level semantic knowledge or low-level perceptual features (e.g., color, texture, spatial layout)? As shown in Figure~\ref{fig:multiplot}(F), the bright diagonal blocks in the cross-modal RSM confirm that the features extracted from EEG are semantically meaningful at the category level. Furthermore, the RSM reveals notable off-diagonal inter-class similarity. An analysis of the top-5 retrieval results suggests this is because the model also leverages shared low-level features—such as color, texture, and orientation—for retrieval (Retrieval Case Analysis detailed in Appendix B.3). The bright main diagonal in the cross-modal RSM explains our high 60.2\% Top-1 retrieval accuracy. In summary, our adaptive teaching enables the model to extract a hybrid representation from EEG, encoding both conceptual semantics and fundamental perceptual properties.


\section{Conclusion}
\label{sec:conclusion}

In this work, we reframed the vision-brain alignment problem by identifying its fundamental asymmetry, which we deconstructed into Fidelity Gap and Semantic Gap. To bridge this vision-brain gap, we proposed the Adaptive Teaching Paradigm, a conceptual shift that empowers the visual ``teacher" to dynamically shrink and adapt its knowledge to match the capacity of the EEG ``student". We implemented this with the ShrinkAdapter, a simple yet effective module whose residual-free, bottleneck design is guided by the Information Bottleneck principle. We also proposed the Shared Temporal Attention Encoder to mitigate temporal noise. Through extensive experiments, we validated the underlying rationale and effectiveness of our paradigm and core components. Our approach not only achieved state-of-the-art performance by a remarkable margin but also provides valuable insights for asymmetric alignment tasks.

\setcounter{secnumdepth}{0}
\section{Acknowledgments}
This work was supported in part by the National Natural Science Foundation of China under Grants No. U25A20531 and U22A2096. We gratefully acknowledge the High Performance Computing Center of Xidian University for providing computational resources for this work.

\bibliography{main}

@article{michel2012towards,
  title={Towards the utilization of EEG as a brain imaging tool},
  author={Michel, Christoph M and Murray, Micah M},
  journal={Neuroimage},
  volume={61},
  number={2},
  pages={371--385},
  year={2012},
  publisher={Elsevier}
}

@article{grootswagers2019representational,
  title={The representational dynamics of visual objects in rapid serial visual processing streams},
  author={Grootswagers, Tijl and Robinson, Amanda K and Carlson, Thomas A},
  journal={NeuroImage},
  volume={188},
  pages={668--679},
  year={2019},
  publisher={Elsevier}
}

@article{keysers2001speed,
  title={The speed of sight},
  author={Keysers, Christian and Xiao, D-K and F{\"o}ldi{\'a}k, Peter and Perrett, David I},
  journal={Journal of cognitive neuroscience},
  volume={13},
  number={1},
  pages={90--101},
  year={2001},
  publisher={MIT Press One Rogers Street, Cambridge, MA 02142-1209, USA journals-info~…}
}

@article{song2025recognizing,
  title={Recognizing Natural Images From EEG With Language-Guided Contrastive Learning},
  author={Song, Yonghao and Wang, Yijun and He, Huiguang and Gao, Xiaorong},
  journal={IEEE Transactions on Neural Networks and Learning Systems},
  year={2025},
  publisher={IEEE}
}

@inproceedings{wei2024mb2c,
  title={MB2C: Multimodal Bidirectional Cycle Consistency for Learning Robust Visual Neural Representations},
  author={Wei, Yayun and Cao, Lei and Li, Hao and Dong, Yilin},
  booktitle={Proceedings of the 32nd ACM International Conference on Multimedia},
  pages={8992--9000},
  year={2024}
}

@misc{chen2024visualneuraldecodingimproved,
      title={Visual Neural Decoding via Improved Visual-EEG Semantic Consistency}, 
      author={Hongzhou Chen and Lianghua He and Yihang Liu and Longzhen Yang},
      year={2024},
      eprint={2408.06788},
      archivePrefix={arXiv},
      primaryClass={cs.CV},
      url={https://arxiv.org/abs/2408.06788}, 
}

@inproceedings{wu2025bridging,
  title={Bridging the Vision-Brain Gap with an Uncertainty-Aware Blur Prior},
  author={Wu, Haitao and Li, Qing and Zhang, Changqing and He, Zhen and Ying, Xiaomin},
  booktitle={Proceedings of the Computer Vision and Pattern Recognition Conference},
  pages={2246--2257},
  year={2025}
}

@inproceedings{zhang2025cognitioncapturer,
  title={Cognitioncapturer: Decoding visual stimuli from human eeg signal with multimodal information},
  author={Zhang, Kaifan and He, Lihuo and Jiang, Xin and Lu, Wen and Wang, Di and Gao, Xinbo},
  booktitle={Proceedings of the AAAI Conference on Artificial Intelligence},
  pages={14486--14493},
  year={2025}
}

@inproceedings{he2016deep,
  title={Deep residual learning for image recognition},
  author={He, Kaiming and Zhang, Xiangyu and Ren, Shaoqing and Sun, Jian},
  booktitle={Proceedings of the IEEE conference on computer vision and pattern recognition},
  pages={770--778},
  year={2016}
}

@article{ren2021reconstructing,
  title={Reconstructing seen image from brain activity by visually-guided cognitive representation and adversarial learning},
  author={Ren, Ziqi and Li, Jie and Xue, Xuetong and Li, Xin and Yang, Fan and Jiao, Zhicheng and Gao, Xinbo},
  journal={NeuroImage},
  volume={228},
  pages={117602},
  year={2021},
  publisher={Elsevier}
}

@inproceedings{takagi2023high,
  title={High-resolution image reconstruction with latent diffusion models from human brain activity},
  author={Takagi, Yu and Nishimoto, Shinji},
  booktitle={Proceedings of the IEEE/CVF conference on computer vision and pattern recognition},
  pages={14453--14463},
  year={2023}
}

@misc{kumar2022finetuningdistortpretrainedfeatures,
      title={Fine-Tuning can Distort Pretrained Features and Underperform Out-of-Distribution}, 
      author={Ananya Kumar and Aditi Raghunathan and Robbie Jones and Tengyu Ma and Percy Liang},
      year={2022},
      eprint={2202.10054},
      archivePrefix={arXiv},
      primaryClass={cs.LG},
      url={https://arxiv.org/abs/2202.10054}, 
}

@article{wang2021combating,
  title={Combating noise: semi-supervised learning by region uncertainty quantification},
  author={Wang, Zhenyu and Li, Ya-Li and Guo, Ye and Wang, Shengjin},
  journal={Advances in Neural Information Processing Systems},
  volume={34},
  pages={9534--9545},
  year={2021}
}

@misc{tishby2000informationbottleneckmethod,
      title={The information bottleneck method}, 
      author={Naftali Tishby and Fernando C. Pereira and William Bialek},
      year={2000},
      eprint={physics/0004057},
      archivePrefix={arXiv},
      primaryClass={physics.data-an},
      url={https://arxiv.org/abs/physics/0004057}, 
}

@inproceedings{li2024visual,
 author = {Li, Dongyang and Wei, Chen and Li, Shiying and Zou, Jiachen and Liu, Quanying},
 booktitle = {Advances in Neural Information Processing Systems},
 pages = {102822--102864},
 publisher = {Curran Associates, Inc.},
 title = {Visual Decoding and Reconstruction via EEG Embeddings with Guided Diffusion},
 url = {https://proceedings.neurips.cc/paper_files/paper/2024/file/ba5f1233efa77787ff9ec015877dbd1f-Paper-Conference.pdf},
 volume = {37},
 year = {2024}
}

@misc{ilharco2021openclip,
  author = {Ilharco, Gabriel and Wortsman, Mitchell and Wightman, Ross and Gordon, Cade and Carlini, Nicholas and Taori, Rohan and Dave, Achal and Shankar, Vaishaal and Namkoong, Hongseok and Miller, John and Hajishirzi, Hannaneh and Farhadi, Ali and Schmidt, Ludwig},
  title = {OpenCLIP},
  howpublished = {\url{https://doi.org/10.5281/zenodo.5143773}},
  year = {2021},
  note = {Accessed: 2025-07-30}
}

@article{schirrmeister2017deep,
  title={Deep learning with convolutional neural networks for EEG decoding and visualization},
  author={Schirrmeister, Robin Tibor and Springenberg, Jost Tobias and Fiederer, Lukas Dominique Josef and Glasstetter, Martin and Eggensperger, Katharina and Tangermann, Michael and Hutter, Frank and Burgard, Wolfram and Ball, Tonio},
  journal={Human brain mapping},
  volume={38},
  number={11},
  pages={5391--5420},
  year={2017},
  publisher={Wiley Online Library}
}

@article{lawhern2018eegnet,
  title={EEGNet: a compact convolutional neural network for EEG-based brain--computer interfaces},
  author={Lawhern, Vernon J and Solon, Amelia J and Waytowich, Nicholas R and Gordon, Stephen M and Hung, Chou P and Lance, Brent J},
  journal={Journal of neural engineering},
  volume={15},
  number={5},
  pages={056013},
  year={2018},
  publisher={iOP Publishing}
}

@misc{chen2024mindseyeimagerecognition,
      title={Mind's Eye: Image Recognition by EEG via Multimodal Similarity-Keeping Contrastive Learning}, 
      author={Chi-Sheng Chen and Chun-Shu Wei},
      year={2024},
      eprint={2406.16910},
      archivePrefix={arXiv},
      primaryClass={eess.SP},
      url={https://arxiv.org/abs/2406.16910}, 
}

@article{thorpe1996speed,
  title={Speed of processing in the human visual system},
  author={Thorpe, Simon and Fize, Denis and Marlot, Catherine},
  journal={nature},
  volume={381},
  number={6582},
  pages={520--522},
  year={1996},
  publisher={Nature Publishing Group UK London}
}

@article{cichy2020eeg,
  title={AM/EEG-fMRI fusion primer: resolving human brain responses in space and time},
  author={Cichy, Radoslaw M and Oliva, Aude},
  journal={Neuron},
  volume={107},
  number={5},
  pages={772--781},
  year={2020},
  publisher={Elsevier}
}

@article{scotti2024reconstructing,
  title={Reconstructing the mind's eye: fMRI-to-image with contrastive learning and diffusion priors},
  author={Scotti, Paul and Banerjee, Atmadeep and Goode, Jimmie and Shabalin, Stepan and Nguyen, Alex and Dempster, Aidan and Verlinde, Nathalie and Yundler, Elad and Weisberg, David and Norman, Kenneth and others},
  journal={Advances in Neural Information Processing Systems},
  volume={36},
  year={2024}
}

@inproceedings{song2024decoding,
  title = {Decoding {{Natural Images}} from {{EEG}} for {{Object Recognition}}},
  author = {Song, Yonghao and Liu, Bingchuan and Li, Xiang and Shi, Nanlin and Wang, Yijun and Gao, Xiaorong},
  booktitle = {International {{Conference}} on {{Learning Representations}}},
  year = {2024},
}

@article{gifford2022large,
  title={A large and rich EEG dataset for modeling human visual object recognition},
  author={Gifford, Alessandro T and Dwivedi, Kshitij and Roig, Gemma and Cichy, Radoslaw M},
  journal={NeuroImage},
  volume={264},
  pages={119754},
  year={2022},
  publisher={Elsevier}
}

@misc{oord2019representationlearningcontrastivepredictive,
      title={Representation Learning with Contrastive Predictive Coding}, 
      author={Aaron van den Oord and Yazhe Li and Oriol Vinyals},
      year={2019},
      eprint={1807.03748},
      archivePrefix={arXiv},
      primaryClass={cs.LG},
      url={https://arxiv.org/abs/1807.03748}, 
}

@article{hebart2023things,
  title={THINGS-data, a multimodal collection of large-scale datasets for investigating object representations in human brain and behavior},
  author={Hebart, Martin N and Contier, Oliver and Teichmann, Lina and Rockter, Adam H and Zheng, Charles Y and Kidder, Alexis and Corriveau, Anna and Vaziri-Pashkam, Maryam and Baker, Chris I},
  journal={Elife},
  volume={12},
  pages={e82580},
  year={2023},
  publisher={eLife Sciences Publications Limited}
}

@inproceedings{radford2021learning,
  title={Learning transferable visual models from natural language supervision},
  author={Radford, Alec and Kim, Jong Wook and Hallacy, Chris and Ramesh, Aditya and Goh, Gabriel and Agarwal, Sandhini and Sastry, Girish and Askell, Amanda and Mishkin, Pamela and Clark, Jack and others},
  booktitle={International conference on machine learning},
  pages={8748--8763},
  year={2021},
  organization={PMLR}
}

@article{du2023decoding,
  title={Decoding Visual Neural Representations by Multimodal Learning of Brain-Visual-Linguistic Features},
  author={Du, Changde and Fu, Kaicheng and Li, Jinpeng and He, Huiguang},
  journal={IEEE Transactions on Pattern Analysis and Machine Intelligence},
  year={2023},
  publisher={IEEE}
}

@inproceedings{ijcai2019p192,
  title     = {Decoding EEG by Visual-guided Deep Neural Networks},
  author    = {Jiao, Zhicheng and You, Haoxuan and Yang, Fan and Li, Xin and Zhang, Han and Shen, Dinggang},
  booktitle = {Proceedings of the Twenty-Eighth International Joint Conference on
               Artificial Intelligence, {IJCAI-19}},
  publisher = {International Joint Conferences on Artificial Intelligence Organization},
  pages     = {1387--1393},
  year      = {2019},
  month     = {7},
  doi       = {10.24963/ijcai.2019/192},
  url       = {https://doi.org/10.24963/ijcai.2019/192},
}

@inproceedings{spampinato2017deep,
  title={Deep learning human mind for automated visual classification},
  author={Spampinato, Concetto and Palazzo, Simone and Kavasidis, Isaak and Giordano, Daniela and Souly, Nasim and Shah, Mubarak},
  booktitle={Proceedings of the IEEE conference on computer vision and pattern recognition},
  pages={6809--6817},
  year={2017}
}

@misc{oquab2024dinov2learningrobustvisual,
      title={DINOv2: Learning Robust Visual Features without Supervision}, 
      author={Maxime Oquab and Timothée Darcet and Théo Moutakanni and Huy Vo and Marc Szafraniec and Vasil Khalidov and Pierre Fernandez and Daniel Haziza and Francisco Massa and Alaaeldin El-Nouby and Mahmoud Assran and Nicolas Ballas and Wojciech Galuba and Russell Howes and Po-Yao Huang and Shang-Wen Li and Ishan Misra and Michael Rabbat and Vasu Sharma and Gabriel Synnaeve and Hu Xu and Hervé Jegou and Julien Mairal and Patrick Labatut and Armand Joulin and Piotr Bojanowski},
      year={2024},
      eprint={2304.07193},
      archivePrefix={arXiv},
      primaryClass={cs.CV},
      url={https://arxiv.org/abs/2304.07193}, 
}
\clearpage

\setcounter{secnumdepth}{2}

\twocolumn[ 
\begin{center} 

    {\huge\bfseries Supplementary Material of\par} 
    \vspace{0.5em}
    {\huge\bfseries Shrinking the Teacher: An Adaptive Teaching Paradigm for Asymmetric EEG-Vision Alignment \par} 
    \vspace{2.0em}

    {\Large\bfseries 
        Lukun Wu\textsuperscript{\rm 1},
        Jie Li\textsuperscript{\rm 1},
        Ziqi Ren\textsuperscript{\rm 2},
        Kaifan Zhang\textsuperscript{\rm 1},
        Xinbo Gao\textsuperscript{\rm 1}
        \par
    }
    \vspace{1.0em}

    {\large 
        \textsuperscript{\rm 1}School of Electronic Engineering, Xidian University, Xi'an, China\\
        \textsuperscript{\rm 2}School of Life Science and Technology, Xidian University, Xi'an, China\\
        lkwu@stu.xidian.edu.cn, \{leejie, xbgao\}@mail.xidian.edu.cn
        \par
    }
\end{center}
\vspace{3em} 
] 

\appendix  

\setcounter{table}{0}      
\setcounter{figure}{0}     
\setcounter{algorithm}{0}  
\setcounter{equation}{0}   
\section{Experimental details}

\subsection{Datasets details}
The core dataset settings are introduced in the main paper; here we provide additional details.

\subsubsection{THINGS-EEG.} THINGS-EEG~\citep{gifford2022large} is a large-scale EEG dataset, collected using the time-efficient Rapid Serial Visual Presentation (RSVP) paradigm~\citep{keysers2001speed,grootswagers2019representational}. Data are recorded using 64-channel EASYCAP equipment with the standard 10-10 system.

\begin{figure}[h]
  \centering
    \includegraphics[width=0.80\linewidth]{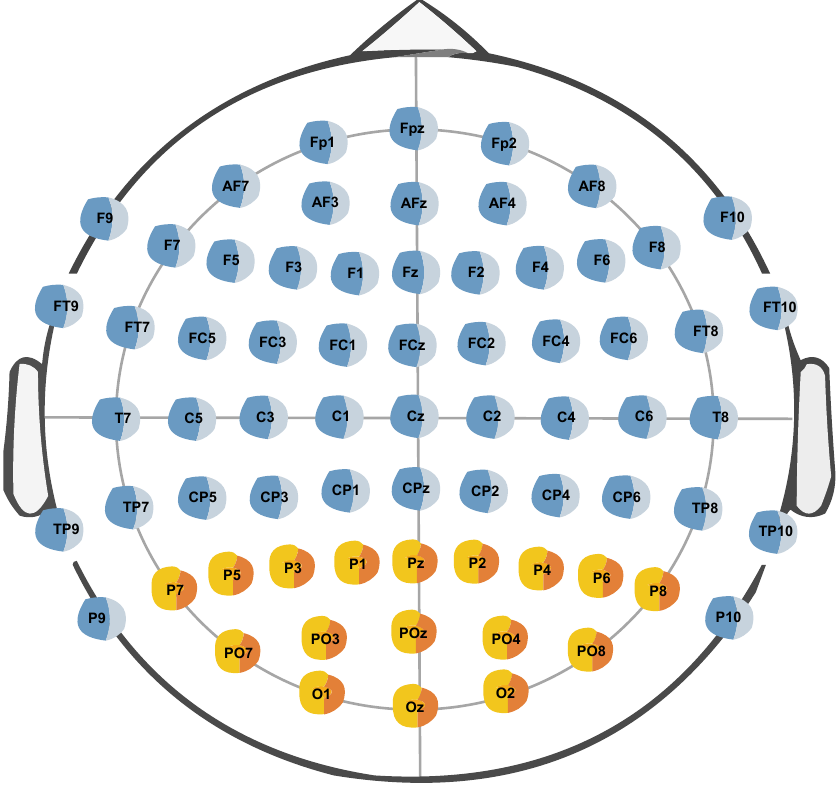}
    \caption{The 17 selected EEG channels from the occipital and parietal lobes used in our study, highlighted in orange. The layout follows the standard 10-10 system.}
    \label{fig:channels}
\end{figure}

For preprocessing, we follow the method in~\cite{wu2025bridging}: raw EEG data (63 channels, 1000 Hz sample rate) is filtered to [0.1, 100] Hz, epoched into trials from 0 to 1000 ms after stimuli onset (with baseline correction using the prior 200 ms average), and downsampled to 250 Hz. While we observe that downsampling to 100 Hz also yields strong performance, we retain the 250 Hz sampling rate for all experiments to ensure a fair and direct comparison with the baseline methods. Following UBP, we retain 17 channels over the occipital and parietal cortex (related to visual processing): P7, P5, P3, P1, Pz, P2, P4, P6, P8, PO7, PO3, POz, PO4, PO8, O1, Oz, O2, as illustrated in Figure~\ref{fig:channels}. The data is stored in float16 format.

\subsubsection{THINGS-MEG.} THINGS-MEG~\cite{hebart2023things} includes four participants and is characterized by 271 channels. Raw MEG data are epoched into trials from 0 to 1000 ms after stimuli onset. Its experimental design uses a relatively long stimulus duration of 500 ms, followed by a blank screen lasting 1000 ± 200 ms. For preprocessing, we apply a [0.1, 100] Hz band-pass filter and perform baseline correction after downsampling the data to 200 Hz. The data is stored in float16 format.

We included both the THINGS-EEG and THINGS-MEG datasets for two key reasons. Most importantly, both datasets clearly exhibit the ``Asymmetric Gap"—the foundational problem our ATS is designed to address. Secondly, their shared protocols and visual stimuli provide a valuable opportunity to validate the robustness of our approach across different, yet related, neuroimaging modalities, despite their distinct signal dynamics.

\subsection{Implementation details}
\subsubsection{Environment.}
Our implementation is based on Python 3.8.19, CUDA 12.2, and PyTorch 2.4.1. All dependencies are detailed in the ats.yml file within the project repository. Experiments are conducted on a server equipped with an Intel(R) Xeon(R) Gold 6330 CPU and 256 GB of system memory, using eight NVIDIA GeForce RTX 3090 GPUs for most tasks. However, the MEG experiments (Appendix~B.2) were run on a configuration with eight NVIDIA GeForce RTX 4090 GPUs.
\subsubsection{Training Configuration.}
We adopt a uniform training protocol across all experiments. All models are trained for 150 epochs with a batch size of 1024. We use the AdamW optimizer with a weight decay of 1e-4. The learning rate is initialized at 1e-4 and decayed by a factor of 0.1 every 50 epochs using a step-wise scheduler. To prevent overfitting, we employ an early stopping strategy that halts training when validation performance ceases to improve. For the uncertainty-aware prior blur in UBP~\cite{wu2025bridging}, we retain the original parameter configurations.

\subsubsection{Robustness and Training Protocol Analysis.}

\begin{table*}[t]
  \centering
  \small
  \setlength{\tabcolsep}{10pt} 
  \begin{tabular}{llcccccc}
    \toprule
    \multirow{2}{*}{Training Protocol} & \multirow{2}{*}{Method} & \multicolumn{2}{c}{Mean ± SD (\%)} & \multicolumn{2}{c}{95\% CI (\%)} & \multicolumn{2}{c}{Cohen's d} \\
    \cmidrule(lr){3-4} \cmidrule(lr){5-6} \cmidrule(lr){7-8}
                                       &                         & Top-1             & Top-5             & Top-1                 & Top-5                 & Top-1     & Top-5     \\
    \midrule
    \multicolumn{8}{c}{\textbf{EEG Intra-subject}: train and test on one subject} \\
    \midrule
    \multirow{2}{*}{UBP's Protocol}    & UBP                     & 50.81 ± 0.42      & 79.60 ± 0.37      & {[50.37, 51.25]}      & {[79.21, 79.99]}      & \multirow{2}{*}{8.55}  & \multirow{2}{*}{15.56} \\
                                       & ATS                     & 56.60 ± 0.72      & 86.32 ± 0.33      & {[55.85, 57.35]}      & {[85.98, 86.67]}      &                        &                        \\
    \midrule
    \multirow{2}{*}{ATS's Protocol}    & UBP                     & 51.09 ± 0.58      & 79.52 ± 0.24      & {[50.48, 51.70]}      & {[79.26, 79.77]}      & \multirow{2}{*}{7.80}  & \multirow{2}{*}{20.96} \\
                                       & ATS                     & 59.07 ± 0.68      & 87.15 ± 0.33      & {[58.36, 59.78]}      & {[86.80, 87.50]}      &                        &                        \\
    \midrule
    \multicolumn{8}{c}{\textbf{MEG Intra-subject}: train and test on one subject} \\
    \midrule
    \multirow{2}{*}{UBP's Protocol}    & UBP                     & 26.87 ± 1.08      & 55.81 ± 1.12      & {[25.74, 28.01]}      & {[54.64, 56.99]}      & \multirow{2}{*}{2.91}  & \multirow{2}{*}{1.72}  \\
                                       & ATS                     & 30.19 ± 1.12      & 58.59 ± 1.23      & {[29.01, 31.36]}      & {[57.30, 59.87]}      &                        &                        \\
    \midrule
    \multirow{2}{*}{ATS's Protocol}    & UBP                     & 29.79 ± 0.55      & 57.69 ± 0.89      & {[29.22, 30.37]}      & {[56.75, 58.62]}      & \multirow{2}{*}{2.40}  & \multirow{2}{*}{2.86}  \\
                                       & ATS                     & 33.27 ± 1.39      & 61.29 ± 1.05      & {[31.81, 34.74]}      & {[60.19, 62.39]}      &                        &                        \\
    \bottomrule
    \end{tabular}
    \caption{Robustness and training protocol analysis on the THINGS-EEG and THINGS-MEG datasets. We compare ATS with the UBP baseline under two different training protocols. All results are averaged over 6 random seeds. Our method (ATS) consistently outperforms the baseline across both datasets and protocols, with Cohen's d values indicating a large to very large effect size. This highlights the statistical significance and robustness of our approach.}
    \label{tab:robustness}
  
\end{table*}
While the main paper reports results from a single run (seed=0) under our proposed protocol (150 epochs with a scheduler), we conducted a more rigorous analysis here to validate the robustness of our method. We re-evaluated the key comparisons on both datasets across six different random seeds, calculating the mean, standard deviation (SD), 95\% confidence intervals (CI), and Cohen's d for effect size. As detailed in Table~\ref{tab:robustness}, the results confirm that the performance gains from ATS are statistically robust across different protocols, consistently demonstrating large effect sizes.
\subsubsection{Attribution Analysis.}
To disentangle the contributions of our proposed components (the ShrinkAdapter and STAE) from the blur prior introduced in UBP, we conducted a step-wise attribution analysis. We evaluated performance under two conditions: with and without the blur prior. As shown in Table~\ref{tab:attribution}, both the ShrinkAdapter and STAE provide significant performance gains independently. Notably, building on the effective blur prior, ATS further adaptively and flexibly bridges the modality gap to achieve the largest overall improvement. Even without the blur prior, the components of ATS remain significantly effective, demonstrating the robustness of our adaptive teaching paradigm.

\begin{table}[htbp]
\centering
\small
\setlength{\tabcolsep}{3pt}
\begin{tabular}{lccc}
\toprule
\textbf{Configuration} & \textbf{UBP} & \textbf{+ ShrinkAdapter} & \textbf{+ STAE (Full ATS)} \\
\midrule
w/ Blur Prior          & 50.4         & 56.8 (+6.4)              & 59.6 (+9.2)                \\
w/o Blur Prior         & 42.3         & 45.7 (+3.4)              & 48.2 (+5.9)                \\
\bottomrule
\end{tabular}
\caption{Attribution analysis of ATS components. We report the Top-1 accuracy (\%) and the cumulative improvement over the UBP baseline. Each component consistently contributes to the performance gain, both with and without the UBP blur prior.}
\label{tab:attribution}
\end{table}
\subsubsection{Additional Introduction to Baseline Methods.}
To address poor generalization to unseen classes, Du et al.~\citep{du2023decoding} introduced BraVL. Attributing the issue to underutilized multimodal semantics, their approach uses a Multi-Modal Variational Autoencoder (MMVAE) with a Mixture-of-Products-of-Experts (MoPoE) to better align and extract brain-visual-linguistic features. Li et al.~\citep{li2024visual} proposed an end-to-end EEG-based visual reconstruction zero-shot framework, featuring a tailored brain encoder called the Adaptive Thinking Mapper (ATM). Their work represents the first framework to comprehensively integrate classification, retrieval, and reconstruction tasks for EEG-based visual decoding.
\subsubsection{Hyperparameter Search.}
We determine our key hyperparameters by performing a small-scale grid search centered around the settings reported by UBP. First, with a fixed batch size of 1024, we test learning rates of \{1e-3, 1e-4, 1e-5\} and find 1e-4 to be optimal. Subsequently, fixing the learning rate at 1e-4, we evaluate batch sizes of \{512, 1024, 2048, 4096\}, with 1024 yielding the best performance. Therefore, we adopt a batch size of 1024 and a learning rate of 1e-4 for our experiments.

\subsubsection{Ablation Study of ShrinkAdapter Components.}

We conduct a comprehensive ablation study to validate the design of our ShrinkAdapter. This study systematically evaluates the contributions of its essential components, whose design rationales are detailed in the main paper. The ablations cover not only core architectural choices like the residual connection and bottleneck architecture, but also the impact of standard components such as Dropout, LayerNorm, and the GELU activation function. To ensure a fair evaluation of the residual connection, we set the output dimension to 1024 to match the input dimension. As shown in Table~\ref{tab:ablation_shrinkadapter}, the results confirm that each component is integral to the model's performance, validating our overall structural design.

\begin{table}[hb!]
  \centering
  \small
  \setlength{\tabcolsep}{2.5pt} 
  \begin{tabular}{lccll}
    \toprule
    \textbf{Components} & \textbf{Top-1 Acc.} & \textbf{Top-5 Acc.} & \textbf{Top-1 $\Delta$} & \textbf{Top-5 $\Delta$} \\
    \midrule
    w/ ResidualAdd      & 54.05 & 83.25 & \textbf{-5.55}*** & \textbf{-4.30}*** \\
    GELU $\rightarrow$ ReLU  & 57.45 & 86.65 & \textbf{-2.15}*   & \textbf{-0.90}    \\
    w/o Dropout          & 57.00 & 86.35 & \textbf{-2.60}*  & \textbf{-1.20}*    \\
    w/o LayerNorm        & 59.25 & 87.45 & \textbf{-0.35}*    & \textbf{-0.10}    \\
    w/o Bottleneck       & 57.80 & 85.90 & \textbf{-1.80}*   & \textbf{-1.65}*   \\
    \midrule
    \textbf{ATS (Ours)} & \textbf{59.60} & \textbf{87.55} & \textbf{--} & \textbf{--} \\
    \bottomrule
  \end{tabular}
    \caption{Ablation study of the ShrinkAdapter components. The last row represents our full model (output dimension 1024), serves as the baseline for comparison. The delta ($\Delta$) columns show the performance change. Asterisks denote the significance level of the performance drop compared to the full model: * $p < 0.05$, ** $p < 0.01$, *** $p < 0.001$. All metrics are reported in percent (\%).}
  \label{tab:ablation_shrinkadapter}
\end{table}
\subsection{Architecture of STAE}

\begin{figure}[h]
  \centering
    \includegraphics[width=0.70\linewidth]{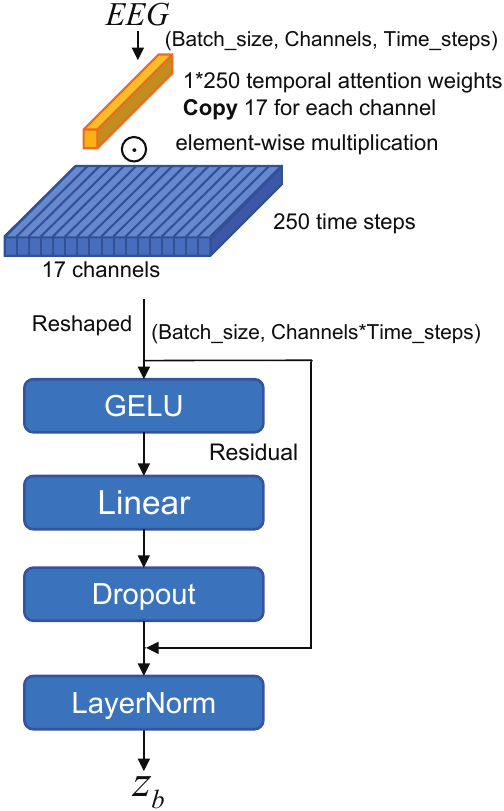}
    \caption{Architecture of the Shared Temporal Attention Encoder (STAE).}
    \label{fig:eegencoder}
\end{figure}

The Shared Temporal Attention Encoder (STAE) is an enhanced version of the simple yet efficient EEGProject from UBP. By incorporating a shared temporal attention mechanism, STAE is designed to automatically focus on time steps most relevant for visual decoding. The complete architecture is illustrated in Figure~\ref{fig:eegencoder}.

\subsection{Algorithm}
The algorithmic flow of our ATS is illustrated in Algorithm~\ref{alg:ats}.

\begin{algorithm}[hb!]
  \caption{Adaptive Teaching System}
  \label{alg:ats}
    \textbf{Input:} Multimodal training dataset $D_{train}$ \\
    \textbf{Model:} ShrinkAdapter $f_A$ with random parameters $\theta$, Brain encoder $f_B$ with random parameters $\phi$, pretrained vision encoder $f_V$ with parameters $\psi$, temperature parameter $\tau$, learning rate $\eta$ \\
    \textbf{Output:} Trained model $f_A$, $f_B$
  \begin{algorithmic}[1]
    \FOR{each iteration}
        \STATE Obtain training sample $(x_v, x_b)$ from dataset $D_{train}$
        \STATE Obtain $\tilde{x}_v$ by applying UBP prior blur settings
        \STATE $z_v \gets f_A(f_V(\tilde{x}_v))$; $z_b \gets f_B(x_b)$
        \STATE Compute loss $\mathcal{L}_{SCE}$ 
        \STATE Update $f_A$ parameters $\theta \leftarrow \theta - \eta \nabla \mathcal{L}_{SCE}$
        \STATE Update $f_B$ parameters $\phi \leftarrow \phi - \eta \nabla \mathcal{L}_{SCE}$
    \ENDFOR
    \STATE \textbf{return} trained model $f_A$, $f_B$
  \end{algorithmic}
\end{algorithm}

\begin{table*}[ht]
  \centering
  \small
  \setlength{\tabcolsep}{1.0pt} 
  \begin{tabular}{lcccccccccccccccccccccc}
  \toprule
  \multirow{2}{*}{Method} & \multicolumn{2}{c}{Subject 1} & \multicolumn{2}{c}{Subject 2} & \multicolumn{2}{c}{Subject 3} & \multicolumn{2}{c}{Subject 4} & \multicolumn{2}{c}{Subject 5} & \multicolumn{2}{c}{Subject 6} & \multicolumn{2}{c}{Subject 7} & \multicolumn{2}{c}{Subject 8} & \multicolumn{2}{c}{Subject 9} & \multicolumn{2}{c}{Subject 10} & \multicolumn{2}{c}{Avg} \\
  \cmidrule(lr){2-3} \cmidrule(lr){4-5} \cmidrule(lr){6-7} \cmidrule(lr){8-9} \cmidrule(lr){10-11} \cmidrule(lr){12-13} \cmidrule(lr){14-15} \cmidrule(lr){16-17} \cmidrule(lr){18-19} \cmidrule(lr){20-21} \cmidrule(lr){22-23}
  & top-1 & top-5 & top-1 & top-5 & top-1 & top-5 & top-1 & top-5 & top-1 & top-5 & top-1 & top-5 & top-1 & top-5 & top-1 & top-5 & top-1 & top-5 & top-1 & top-5 & top-1 & top-5 \\
  \midrule
  \multicolumn{23}{c}{\textbf{Intra-subject}: train and test on one subject} \\
  \midrule
  BraVL & 6.1 & 17.9 & 4.9 & 14.9 & 5.6 & 17.4 & 5.0 & 15.1 & 4.0 & 13.4 & 6.0 & 18.2 & 6.5 & 20.4 & 8.8 & 23.7 & 4.3 & 14.0 & 7.0 & 19.7 & 5.8 & 17.5 \\
  NICE & 13.2 & 39.5 & 13.5 & 40.3 & 14.5 & 42.7 & 20.6 & 52.7 & 10.1 & 31.5 & 16.5 & 44.0 & 17.0 & 42.1 & 22.9 & 56.1 & 15.4 & 41.6 & 17.4 & 45.8 & 16.1 & 43.6 \\
  NICE-S & 13.3 & 40.2 & 12.1 & 36.1 & 15.3 & 39.6 & 15.9 & 49.0 & 9.8 & 34.4 & 14.2 & 42.4 & 17.9 & 43.6 & 18.2 & 50.2 & 14.4 & 38.7 & 16.0 & 42.8 & 14.7 & 41.7 \\
  NICE-G & 15.2 & 40.1 & 13.9 & 40.1 & 15.7 & 42.7 & 17.6 & 48.9 & 9.0 & 29.7 & 16.3 & 44.4 & 14.9 & 43.1 & 20.3 & 52.1 & 14.1 & 39.7 & 19.6 & 46.7 & 15.6 & 42.8 \\
  MB2C & 23.6  & 56.3 & 22.6 & 50.5 & 26.3 & 60.1 & 34.8 & 67.0 & 21.3 & 53.0 & 31.0 & 62.3 & 25.0 & 54.8 & 39.0 & 69.3 & 27.5 & 59.3 & 33.1 &70.8 & 28.4 & 60.3 \\
  ATM-S & 25.6 & 60.4 & 22.0 & 54.5 & 24.0 & 62.4 & 31.4 & 60.9 & 12.9 & 43.0 & 21.4 & 51.1 & 30.5 & 61.5 & 38.8 & 72.0 & 34.4 & 51.5 & 29.1 & 63.5 & 28.5 & 60.4 \\
  CogCap & 31.4 & 79.6 & 31.4 & 77.8 & 38.1 & 85.6 & 40.3 & 85.8 & 24.4 & 66.3 & 34.8 & 78.7 & 34.6 & 80.9 & 48.1 & 88.6 & 37.4 & 79.3 & 35.5 & 79.2 & 35.6 & 80.2 \\
  VE-SDN & 32.6 & 63.7 & 34.4 & 69.9 & 38.7 & 73.5 & 39.8 & 72.0 & 29.4 & 58.6 & 34.5 & 68.8 & 34.5 & 68.3 & 49.3 & 79.8 & 39.0 & 69.6 & 39.8 & 75.3 & 37.2 & 69.9 \\
    UBP & 40.5 & 71.0 & 49.5 & 82.5 & 49.5 & 82.0 & 49.5 & 76.0 & 45.0 & 73.0 & 56.5 & 83.0 & 48.5 & 80.0 & 57.0 & 86.0 & 44.0 & 76.0 & 64.0 & 87.5 & 50.4 & 79.7 \\
  \midrule
  ATS(Ours) & \textbf{53.0} & \textbf{79.0} & \textbf{62.0} & \textbf{87.5} & \textbf{61.5} & \textbf{89.0} & \textbf{57.0} & \textbf{86.5} & \textbf{55.0} & \textbf{84.0} & \textbf{68.0} & \textbf{90.5} & \textbf{53.0} & \textbf{84.0} & \textbf{66.5} & \textbf{91.0} & \textbf{58.5} & \textbf{86.0} & \textbf{67.5} & \textbf{89.0} & \textbf{60.2} & \textbf{86.7} \\

  \midrule
  \multicolumn{23}{c}{\textbf{Inter-subject}: leave one subject out for test} \\
  \midrule
  BraVL & 2.3 & 8.0 & 1.5 & 6.3 & 1.4 & 5.9 & 1.7 & 6.7 & 1.5 & 5.6 & 1.8 & 7.2 & 2.1 & 8.1 & 2.2 & 7.6 & 1.6 & 6.4 & 2.3 & 8.5 & 1.8 & 7.0 \\
  NICE & 7.6 & 22.8 & 5.9 & 20.5 & 6.0 & 22.3 & 6.3 & 20.7 & 4.4 & 18.3 & 5.6 & 22.2 & 5.6 & 19.7 & 6.3 & 22.0 & 5.7 & 17.6 & 8.4 & 28.3 & 6.2 & 21.4 \\
  NICE-S & 7.0 & 22.6 & 6.6 & 23.2 & 7.5 & 23.7 & 5.4 & 21.4 & 6.4 & 22.2 & 7.5 & 22.5 & 3.8 & 19.1 & 8.5 & 24.4 & 7.4 & 22.3 & 9.8 & 29.6 & 7.0 & 23.1 \\
  NICE-G & 5.9 & 21.4 & 6.4 & 22.7 & 5.5 & 20.1 & 6.1 & 21.0 & 4.7 & 19.5 & 6.2 & 22.5 & 5.9 & 19.1 & 7.3 & 25.3 & 4.8 & 18.3 & 6.2 & 26.3 & 5.9 & 21.6 \\
  ATM-S & 10.5 & 26.8 & 7.1 & 24.8 & \textbf{11.9} & \textbf{33.8} & \textbf{14.7} & 39.4 & 7.0 & 23.9 & 11.1 & 35.8 & \textbf{16.1} & \textbf{43.5} & \textbf{15.0} & \textbf{40.3} & 4.9 & 22.7 & 20.5 & 46.5 & 11.8 & 33.7 \\
  UBP & 11.5 & 29.7 & 15.5 & 40.0 & 9.8 & 27.0 & 13.0 & 32.3 & \textbf{8.8} & \textbf{33.8} & 11.7 & 31.0 & 10.2 & 23.8 & 12.2 & 32.2 & \textbf{15.5} & \textbf{40.5} & 16.0 & 43.5 & 12.4 & 33.4 \\
  \midrule 
  ATS(Ours) & \textbf{17.0} & \textbf{37.5} & \textbf{21.5} & \textbf{49.5} & 7.5 & 20.0 & 13.5 & \textbf{41.5} & 7.0 & 27.0 & \textbf{14.5} & \textbf{37.0} & 10.0 & 33.5 & 13.0 & 30.5 & 9.5 & 28.5 & \textbf{26.0} & \textbf{52.5} & \textbf{14.0} & \textbf{35.8} \\

  \bottomrule
  \end{tabular}
  \caption{Top-1 and Top-5 accuracy (\%) for 200-way zero-shot retrieval on THINGS-EEG. Compared to the UBP baseline, the improvement of our ATS method is very highly significant in the intra-subject setting ($p < 0.001$), but not significant in the inter-subject setting ($p > 0.05$).}
  \label{tab:EEG_ALL}
  \end{table*}

\section{Results details}

\subsection{Full Experimental Results}

\begin{figure}[h]
  \centering
    \includegraphics[width=1.0\linewidth]{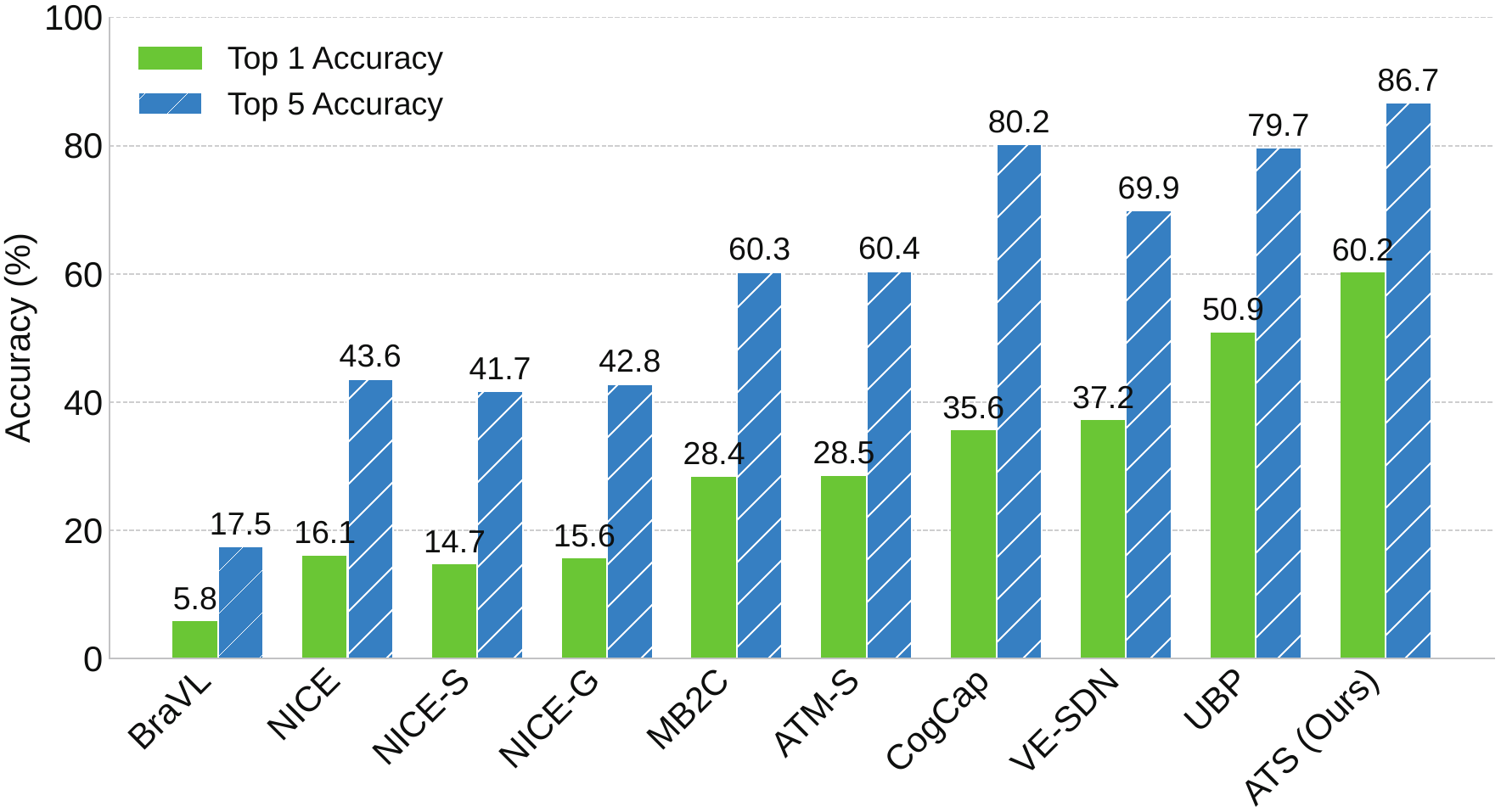}
    \caption{Zero-shot retrieval accuracy comparison on THINGS-EEG. The random chance level is 0.5\%.}
    \label{fig:zero_shot_accuracy_comparison}
\end{figure}

We evaluate our method under both intra-subject and inter-subject settings. The complete quantitative results are presented in Table~\ref{tab:EEG_ALL} and Table~\ref{tab:meg_all}, while Figure~\ref{fig:zero_shot_accuracy_comparison} illustrates the performance comparison.

The results demonstrate that our method achieves statistically significant improvements over the UBP baseline in the intra-subject setting across both THINGS-EEG and THINGS-MEG, as determined by the Wilcoxon signed-rank test. In the inter-subject setting, while our method shows an overall performance gain, its effectiveness varies across individuals. We attribute this variance primarily to the significant inter-subject variability in brain signals, a challenge our current model does not specifically address.


\begin{table}[h]
  \centering
  \small
  \setlength{\tabcolsep}{0.8pt} 
  \begin{tabular}{lcccccccccc}
    \toprule
    \multirow{2}{*}{Method}& \multicolumn{2}{c}{Subject 1} & \multicolumn{2}{c}{Subject 2} & \multicolumn{2}{c}{Subject 3} & \multicolumn{2}{c}{Subject 4}& \multicolumn{2}{c}{Avg} \\
    \cmidrule(r){2-3} \cmidrule(r){4-5} \cmidrule(r){6-7} \cmidrule(r){8-9} \cmidrule(r){10-11}
    & top-1 & top-5 & top-1 & top-5 & top-1 & top-5 & top-1 & top-5 & top-1 & top-5 \\
    \midrule
    \multicolumn{11}{c}{\textbf{Intra-subject}: train and test on one subject} \\
    \midrule
    NICE & 9.6 & 27.8 & 18.5 & 47.8 & 14.2 & 41.6 & 9.0 & 26.6 & 12.8 & 36.0\\   
    NICE-S & 9.8 & 27.8 & 18.6 & 46.4 & 10.5 & 38.4 & 11.7 & 27.2 & 12.7 & 35.0\\ 
    NICE-G & 8.7 & 30.5 & 21.8 & 56.6 & 16.5 & 49.7 & 10.3 & 32.3 & 14.3 & 42.3\\ 
    UBP & 17.0 & 38.5 & 51.0 & 81.5 & 30.0 & 63.0 & 19.5 & 44.5 & 29.4 & 56.9 \\ 
    \midrule

    \textbf{ATS} & \textbf{18.0} & \textbf{43.0} & \textbf{56.0} & \textbf{89.0} & \textbf{36.0} & \textbf{67.0} & \textbf{19.5} & \textbf{50.0} & \textbf{32.4} & \textbf{62.3} \\
    \midrule
    \multicolumn{11}{c}{\textbf{Inter-subject}: leave one subject out for test} \\
    \midrule
    UBP &2.0 & 5.7 & 1.5 & \textbf{17.2} & 2.7 & 10.5 & \textbf{2.5} & 8.0 & 2.2 & 10.4\\
    \midrule 
    \textbf{ATS} & \textbf{2.0} & \textbf{6.0} & \textbf{6.0} & 14.0 & \textbf{4.5} & \textbf{15.0} & 1.0 & \textbf{9.5} & \textbf{3.4} & \textbf{11.2}\\ 
    \bottomrule
  \end{tabular}
  \caption{Top-1 and Top-5 accuracy (\%) for 200-way zero-shot retrieval on THINGS-MEG. Compared to the UBP baseline, the improvement of our ATS method is highly significant in the intra-subject setting ($p < 0.01$), but not significant in the inter-subject setting ($p > 0.05$).}
  \label{tab:meg_all}
\end{table}

\subsection{Study on Different Vision Encoders}

To investigate the generalizability of our ShrinkAdapter, we evaluate its interaction with various EEG encoders when paired with vision encoders of different scales: the OpenCLIP RN50 and the more powerful OpenCLIP ViT-L/14. The results are presented in Table~\ref{tab:compare_RN50} and Table~\ref{tab:compare_vit-l-14}.

A clear, overarching conclusion emerges from these experiments: for capable EEG encoders like EEGProject and our STAE, applying the ShrinkAdapter yields significant performance gains across both RN50 and ViT-L/14. This robustly demonstrates the effectiveness and generalizability of our adaptive teaching paradigm.

However, a more nuanced interaction is revealed when considering weaker EEG encoders (e.g., ShallowNet, DeepNet, EEGNet). Across both vision encoder scales, applying the ShrinkAdapter to these weaker students is consistently detrimental to performance. This finding suggests a crucial precondition for our adaptive teaching paradigm: the student model must possess adequate representational capacity. We posit that weaker encoders, with their limited architectural flexibility, struggle to map noisy EEG signals to a semantically structured feature space. In this scenario, the joint optimization process leads to a \textbf{destructive co-adaptation}: to minimize the alignment loss, the ShrinkAdapter is forced to learn a ``destructive" transformation, degrading the high-quality vision features to mimic the student's poor and disorganized representations. This ultimately traps the system in a vicious cycle, harming overall performance. In contrast, stronger encoders like STAE have the potential to form a well-structured latent space. For them, the ShrinkAdapter can effectively learn to prune the teacher's features in a meaningful way, simplifying the alignment task and leading to a \textbf{virtuous cycle} of co-adaptation.
\begin{table*}[htbp]
\centering
\small
\setlength{\tabcolsep}{4.0pt} 
\begin{tabular}{lccccccccccc}
    \toprule
    EEG         & \multicolumn{8}{c}{CLIP Models} & \multicolumn{3}{c}{DINOv2 Models} \\
    \cmidrule(lr){2-9} \cmidrule(lr){10-12}
    Top-1 ACC   & RN50 & RN101 & ViT-B-16 & ViT-B-32 & ViT-L-14 & ViT-H-14 & ViT-g-14 & ViT-bigG-14 & ViT-B-14 & ViT-L-14 & ViT-G-14 \\
    \midrule
    UBP         & 50.4 & 46.3  & 44.5     & 48.3     & 44.2     & 44.3     & 40.8     & 43.8        & 26.3     & 19.7     & 20.3     \\
    ATS         & 59.6 & 51.1  & 50.7     & 55.4     & 50.2     & 49.5     & 44.3     & 47.6        & 31.2     & 22.7     & 23.5     \\
    \midrule
    ratio(best) & 0.25 & 1     & 1        & 0.5      & 1        & 0.25     & 0.5      & 0.25        & 0.5      & 1        & 0.25     \\     
    \bottomrule
  \end{tabular}
  \caption{Top-1 accuracy (\%) for 200-way zero-shot retrieval on the THINGS-EEG dataset across various vision encoders used as teachers. ATS consistently outperforms the UBP baseline across all teacher models, demonstrating its robustness and effectiveness. The "ratio(best)" row indicates the optimal bottleneck compression ratio for the ShrinkAdapter, where the latent dimension is kept consistent with that of the original teacher model.}
  \label{tab:cross_teacher_EEG}
\end{table*}

\begin{table*}[htbp]
\centering
\small
\setlength{\tabcolsep}{4.0pt} 
\begin{tabular}{lccccccccccc}
    \toprule
    MEG         & \multicolumn{8}{c}{CLIP Models} & \multicolumn{3}{c}{DINOv2 Models} \\
    \cmidrule(lr){2-9} \cmidrule(lr){10-12}
    Top-1 ACC   & RN50 & RN101 & ViT-B-16 & ViT-B-32 & ViT-L-14 & ViT-H-14 & ViT-g-14 & ViT-bigG-14 & ViT-B-14 & ViT-L-14 & ViT-G-14 \\
    \midrule
    UBP         & 20.6 & 19.0  & 23.1     & 24.1     & 25.1     & 28.3     & 22.0     & 27.6        & 13.5     & 10.1     & 10.5     \\
    ATS         & 28.6 & 25.4  & 29.6     & 26.4     & 26.3     & 27.8     & 21.9     & 26.0        & 15.6     & 11.4     & 11.8     \\
    \midrule
    ratio(best) & 0.25 & 0.25     & 0.25        & 0.25      & 0.25        & 0.25     & 0.25     & 0.25        & 1     & 1        & 0.25     \\     
    \bottomrule
  \end{tabular}
  \caption{Top-1 accuracy (\%) for 200-way zero-shot retrieval on the THINGS-MEG dataset across various vision encoders used as teachers. }
  \label{tab:cross_teacher_MEG}
\end{table*}

As a further supplement, we tested the robustness of ATS across an even wider range of teacher models (vision encoders) on THINGS-EEG. We selected 10 encoders from two distinct families: the semantically rich CLIP models and the self-supervised DINOv2 models. The results are presented in Table~\ref{tab:cross_teacher_EEG} for THINGS-EEG.

This analysis provides two key insights. First, ATS robustly outperforms the UBP baseline across diverse teacher sizes (e.g., ViT-B vs. ViT-L) and types (CLIP vs. DINOv2), confirming that our method is not overfitting to the properties of any single teacher. Second, the performance trends validate our ``Asymmetric Gap" hypothesis: overly complex teachers exacerbate the modality asymmetry. Detailed experimental results are presented in Tables 9--19.

The experimental results on the THINGS-MEG dataset are presented in Table~\ref{tab:cross_teacher_MEG}. Overall, ATS remains effective, outperforming the baseline on most models. However, unlike the trend observed on EEG, performance on MEG tends to improve with more complex teacher models. This may be attributed to the intrinsic properties of the MEG data, such as reduced temporal aliasing due to its distinct experimental settings. This intriguing observation warrants further investigation. Detailed experimental results are presented in Tables 20--30.

Collectively, these studies confirm that ATS is a generalizable framework, while also revealing the critical importance of sufficient student capacity for adaptive teaching to be effective. 

\subsection{Retrieval Case Analysis}
To provide an intuitive understanding of our model's decoding behavior, we present a qualitative analysis of Top-5 retrieval cases from the THINGS-EEG dataset (subject 10). This case study reveals a critical insight: the features decoded from EEG are not purely abstract but are hybrid representations, jointly encoding both high-level semantic concepts and low-level visual properties like \textbf{color, texture, spatial layout, orientation, and quantity}.

\subsubsection{Good Cases}
\begin{figure}[htbp]
  \centering
    \includegraphics[width=0.89\linewidth]{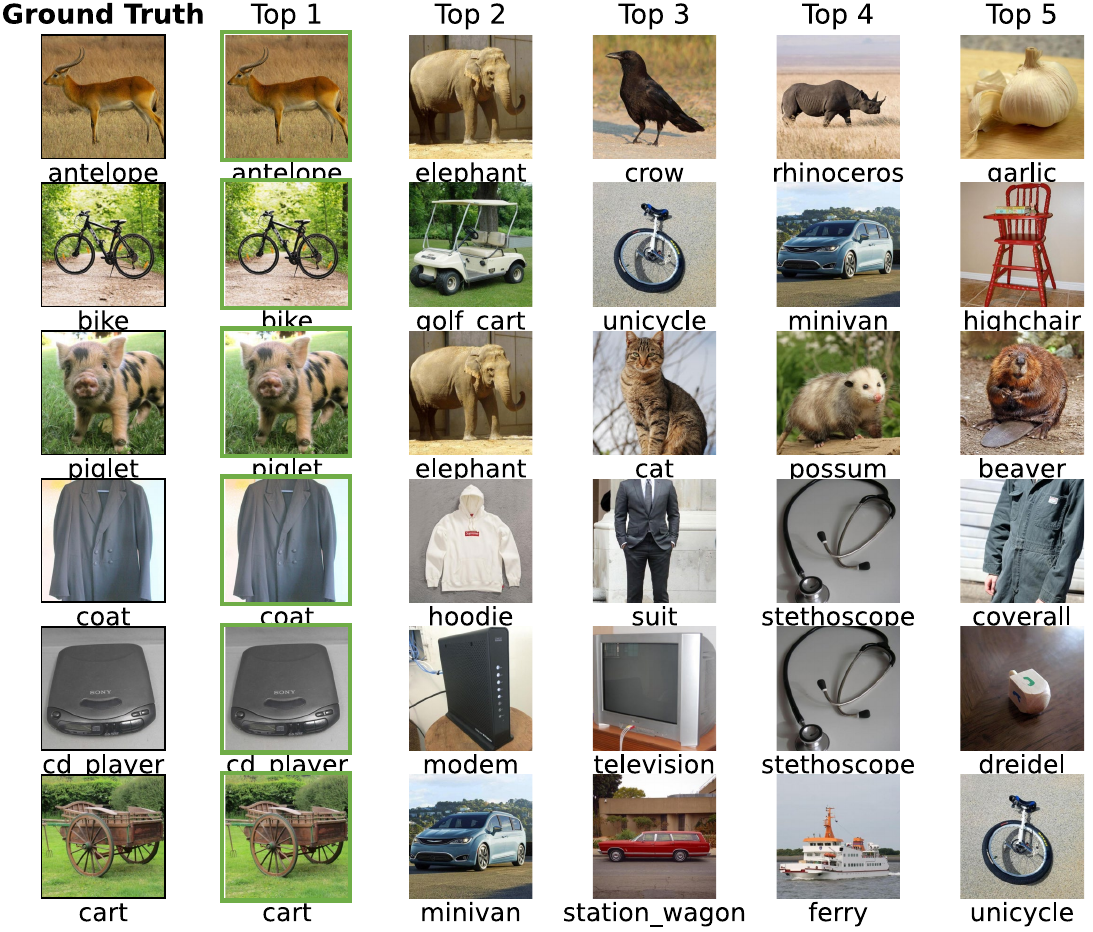} 
    \caption{Good Cases: Top-5 retrieval results for various stimuli (subject 10). }
    \label{fig:good_cases}
\end{figure}

Figure~\ref{fig:good_cases} presents successful retrieval examples. In these cases, the model not only retrieves the correct target at Top-1 but also shows strong semantic consistency within the Top-5 results (e.g., retrieving other animals for an animal stimulus). This suggests that the decoded EEG features contain category-level semantic information.

\subsubsection{Perceptual-driven Cases}
\begin{figure}[htbp]
  \centering
    \includegraphics[width=0.89\linewidth]{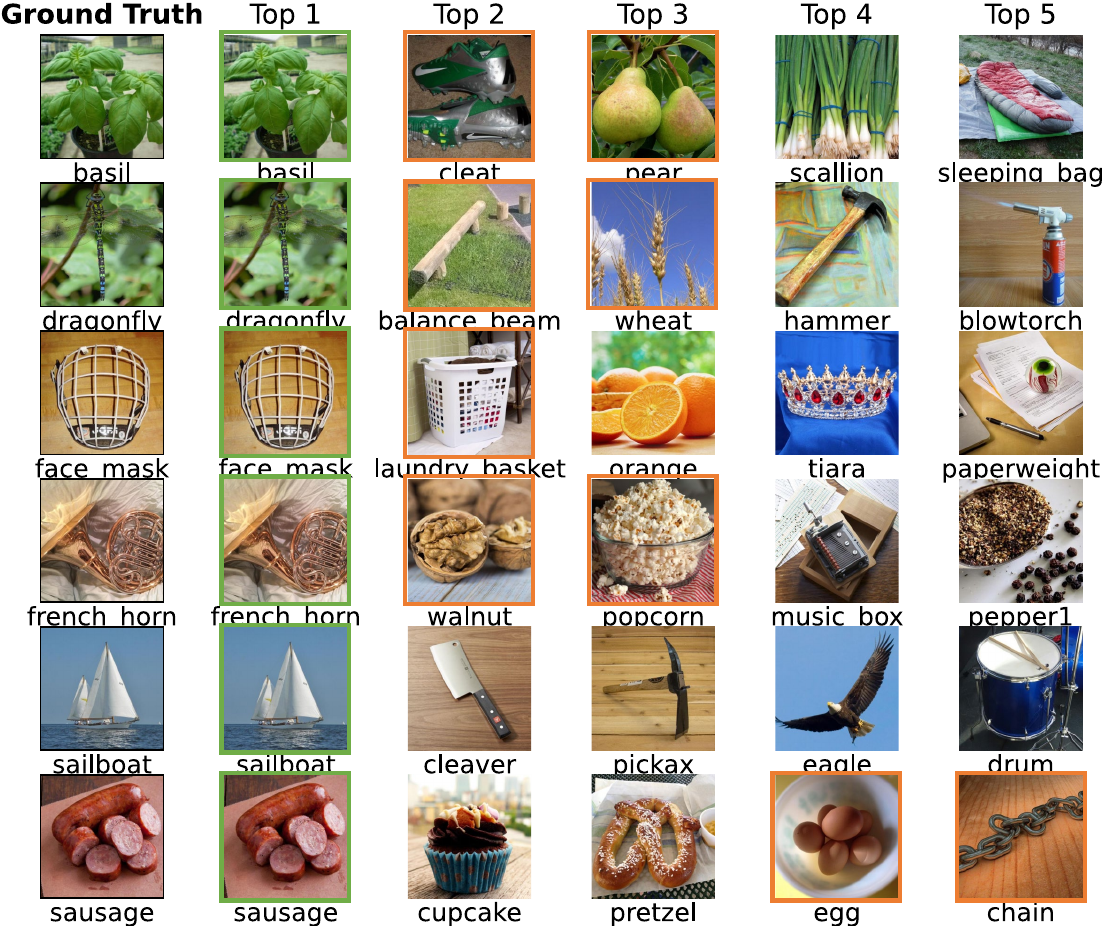}
    \caption{Perceptual-driven Cases: Top-5 retrieval results for various stimuli (subject 10). }
    \label{fig:perceptual_driven_cases}
\end{figure}

As illustrated in Figure~\ref{fig:perceptual_driven_cases}, we observe more revealing cases where the model prioritizes perceptual similarity over semantic correctness. These instances suggest the hybrid nature of the decoded features. For example:
\begin{itemize}
    \item \textbf{Color (Row 1):} When presented with the image of \textit{basil}, which is predominantly green, the model retrieves a \textit{cleat}  and a \textit{pear} in its Top-3. These items are semantically unrelated but share the same dominant green color as the target.
    \item \textbf{Texture and Pattern (Row 2, 3, 4):} The model's focus on perceptual patterns is evident across multiple cases. For the stimulus of a \textit{dragonfly}, it retrieves a \textit{balance beam} and \textit{wheat} in its Top-3, likely driven by their shared elongated shape, orientation, and fine-grained texture. Similarly, for the \textit{face mask} with its distinct grid-like structure, the Top-2 retrieval is a \textit{laundry basket}, focusing on the shared pattern while ignoring the semantic category. Furthermore, for the \textit{french horn}, the model retrieves a \textit{walnut} and \textit{popcorn}, which share a complex, irregular texture and a similar color palette with the target.

    \item \textbf{Shape and Quantity (Row 6):} For the stimulus of sliced \textit{sausage}, the model retrieves \textit{eggs} (Top-4) and a \textit{chain} (Top-5). While semantically distinct, both results share core visual properties with the target: the presence of multiple, repeated oval-shaped elements. This indicates the model is sensitive to both the shape of individual components and their numerosity.
\end{itemize}
In these cases, retrieved images may be semantically unrelated to the target but share a salient low-level feature.

\subsubsection{Bad Cases}
\begin{figure}[htbp]
  \centering
    \includegraphics[width=0.89\linewidth]{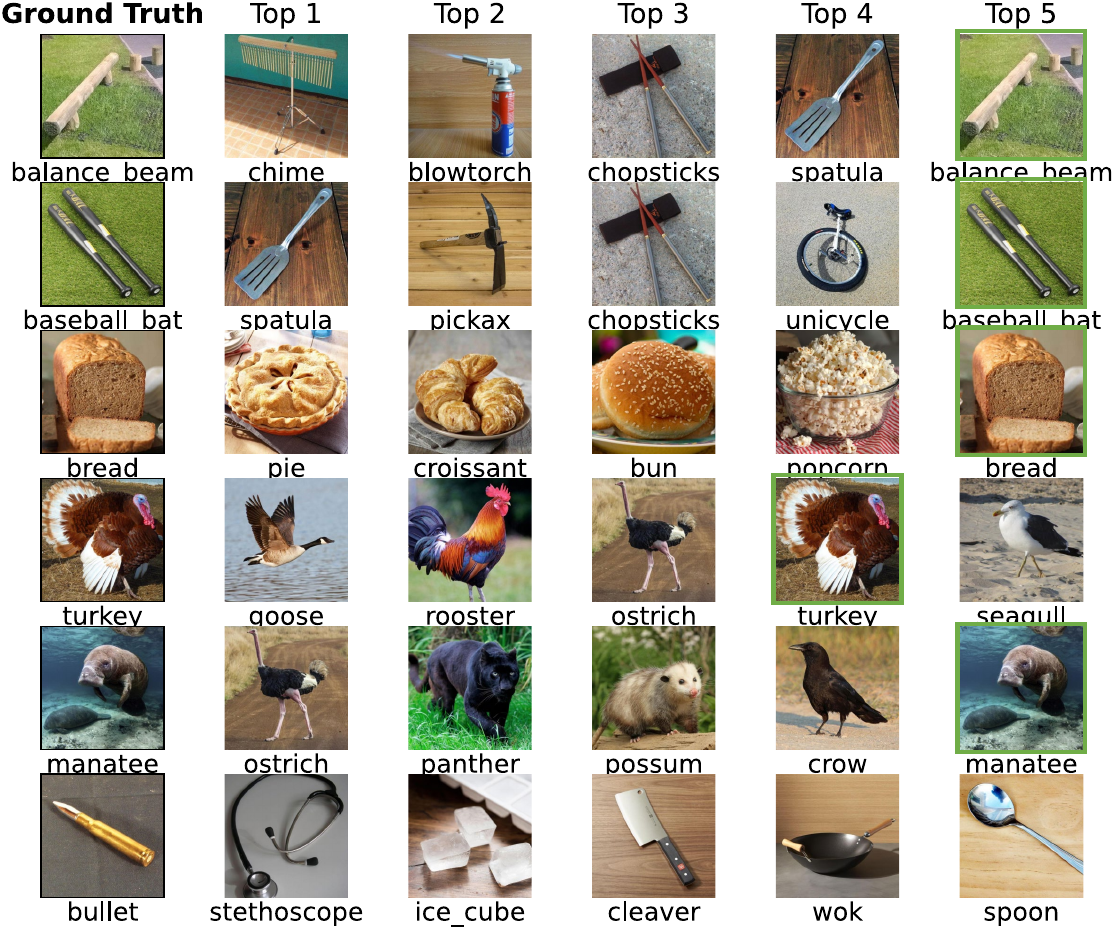}
    \caption{Bad Cases: Top-5 retrieval results for various stimuli (subject 10). }
    \label{fig:bad_cases}
\end{figure}

Examples of failed retrievals are presented in Figure~\ref{fig:bad_cases}. These typically involve confusion between semantically adjacent or visually similar categories, illustrating the current limitations of the decoding process.

Collectively, these results suggest that our method decodes a rich, hybrid feature space from EEG signals.

\subsection{More Evaluation Metrics}

To comprehensively evaluate model performance, we employ three key metrics.

\paragraph{Top-K Accuracy.}
This is the standard metric for retrieval tasks. It measures whether the ground-truth item appears within the top $K$ ranked results for a given query $q$. The accuracy for a single query is defined as:
\begin{equation}
\text{Top-}K(q) =
\begin{cases}
1, & \text{if } \text{rank}(q) \leq K, \\
0, & \text{otherwise},
\end{cases}
\end{equation}
where $\text{rank}(q)$ is the rank of the ground-truth item. The overall Top-K Accuracy is the average over all queries in the test set.

\paragraph{Mean Average Precision (mAP).}
In line with UBP, we also adopt mAP for a more granular assessment of ranking quality. Since each query in our task has a single correct item, the Average Precision (AP) for a query $q$ simplifies to the reciprocal of the ground-truth item's rank:
\begin{equation}
\text{AP}(q) = \frac{1}{\text{rank}(q)}
\end{equation}
Unlike Top-K accuracy, which treats all ranks within K equally, mAP rewards higher rankings more strongly and penalizes lower rankings across the entire list. It thus offers a more holistic measure of retrieval performance. The final mAP is the mean of AP values across all queries.

\paragraph{Similarity.}
While the above metrics focus on ranking, the Similarity score directly measures the alignment quality in the embedding space. It is defined as the mean of the diagonal elements in the retrieval similarity matrix. This score quantifies how closely the model maps ground-truth pairs, independent of their ranking against other non-matching items.

The comprehensive results presented in Table~\ref{tab:metrics} demonstrate that our proposed ATS framework achieves superior performance across all three metrics.
\begin{table}[h]
\centering
\small
\setlength{\tabcolsep}{1mm}
\begin{tabular}{lllll}
\toprule
\textbf{Method} & \textbf{Top-1 Acc} & \textbf{Top-5 Acc} & \textbf{mAP*} & \textbf{Similarity*} \\
\midrule
UBP & 50.9 & 79.7 & 63.8 & 0.199 \\
\textbf{ATS (Ours)} & \textbf{60.2} {\scriptsize (+9.3)} & \textbf{86.7} {\scriptsize (+7.0)} & \textbf{72.0} {\scriptsize (+8.2)} & \textbf{0.515} {\scriptsize (+0.316)} \\
\bottomrule
\end{tabular}
\caption{Retrieval performance on the THINGS-EEG dataset. Accuracy and mAP metrics are reported in percent (\%).}
\label{tab:metrics}
\end{table}

\begin{table*}[t!]
  \centering
  \small
  \setlength{\tabcolsep}{0.8pt} 
  
  \begin{tabular}{lcccccccccccccccccccccc}
    \toprule
    \multirow{2}{*}{Backbone}& \multicolumn{2}{c}{Subject 1} & \multicolumn{2}{c}{Subject 2} & \multicolumn{2}{c}{Subject 3} & \multicolumn{2}{c}{Subject 4}  & \multicolumn{2}{c}{Subject 5}  & \multicolumn{2}{c}{Subject 6}  & \multicolumn{2}{c}{Subject 7}  & \multicolumn{2}{c}{Subject 8} & \multicolumn{2}{c}{Subject 9} & \multicolumn{2}{c}{Subject 10} & \multicolumn{2}{c}{Avg} \\
    \cmidrule(r){2-3} \cmidrule(r){4-5} \cmidrule(r){6-7} \cmidrule(r){8-9} \cmidrule(r){10-11} \cmidrule(r){12-13} \cmidrule(r){14-15} \cmidrule(r){16-17} \cmidrule(r){18-19} \cmidrule(r){20-21} \cmidrule(r){22-23}
    & top-1 & top-5 & top-1 & top-5 & top-1 & top-5 & top-1 & top-5 & top-1 & top-5 & top-1 & top-5 & top-1 & top-5 & top-1 & top-5 & top-1 & top-5 & top-1 & top-5 & top-1 & top-5\\
    \midrule
    ShallowNet & 24.5 & 56.5 & 26.5 & 63.5 & 33 & 67.5 & 37 & 68.5 & 21.0 & 48 & 39.5 & 71.0 & 37.0 & 66.5 & 43 & 77.5 & 28.0 & 63.5 & 42.0 & 72.5 & 33.15 & 65.5 \\
    w/ ShrinkA & 28.0 & 60.5 & 27.5 & 58.0 & 30.5 & 69.5 & 36.0 & 68.0 & 17.0 & 45.5 & 38.0 & 70 & 34.5 & 65.5 & 43.5 & 76.5 & 27.5 & 60.5 & 39.5 & 77.0 & 32.2 & 65.1 \\
    \midrule
    DeepNet & 9.0 & 32.5 & 11.5 & 37.0 & 10.0 & 40.5 & 16.5 & 45.5 & 5.5 & 23.0 & 16.5 & 40.0 & 15.5 & 37.0 & 14.5 & 41.5 & 16.5 & 41.5 & 16.5 & 46.0 & 13.2 & 38.45 \\
    w/ ShrinkA & 3.0 & 11.0 & 2.0 & 10.5 & 2.5 & 12.0 & 5.0 & 23.5 & 1.0 & 6.5 & 1.0 & 6.5 & 4.5 & 13.5 & 2.5 & 13.0 & 2.0 & 12.0 & 6.5 & 24.5 & 3.0 & 13.3 \\
    \midrule
    EEGNet & 21.0 & 46.5 & 25.5 & 59.0 & 26.0 & 59.5 & 34.5 & 64.0 & 20.0 & 49.5 & 31.5 & 66.5 & 24.5 & 56.0 & 37.0 & 71.0 & 26.5 & 57.5 & 32.5 & 69.5 & 27.9 & 59.9 \\
    w/ ShrinkA & 11.5 & 37.0 & 24.5 & 54.5 & 23.0 & 52.5 & 30.5 & 60.0 & 11.0 & 38.0 & 31.0 & 64.5 & 23.0 & 46.5 & 29.5 & 65.5 & 20.5 & 52.0 & 34.0 & 73.5 & 23.85 & 54.4 \\
    \midrule
    TSConv & 39.0 & 74.0 & 31.5 & 72.0 & 42.5 & 76.0 & 44.5 & 76.5 & 30.0 & 63.0 & 44.0 & 75.5 & 37.5 & 70.0 & 50.0 & 81.5 & 33.5 & 70.5 & 47.0 & 80.5 & 39.95 & 73.95 \\
    w/ ShrinkA & 36.0 & 70.5 & 37.5 & 74.5 & 44.0 & 79.5 & 50.5 & 76.0 & 34.5 & 67.0 & 55.5 & 83.5 & 40.5 & 70.5 & 55.0 & 87.5 & 39.5 & 73.5 & 61.0 & 84.5 & 45.4 & 76.7 \\
    \midrule
    EEGProject & 40.5 & 71.0 & 49.5 & 82.5 & 49.5 & 82.0 & 49.5 & 76.0 & 45.0 & 73.0 & 56.0 & 83.0 & 48.5 & 80.0 & 57.0 & 86.0 & 44.0 & 76.0 & 64.0 & 87.5 & 50.35 & 79.7 \\
    w/ ShrinkA & 43.0 & 76.5 & 57.0 & 85.5 & 60.0 & 88.0 & 54.5 & 84.0 & 50.5 & 82.0 & 65.5 & 88.5 & 52.5 & 85.0 & 67.5 & 89.0 & 46.5 & 85.0 & 70.0 & 91.0 & 56.7 & 85.45 \\
    \midrule
    STAE (Ours) & 43.0 & 71.5 & 52.0 & 82.0 & 49.0 & 82.0 & 49.5 & 76.0 & 45.5 & 75.5 & 58.5 & 82.5 & 50.0 & 77.5 & 59.5 & 87.5 & 47.5 & 76.5 & 57.5 & 86.5 & 51.2 & 79.75 \\
    w/ ShrinkA & 50.5 & 78.0 & 59.5 & 89.0 & 60.5 & 89.0 & 59.5 & 88.0 & 58.5 & 86.0 & 64.5 & 91.5 & 55.0 & 86.0 & 67.5 & 91.0 & 55.0 & 88.5 & 65.5 & 88.5 & 59.6 & 87.55 \\
    \bottomrule
  \end{tabular}
  \caption{Performance comparison on THINGS-EEG using OpenCLIP RN50 as the vision encoder (latent dimension: 1024). For each backbone, we report Top-1 and Top-5 accuracy (\%) with and without our ShrinkAdapter. Key findings are: (1) Applying the ShrinkAdapter yields significant improvements for both EEGProject and our STAE ($p < 0.001$). (2) With the ShrinkAdapter, our STAE significantly outperforms EEGProject on both Top-1 ($p < 0.05$) and Top-5 ($p < 0.01$) accuracy.}
  \label{tab:compare_RN50}
\end{table*}

\begin{table*}[t!]
  \centering
  \small
  \setlength{\tabcolsep}{0.8pt} 
  
  \begin{tabular}{lcccccccccccccccccccccc}
    \toprule
    \multirow{2}{*}{Backbone}& \multicolumn{2}{c}{Subject 1} & \multicolumn{2}{c}{Subject 2} & \multicolumn{2}{c}{Subject 3} & \multicolumn{2}{c}{Subject 4}  & \multicolumn{2}{c}{Subject 5}  & \multicolumn{2}{c}{Subject 6}  & \multicolumn{2}{c}{Subject 7}  & \multicolumn{2}{c}{Subject 8} & \multicolumn{2}{c}{Subject 9} & \multicolumn{2}{c}{Subject 10} & \multicolumn{2}{c}{Avg} \\
    \cmidrule(r){2-3} \cmidrule(r){4-5} \cmidrule(r){6-7} \cmidrule(r){8-9} \cmidrule(r){10-11} \cmidrule(r){12-13} \cmidrule(r){14-15} \cmidrule(r){16-17} \cmidrule(r){18-19} \cmidrule(r){20-21} \cmidrule(r){22-23}
    & top-1 & top-5 & top-1 & top-5 & top-1 & top-5 & top-1 & top-5 & top-1 & top-5 & top-1 & top-5 & top-1 & top-5 & top-1 & top-5 & top-1 & top-5 & top-1 & top-5 & top-1 & top-5\\
    \midrule
    ShallowNet  & 18.0 & 51.0 & 23.0 & 54.5 & 26.5 & 60.0 & 31.0 & 66.0 & 18.0 & 51.0 & 33.5 & 71.5 & 26.5 & 61.5 & 38.0 & 72.5 & 22.0 & 59.5 & 34.5 & 70.5 & 27.1 & 61.8 \\
    w/ ShrinkA  & 22.0 & 52.5 & 25.5 & 57.0 & 39.5 & 70.0 & 42.5 & 72.5 & 28.0 & 57.0 & 42.0 & 73.0 & 32.0 & 67.5 & 45.5 & 76.0 & 32.0 & 65.0 & 41.5 & 74.5 & 35.1 & 66.5 \\ \midrule
    DeepNet     & 10.5 & 32.0 & 12.0 & 35.0 & 15.0 & 35.0 & 17.0 & 44.5 & 6.5  & 24.5 & 18.5 & 38.5 & 13.0 & 40.0 & 15.0 & 40.0 & 16.0 & 39.5 & 14.0 & 43.5 & 13.8 & 37.3 \\
    w/ ShrinkA  & 3.5  & 18.0 & 6.5  & 22.0 & 7.0  & 28.0 & 19.0 & 45.5 & 3.5  & 19.0 & 9.0  & 27.0 & 6.0  & 22.5 & 8.5  & 22.0 & 9.0  & 24.5 & 15.5 & 42.0 & 8.8  & 27.1 \\ \midrule
    EEGNet      & 17.0 & 46.0 & 24.0 & 51.5 & 24.0 & 56.0 & 32.0 & 63.0 & 21.0 & 48.0 & 30.0 & 64.0 & 24.0 & 57.0 & 27.0 & 63.0 & 22.0 & 55.0 & 28.5 & 66.0 & 25.0 & 57.0 \\
    w/ ShrinkA  & 20.5 & 48.5 & 31.0 & 60.0 & 28.5 & 59.0 & 37.5 & 66.5 & 22.0 & 54.5 & 33.5 & 69.5 & 35.5 & 61.0 & 36.0 & 70.0 & 27.5 & 61.0 & 46.5 & 80.0 & 31.9 & 63.0 \\ \midrule
    TSConv      & 25.5 & 64.0 & 26.0 & 65.0 & 33.0 & 68.0 & 37.0 & 71.5 & 29.0 & 56.0 & 38.5 & 79.0 & 32.0 & 67.0 & 46.5 & 79.0 & 30.5 & 64.0 & 42.5 & 74.5 & 34.1 & 68.8 \\
    w/ ShrinkA  & 30.0 & 61.0 & 29.5 & 66.5 & 40.5 & 69.5 & 42.5 & 77.0 & 30.5 & 57.0 & 46.5 & 79.0 & 36.0 & 69.0 & 47.0 & 80.5 & 37.0 & 68.0 & 49.0 & 79.0 & 38.9 & 70.7 \\ \midrule
    EEGProject  & 30.0 & 66.5 & 41.5 & 74.0 & 44.0 & 79.5 & 46.0 & 74.0 & 32.0 & 69.0 & 54.0 & 83.0 & 39.0 & 74.5 & 58.0 & 81.0 & 43.5 & 74.5 & 54.0 & 84.0 & 44.2 & 76.0 \\
    w/ ShrinkA  & 33.5 & 66.5 & 43.0 & 73.5 & 44.5 & 80.5 & 48.0 & 79.5 & 39.5 & 70.5 & 53.0 & 83.5 & 44.0 & 78.0 & 59.5 & 82.0 & 44.5 & 74.5 & 57.0 & 87.0 & 46.7 & 77.6 \\ \midrule
    STAE (Ours) & 32.0 & 68.0 & 44.0 & 76.0 & 44.5 & 81.5 & 46.0 & 75.0 & 35.0 & 75.0 & 53.5 & 83.0 & 42.0 & 76.0 & 58.5 & 82.5 & 44.5 & 75.5 & 54.0 & 85.5 & 45.4 & 77.8 \\
    w/ ShrinkA  & 42.0 & 69.0 & 48.0 & 79.0 & 46.5 & 81.0 & 52.5 & 83.0 & 42.0 & 75.5 & 53.0 & 87.0 & 46.0 & 80.5 & 61.0 & 82.5 & 51.0 & 77.0 & 60.0 & 87.0 & 50.2 & 80.2 \\
    \bottomrule
  \end{tabular}
  \caption{Performance comparison on THINGS-EEG using OpenCLIP ViT-L/14 as the vision encoder  (latent dimension: 768). For each backbone, we report Top-1 and Top-5 accuracy (\%) with and without our ShrinkAdapter. Key findings are: (1) Applying the ShrinkAdapter yields significant improvements for both EEGProject and our STAE ($p < 0.01$). (2) With the ShrinkAdapter, our STAE significantly outperforms EEGProject on Top-5 accuracy ($p < 0.01$), while the difference on Top-1 is not significant ($p > 0.05$).}
  \label{tab:compare_vit-l-14}
\end{table*}

\begin{table*}[t!]
  \centering
  \small
  \setlength{\tabcolsep}{0.8pt} 
  
  \begin{tabular}{lcccccccccccccccccccccc}
    \toprule
    \multirow{2}{*}{Backbone}& \multicolumn{2}{c}{Subject 1} & \multicolumn{2}{c}{Subject 2} & \multicolumn{2}{c}{Subject 3} & \multicolumn{2}{c}{Subject 4}  & \multicolumn{2}{c}{Subject 5}  & \multicolumn{2}{c}{Subject 6}  & \multicolumn{2}{c}{Subject 7}  & \multicolumn{2}{c}{Subject 8} & \multicolumn{2}{c}{Subject 9} & \multicolumn{2}{c}{Subject 10} & \multicolumn{2}{c}{Avg} \\
    \cmidrule(r){2-3} \cmidrule(r){4-5} \cmidrule(r){6-7} \cmidrule(r){8-9} \cmidrule(r){10-11} \cmidrule(r){12-13} \cmidrule(r){14-15} \cmidrule(r){16-17} \cmidrule(r){18-19} \cmidrule(r){20-21} \cmidrule(r){22-23}
    & top-1 & top-5 & top-1 & top-5 & top-1 & top-5 & top-1 & top-5 & top-1 & top-5 & top-1 & top-5 & top-1 & top-5 & top-1 & top-5 & top-1 & top-5 & top-1 & top-5 & top-1 & top-5\\
    \midrule
    ShallowNet  & 22.0 & 56.0 & 20.0 & 60.0 & 29.0 & 61.0 & 31.5 & 64.5 & 16.0 & 46.0 & 33.0 & 72.5 & 23.5 & 62.0 & 34.5 & 69.5 & 22.0 & 63.0 & 35.5 & 72.5 & 26.7 & 62.7 \\
    w/ ShrinkA  & 25.5 & 53.0 & 27.5 & 60.0 & 35.5 & 68.0 & 34.5 & 73.0 & 22.0 & 51.0 & 34.0 & 70.0 & 31.5 & 63.5 & 39.0 & 73.5 & 33.5 & 62.0 & 37.5 & 78.0 & 32.1 & 65.2 \\ \midrule
    DeepNet     & 8.5  & 27.5 & 11.0 & 34.5 & 9.5  & 32.0 & 15.5 & 42.0 & 7.5  & 22.5 & 13.5 & 36.5 & 8.5  & 30.5 & 12.0 & 34.5 & 15.5 & 40.0 & 16.5 & 49.0 & 11.8 & 34.9 \\
    w/ ShrinkA  & 2.5  & 10.5 & 3.0  & 7.0  & 1.5  & 9.0  & 8.0  & 24.0 & 1.0  & 7.5  & 1.5  & 7.5  & 5.0  & 15.5 & 4.0  & 12.5 & 1.5  & 7.0  & 7.5  & 29.5 & 3.6  & 13.0 \\ \midrule
    EEGNet      & 20.5 & 44.0 & 15.5 & 55.0 & 21.5 & 52.0 & 23.0 & 57.5 & 18.0 & 47.0 & 29.0 & 59.5 & 18.0 & 42.5 & 19.5 & 61.5 & 25.0 & 54.5 & 29.0 & 62.0 & 21.9 & 53.6 \\
    w/ ShrinkA  & 17.0 & 37.5 & 24.5 & 56.0 & 25.5 & 58.5 & 27.5 & 63.0 & 14.5 & 40.5 & 37.5 & 65.5 & 18.5 & 45.5 & 30.0 & 60.5 & 26.5 & 56.5 & 32.0 & 69.0 & 25.4 & 55.3 \\ \midrule
    TSConv      & 34.0 & 67.5 & 25.5 & 64.0 & 32.5 & 67.5 & 36.0 & 72.5 & 25.5 & 57.5 & 40.5 & 78.0 & 26.5 & 64.5 & 43.0 & 79.5 & 33.0 & 72.0 & 40.5 & 74.5 & 33.7 & 69.8 \\
    w/ ShrinkA  & 36.0 & 68.5 & 34.0 & 70.5 & 44.5 & 75.5 & 41.5 & 75.0 & 31.5 & 61.0 & 41.5 & 83.0 & 34.5 & 66.5 & 47.0 & 80.5 & 41.0 & 67.5 & 49.0 & 83.0 & 40.1 & 73.1 \\ \midrule
    EEGProject  & 36.5 & 71.5 & 47.0 & 81.0 & 44.5 & 78.5 & 47.0 & 80.0 & 42.5 & 71.0 & 52.5 & 84.0 & 45.0 & 78.0 & 54.5 & 81.0 & 40.5 & 74.0 & 52.5 & 87.0 & 46.3 & 78.6 \\
    w/ ShrinkA  & 39.5 & 70.0 & 49.5 & 79.0 & 51.5 & 84.5 & 47.0 & 86.5 & 39.5 & 72.5 & 51.5 & 85.0 & 44.5 & 77.0 & 56.5 & 86.0 & 46.5 & 80.0 & 57.5 & 91.0 & 48.4 & 81.2 \\ \midrule
    STAE (Ours) & 37.0 & 71.5 & 44.0 & 83.0 & 46.0 & 78.0 & 45.0 & 79.0 & 41.0 & 72.0 & 51.5 & 84.0 & 37.5 & 73.0 & 56.5 & 84.5 & 43.5 & 76.0 & 49.5 & 85.0 & 45.2 & 78.6 \\
    w/ ShrinkA  & 43.5 & 74.0 & 48.5 & 85.0 & 55.5 & 85.5 & 51.0 & 88.0 & 45.5 & 77.5 & 55.0 & 89.0 & 45.5 & 79.0 & 59.0 & 88.5 & 49.0 & 85.0 & 58.5 & 89.0 & 51.1 & 84.1 \\
    \bottomrule
  \end{tabular}
  \caption{Performance comparison on THINGS-EEG using OpenCLIP RN101 as the vision encoder (latent dimension: 512). }
  \label{tab:compare_RN101}
\end{table*}

\begin{table*}[t!]
  \centering
  \small
  \setlength{\tabcolsep}{0.8pt} 
  
  \begin{tabular}{lcccccccccccccccccccccc}
    \toprule
    \multirow{2}{*}{Backbone}& \multicolumn{2}{c}{Subject 1} & \multicolumn{2}{c}{Subject 2} & \multicolumn{2}{c}{Subject 3} & \multicolumn{2}{c}{Subject 4}  & \multicolumn{2}{c}{Subject 5}  & \multicolumn{2}{c}{Subject 6}  & \multicolumn{2}{c}{Subject 7}  & \multicolumn{2}{c}{Subject 8} & \multicolumn{2}{c}{Subject 9} & \multicolumn{2}{c}{Subject 10} & \multicolumn{2}{c}{Avg} \\
    \cmidrule(r){2-3} \cmidrule(r){4-5} \cmidrule(r){6-7} \cmidrule(r){8-9} \cmidrule(r){10-11} \cmidrule(r){12-13} \cmidrule(r){14-15} \cmidrule(r){16-17} \cmidrule(r){18-19} \cmidrule(r){20-21} \cmidrule(r){22-23}
    & top-1 & top-5 & top-1 & top-5 & top-1 & top-5 & top-1 & top-5 & top-1 & top-5 & top-1 & top-5 & top-1 & top-5 & top-1 & top-5 & top-1 & top-5 & top-1 & top-5 & top-1 & top-5\\
    \midrule
    ShallowNet  & 21.5 & 53.0 & 23.5 & 54.5 & 24.5 & 62.5 & 35.0 & 67.5 & 20.5 & 44.0 & 29.5 & 65.0 & 29.0 & 62.5 & 34.5 & 72.0 & 25.5 & 56.5 & 37.0 & 69.0 & 28.1 & 60.7 \\
    w/ ShrinkA  & 23.0 & 53.0 & 31.5 & 58.5 & 34.0 & 67.5 & 35.0 & 69.5 & 21.5 & 56.0 & 38.5 & 69.5 & 34.0 & 61.0 & 39.0 & 75.5 & 27.5 & 61.5 & 40.0 & 72.0 & 32.4 & 64.4 \\ \midrule
    DeepNet     & 11.0 & 26.0 & 8.0  & 32.5 & 9.5  & 32.5 & 21.0 & 47.5 & 8.5  & 25.5 & 11.5 & 33.5 & 9.5  & 30.5 & 10.5 & 34.0 & 15.0 & 42.0 & 10.5 & 40.0 & 11.5 & 34.4 \\
    w/ ShrinkA  & 3.5  & 13.5 & 5.5  & 18.5 & 5.5  & 23.5 & 14.5 & 32.0 & 3.0  & 11.0 & 7.5  & 21.0 & 5.0  & 17.0 & 7.0  & 18.5 & 4.0  & 15.0 & 9.5  & 31.0 & 6.5  & 20.1 \\ \midrule
    EEGNet      & 18.5 & 44.0 & 19.0 & 50.0 & 26.0 & 53.5 & 28.0 & 62.0 & 18.0 & 43.5 & 27.5 & 60.5 & 23.5 & 49.5 & 29.5 & 60.5 & 24.5 & 50.0 & 28.0 & 59.5 & 24.3 & 53.3 \\
    w/ ShrinkA  & 22.5 & 43.0 & 27.5 & 58.5 & 27.0 & 59.5 & 28.0 & 65.0 & 21.5 & 46.5 & 36.0 & 69.5 & 20.0 & 49.0 & 38.0 & 71.5 & 27.0 & 56.5 & 38.5 & 73.0 & 28.6 & 59.2 \\ \midrule
    TSConv      & 27.5 & 65.0 & 31.5 & 64.5 & 33.0 & 69.0 & 38.5 & 75.0 & 24.0 & 60.5 & 42.5 & 75.0 & 33.5 & 66.0 & 42.0 & 77.5 & 32.5 & 65.5 & 36.5 & 75.5 & 34.2 & 69.4 \\
    w/ ShrinkA  & 34.0 & 65.0 & 40.0 & 69.0 & 38.5 & 72.0 & 40.5 & 78.5 & 31.0 & 57.5 & 48.0 & 81.0 & 36.0 & 66.5 & 52.5 & 82.0 & 33.5 & 70.5 & 48.0 & 78.5 & 40.2 & 72.1 \\ \midrule
    EEGProject  & 37.5 & 66.0 & 46.0 & 77.0 & 45.0 & 77.0 & 45.5 & 74.5 & 34.0 & 67.5 & 47.0 & 79.0 & 41.0 & 76.5 & 49.5 & 82.0 & 44.0 & 77.0 & 55.5 & 88.5 & 44.5 & 76.5 \\
    w/ ShrinkA  & 37.5 & 71.0 & 47.0 & 81.5 & 49.5 & 79.0 & 44.0 & 81.0 & 37.0 & 69.5 & 56.5 & 87.5 & 44.5 & 76.5 & 58.5 & 83.5 & 43.5 & 74.0 & 60.0 & 88.5 & 47.8 & 79.2 \\ \midrule
    STAE (Ours) & 39.0 & 64.5 & 46.0 & 80.5 & 46.5 & 77.0 & 43.5 & 76.5 & 37.0 & 73.0 & 49.5 & 78.5 & 44.0 & 75.5 & 50.5 & 83.5 & 45.5 & 77.0 & 53.0 & 85.5 & 45.5 & 77.2 \\
    w/ ShrinkA  & 44.0 & 74.0 & 54.5 & 85.5 & 49.5 & 82.5 & 49.5 & 82.5 & 43.5 & 73.0 & 53.0 & 85.5 & 48.5 & 77.0 & 61.0 & 87.0 & 47.5 & 79.0 & 56.0 & 87.5 & 50.7 & 81.4 \\
    \bottomrule
  \end{tabular}
  \caption{Performance comparison on THINGS-EEG using OpenCLIP ViT-B/16 as the vision encoder (latent dimension: 512). }
  \label{tab:compare_vit-b-16}
\end{table*}

\begin{table*}[t!]
  \centering
  \small
  \setlength{\tabcolsep}{0.8pt} 
  
  \begin{tabular}{lcccccccccccccccccccccc}
    \toprule
    \multirow{2}{*}{Backbone}& \multicolumn{2}{c}{Subject 1} & \multicolumn{2}{c}{Subject 2} & \multicolumn{2}{c}{Subject 3} & \multicolumn{2}{c}{Subject 4}  & \multicolumn{2}{c}{Subject 5}  & \multicolumn{2}{c}{Subject 6}  & \multicolumn{2}{c}{Subject 7}  & \multicolumn{2}{c}{Subject 8} & \multicolumn{2}{c}{Subject 9} & \multicolumn{2}{c}{Subject 10} & \multicolumn{2}{c}{Avg} \\
    \cmidrule(r){2-3} \cmidrule(r){4-5} \cmidrule(r){6-7} \cmidrule(r){8-9} \cmidrule(r){10-11} \cmidrule(r){12-13} \cmidrule(r){14-15} \cmidrule(r){16-17} \cmidrule(r){18-19} \cmidrule(r){20-21} \cmidrule(r){22-23}
    & top-1 & top-5 & top-1 & top-5 & top-1 & top-5 & top-1 & top-5 & top-1 & top-5 & top-1 & top-5 & top-1 & top-5 & top-1 & top-5 & top-1 & top-5 & top-1 & top-5 & top-1 & top-5\\
    \midrule
    ShallowNet  & 21.0 & 57.0 & 24.0 & 59.0 & 25.5 & 63.5 & 38.0 & 72.0 & 21.5 & 50.5 & 30.0 & 70.0 & 31.0 & 65.0 & 39.0 & 75.5 & 23.0 & 62.5 & 33.0 & 71.5 & 28.6 & 64.7 \\
    w/ ShrinkA  & 24.5 & 53.0 & 26.0 & 61.0 & 34.0 & 67.5 & 35.0 & 70.0 & 19.5 & 45.0 & 32.5 & 67.0 & 30.0 & 67.0 & 41.5 & 73.5 & 30.5 & 59.5 & 36.0 & 77.5 & 31.0 & 64.1 \\ \midrule
    DeepNet     & 10.0 & 26.5 & 10.0 & 36.5 & 8.5  & 38.5 & 15.0 & 43.0 & 10.5 & 26.0 & 12.5 & 37.5 & 12.0 & 32.5 & 12.5 & 29.5 & 11.0 & 39.5 & 14.0 & 49.5 & 11.6 & 35.9 \\
    w/ ShrinkA  & 2.0  & 8.0  & 1.5  & 8.5  & 5.0  & 12.0 & 7.0  & 23.0 & 3.0  & 13.0 & 2.5  & 12.5 & 5.0  & 12.0 & 4.5  & 17.0 & 1.0  & 7.0  & 4.5  & 23.5 & 3.6  & 13.7 \\ \midrule
    EEGNet      & 19.5 & 45.5 & 22.5 & 58.0 & 22.0 & 61.0 & 26.5 & 61.0 & 22.0 & 47.0 & 31.0 & 64.5 & 25.0 & 51.0 & 29.0 & 66.0 & 25.5 & 55.5 & 27.5 & 63.5 & 25.1 & 57.3 \\
    w/ ShrinkA  & 13.0 & 36.5 & 22.0 & 54.5 & 17.5 & 48.0 & 27.5 & 60.5 & 11.0 & 35.0 & 23.5 & 65.5 & 15.0 & 42.0 & 25.5 & 63.0 & 21.5 & 49.0 & 31.5 & 62.0 & 20.8 & 51.6 \\ \midrule
    TSConv      & 30.0 & 65.5 & 29.0 & 66.5 & 38.5 & 71.5 & 37.0 & 75.5 & 30.0 & 61.5 & 45.5 & 79.0 & 31.0 & 69.5 & 44.0 & 82.0 & 34.5 & 68.5 & 39.0 & 72.5 & 35.9 & 71.2 \\
    w/ ShrinkA  & 31.5 & 63.5 & 31.5 & 67.0 & 40.5 & 75.0 & 40.5 & 79.0 & 26.5 & 58.0 & 49.0 & 81.5 & 37.5 & 70.5 & 47.0 & 83.0 & 37.0 & 72.5 & 45.5 & 79.5 & 38.7 & 73.0 \\ \midrule
    EEGProject  & 38.0 & 69.0 & 47.0 & 81.0 & 53.0 & 83.5 & 48.5 & 79.5 & 37.0 & 72.0 & 50.5 & 82.0 & 42.0 & 79.5 & 60.5 & 82.0 & 49.5 & 76.0 & 57.0 & 91.5 & 48.3 & 79.6 \\
    w/ ShrinkA  & 42.0 & 73.5 & 49.5 & 82.0 & 55.0 & 84.0 & 48.0 & 84.0 & 40.5 & 78.0 & 57.5 & 88.5 & 53.5 & 83.5 & 62.0 & 90.5 & 49.5 & 79.0 & 61.5 & 92.5 & 51.9 & 83.6 \\ \midrule
    STAE (Ours) & 38.0 & 71.0 & 48.0 & 82.5 & 55.5 & 84.0 & 50.0 & 82.0 & 44.0 & 75.5 & 55.0 & 83.0 & 47.0 & 78.5 & 57.5 & 82.5 & 54.0 & 80.5 & 55.0 & 87.5 & 50.4 & 80.7 \\
    w/ ShrinkA  & 47.0 & 79.5 & 52.5 & 88.0 & 61.0 & 85.0 & 56.0 & 85.5 & 47.0 & 82.0 & 60.0 & 89.5 & 57.0 & 82.5 & 63.5 & 91.5 & 50.5 & 82.5 & 59.5 & 91.0 & 55.4 & 85.7 \\
    \bottomrule
  \end{tabular}
  \caption{Performance comparison on THINGS-EEG using OpenCLIP ViT-B/32 as the vision encoder (latent dimension: 512). }
  \label{tab:compare_vit-b-32}
\end{table*}

\begin{table*}[t!]
  \centering
  \small
  \setlength{\tabcolsep}{0.8pt} 
  
  \begin{tabular}{lcccccccccccccccccccccc}
    \toprule
    \multirow{2}{*}{Backbone}& \multicolumn{2}{c}{Subject 1} & \multicolumn{2}{c}{Subject 2} & \multicolumn{2}{c}{Subject 3} & \multicolumn{2}{c}{Subject 4}  & \multicolumn{2}{c}{Subject 5}  & \multicolumn{2}{c}{Subject 6}  & \multicolumn{2}{c}{Subject 7}  & \multicolumn{2}{c}{Subject 8} & \multicolumn{2}{c}{Subject 9} & \multicolumn{2}{c}{Subject 10} & \multicolumn{2}{c}{Avg} \\
    \cmidrule(r){2-3} \cmidrule(r){4-5} \cmidrule(r){6-7} \cmidrule(r){8-9} \cmidrule(r){10-11} \cmidrule(r){12-13} \cmidrule(r){14-15} \cmidrule(r){16-17} \cmidrule(r){18-19} \cmidrule(r){20-21} \cmidrule(r){22-23}
    & top-1 & top-5 & top-1 & top-5 & top-1 & top-5 & top-1 & top-5 & top-1 & top-5 & top-1 & top-5 & top-1 & top-5 & top-1 & top-5 & top-1 & top-5 & top-1 & top-5 & top-1 & top-5\\
    \midrule
    ShallowNet  & 21.5 & 51.0 & 22.0 & 55.0 & 28.0 & 64.0 & 37.5 & 66.5 & 21.0 & 47.5 & 31.5 & 68.5 & 25.0 & 60.0 & 36.0 & 72.5 & 27.5 & 59.0 & 40.5 & 68.5 & 29.1 & 61.3 \\
    w/ ShrinkA  & 23.0 & 53.5 & 25.0 & 52.5 & 27.0 & 64.0 & 34.5 & 63.0 & 17.5 & 47.0 & 32.0 & 67.5 & 29.5 & 60.5 & 39.5 & 70.0 & 28.5 & 61.0 & 40.0 & 73.5 & 29.7 & 61.3 \\ \midrule
    DeepNet     & 8.5  & 30.5 & 12.0 & 39.5 & 15.5 & 38.0 & 15.5 & 39.5 & 10.0 & 22.0 & 15.0 & 36.0 & 13.0 & 40.0 & 13.5 & 42.0 & 14.0 & 41.5 & 16.5 & 45.5 & 13.4 & 37.5 \\
    w/ ShrinkA  & 1.5  & 8.5  & 1.0  & 4.5  & 1.0  & 7.0  & 4.5  & 16.5 & 1.0  & 5.0  & 1.5  & 8.0  & 1.5  & 10.5 & 5.0  & 12.5 & 1.5  & 6.0  & 8.5  & 22.5 & 2.7  & 10.1 \\ \midrule
    EEGNet      & 15.0 & 43.0 & 21.0 & 54.5 & 23.5 & 58.0 & 25.5 & 62.0 & 22.0 & 51.0 & 28.0 & 63.5 & 22.5 & 52.0 & 27.5 & 62.5 & 24.0 & 58.5 & 32.0 & 67.5 & 24.1 & 57.3 \\
    w/ ShrinkA  & 10.0 & 29.0 & 19.5 & 45.5 & 16.0 & 41.5 & 26.5 & 55.0 & 11.5 & 29.5 & 23.5 & 54.5 & 18.5 & 47.5 & 24.0 & 63.5 & 21.5 & 49.0 & 34.0 & 64.5 & 20.5 & 48.0 \\ \midrule
    TSConv      & 29.5 & 62.5 & 31.5 & 67.0 & 36.0 & 70.0 & 38.5 & 72.5 & 26.5 & 59.0 & 38.5 & 76.0 & 34.0 & 63.5 & 40.0 & 79.0 & 27.0 & 64.5 & 46.0 & 77.0 & 34.8 & 69.1 \\
    w/ ShrinkA  & 31.5 & 65.0 & 34.5 & 64.5 & 39.5 & 68.0 & 44.0 & 74.0 & 28.5 & 59.5 & 51.0 & 82.0 & 35.0 & 63.0 & 48.0 & 79.5 & 40.5 & 70.0 & 50.0 & 82.0 & 40.3 & 70.8 \\ \midrule
    EEGProject  & 32.5 & 63.5 & 47.5 & 76.0 & 47.0 & 84.5 & 41.0 & 77.0 & 38.0 & 67.5 & 46.0 & 82.0 & 38.5 & 74.5 & 53.0 & 83.0 & 46.5 & 72.0 & 52.5 & 85.0 & 44.3 & 76.5 \\
    w/ ShrinkA  & 40.5 & 68.5 & 44.5 & 75.5 & 51.0 & 83.5 & 42.5 & 79.0 & 42.5 & 73.0 & 52.0 & 88.5 & 36.5 & 76.5 & 54.5 & 86.5 & 40.5 & 75.5 & 59.5 & 88.5 & 46.4 & 79.5 \\ \midrule
    STAE (Ours) & 37.0 & 65.5 & 42.0 & 78.0 & 45.5 & 82.0 & 42.0 & 76.5 & 36.5 & 71.0 & 50.0 & 82.5 & 34.5 & 73.0 & 54.5 & 83.0 & 44.5 & 73.0 & 48.0 & 87.0 & 43.5 & 77.2 \\
    w/ ShrinkA  & 39.5 & 68.5 & 48.0 & 82.5 & 51.5 & 83.0 & 44.5 & 81.0 & 48.5 & 77.0 & 57.0 & 89.0 & 43.0 & 74.5 & 54.0 & 87.0 & 46.0 & 76.5 & 62.5 & 89.0 & 49.5 & 80.8 \\
    \bottomrule
  \end{tabular}
  \caption{Performance comparison on THINGS-EEG using OpenCLIP ViT-H/14 as the vision encoder (latent dimension: 1024).}
  \label{tab:compare_vit-h-14}
\end{table*}

\begin{table*}[t!]
  \centering
  \small
  \setlength{\tabcolsep}{0.8pt} 
  
  \begin{tabular}{lcccccccccccccccccccccc}
    \toprule
    \multirow{2}{*}{Backbone}& \multicolumn{2}{c}{Subject 1} & \multicolumn{2}{c}{Subject 2} & \multicolumn{2}{c}{Subject 3} & \multicolumn{2}{c}{Subject 4}  & \multicolumn{2}{c}{Subject 5}  & \multicolumn{2}{c}{Subject 6}  & \multicolumn{2}{c}{Subject 7}  & \multicolumn{2}{c}{Subject 8} & \multicolumn{2}{c}{Subject 9} & \multicolumn{2}{c}{Subject 10} & \multicolumn{2}{c}{Avg} \\
    \cmidrule(r){2-3} \cmidrule(r){4-5} \cmidrule(r){6-7} \cmidrule(r){8-9} \cmidrule(r){10-11} \cmidrule(r){12-13} \cmidrule(r){14-15} \cmidrule(r){16-17} \cmidrule(r){18-19} \cmidrule(r){20-21} \cmidrule(r){22-23}
    & top-1 & top-5 & top-1 & top-5 & top-1 & top-5 & top-1 & top-5 & top-1 & top-5 & top-1 & top-5 & top-1 & top-5 & top-1 & top-5 & top-1 & top-5 & top-1 & top-5 & top-1 & top-5\\
    \midrule
    ShallowNet  & 18.5 & 52.0 & 19.0 & 53.5 & 28.0 & 64.0 & 31.5 & 69.0 & 20.0 & 45.0 & 29.0 & 63.0 & 26.0 & 59.0 & 30.5 & 72.0 & 26.0 & 57.0 & 33.5 & 69.5 & 26.2 & 60.4 \\
    w/ ShrinkA  & 23.0 & 54.0 & 25.0 & 59.0 & 32.0 & 66.5 & 35.0 & 66.5 & 22.5 & 52.5 & 35.5 & 67.0 & 30.5 & 63.5 & 41.5 & 72.0 & 30.5 & 62.0 & 36.0 & 76.5 & 31.2 & 64.0 \\ \midrule
    DeepNet     & 10.5 & 29.0 & 10.5 & 23.0 & 8.5  & 32.5 & 15.0 & 34.0 & 5.5  & 20.0 & 14.5 & 33.0 & 11.0 & 38.0 & 10.0 & 33.0 & 9.5  & 34.5 & 15.5 & 41.5 & 11.1 & 31.9 \\
    w/ ShrinkA  & 2.5  & 14.0 & 2.0  & 8.0  & 6.5  & 21.5 & 11.5 & 35.5 & 1.0  & 8.0  & 7.5  & 23.5 & 9.0  & 26.5 & 6.5  & 21.5 & 3.0  & 17.0 & 13.0 & 38.5 & 6.3  & 21.4 \\ \midrule
    EEGNet      & 15.5 & 41.5 & 19.5 & 48.5 & 25.5 & 57.5 & 27.0 & 60.0 & 19.0 & 51.5 & 26.5 & 60.5 & 22.0 & 52.0 & 27.0 & 61.0 & 20.5 & 54.0 & 32.0 & 61.0 & 23.5 & 54.8 \\
    w/ ShrinkA  & 21.0 & 39.5 & 19.5 & 49.5 & 24.0 & 53.5 & 28.5 & 63.5 & 18.5 & 44.5 & 30.5 & 66.5 & 24.5 & 54.5 & 31.0 & 71.0 & 29.5 & 58.5 & 36.5 & 73.5 & 26.4 & 57.5 \\ \midrule
    TSConv      & 30.5 & 63.5 & 25.0 & 61.5 & 32.0 & 64.5 & 33.0 & 67.0 & 26.0 & 56.0 & 32.5 & 70.5 & 33.5 & 62.5 & 38.5 & 78.5 & 30.5 & 64.5 & 40.5 & 77.5 & 32.2 & 66.6 \\
    w/ ShrinkA  & 31.5 & 65.5 & 28.0 & 62.5 & 41.5 & 71.0 & 38.5 & 66.0 & 28.0 & 59.5 & 47.0 & 78.5 & 35.0 & 67.0 & 48.5 & 81.0 & 39.5 & 68.5 & 49.5 & 78.0 & 38.7 & 69.8 \\ \midrule
    EEGProject  & 29.0 & 61.0 & 39.5 & 69.0 & 44.0 & 77.0 & 38.0 & 74.0 & 31.5 & 61.0 & 44.0 & 77.5 & 38.0 & 70.5 & 49.5 & 81.5 & 44.5 & 71.0 & 49.5 & 81.5 & 40.8 & 72.4 \\
    w/ ShrinkA  & 29.0 & 66.5 & 43.5 & 74.0 & 46.5 & 81.5 & 46.0 & 76.5 & 35.0 & 68.0 & 45.5 & 78.5 & 36.5 & 71.0 & 54.5 & 88.0 & 39.5 & 73.0 & 55.5 & 89.5 & 43.2 & 76.7 \\ \midrule
    STAE (Ours) & 27.5 & 64.5 & 41.5 & 68.5 & 43.5 & 76.5 & 43.5 & 74.5 & 37.5 & 66.5 & 45.0 & 79.0 & 39.0 & 72.0 & 50.0 & 82.5 & 44.0 & 71.5 & 50.5 & 80.5 & 42.2 & 73.6 \\
    w/ ShrinkA  & 31.5 & 67.0 & 43.0 & 78.5 & 46.5 & 83.0 & 42.5 & 80.0 & 39.0 & 75.0 & 51.5 & 82.5 & 39.5 & 65.5 & 54.0 & 89.0 & 45.5 & 76.5 & 50.0 & 86.0 & 44.3 & 78.3 \\
    \bottomrule
  \end{tabular}
  \caption{Performance comparison on THINGS-EEG using OpenCLIP ViT-g/14 as the vision encoder (latent dimension: 1024).}
  \label{tab:compare_vit-g-14}
\end{table*}

\begin{table*}[t!]
  \centering
  \small
  \setlength{\tabcolsep}{0.8pt} 
  
  \begin{tabular}{lcccccccccccccccccccccc}
    \toprule
    \multirow{2}{*}{Backbone}& \multicolumn{2}{c}{Subject 1} & \multicolumn{2}{c}{Subject 2} & \multicolumn{2}{c}{Subject 3} & \multicolumn{2}{c}{Subject 4}  & \multicolumn{2}{c}{Subject 5}  & \multicolumn{2}{c}{Subject 6}  & \multicolumn{2}{c}{Subject 7}  & \multicolumn{2}{c}{Subject 8} & \multicolumn{2}{c}{Subject 9} & \multicolumn{2}{c}{Subject 10} & \multicolumn{2}{c}{Avg} \\
    \cmidrule(r){2-3} \cmidrule(r){4-5} \cmidrule(r){6-7} \cmidrule(r){8-9} \cmidrule(r){10-11} \cmidrule(r){12-13} \cmidrule(r){14-15} \cmidrule(r){16-17} \cmidrule(r){18-19} \cmidrule(r){20-21} \cmidrule(r){22-23}
    & top-1 & top-5 & top-1 & top-5 & top-1 & top-5 & top-1 & top-5 & top-1 & top-5 & top-1 & top-5 & top-1 & top-5 & top-1 & top-5 & top-1 & top-5 & top-1 & top-5 & top-1 & top-5\\
    \midrule
    ShallowNet  & 19.5 & 51.5 & 21.5 & 60.0 & 29.5 & 65.5 & 39.0 & 70.0 & 19.0 & 51.0 & 37.5 & 68.5 & 30.0 & 64.0 & 34.5 & 71.0 & 26.5 & 60.0 & 33.0 & 71.0 & 29.0 & 63.3 \\
    w/ ShrinkA  & 26.0 & 51.0 & 27.0 & 58.5 & 30.5 & 69.5 & 38.5 & 68.5 & 16.0 & 49.5 & 38.0 & 70.0 & 31.5 & 68.0 & 43.0 & 74.5 & 31.0 & 65.5 & 40.0 & 78.5 & 32.2 & 65.4 \\ \midrule
    DeepNet     & 7.5  & 27.0 & 12.5 & 35.5 & 11.5 & 32.0 & 17.0 & 41.0 & 7.5  & 21.5 & 12.0 & 34.5 & 13.5 & 37.5 & 14.5 & 36.5 & 12.5 & 40.5 & 15.0 & 42.0 & 12.4 & 34.8 \\
    w/ ShrinkA  & 1.0  & 6.0  & 2.0  & 13.5 & 5.5  & 24.0 & 12.5 & 34.0 & 1.0  & 6.5  & 3.0  & 12.5 & 4.5  & 16.0 & 4.5  & 14.5 & 1.0  & 5.5  & 7.0  & 25.0 & 4.2  & 15.8 \\ \midrule
    EEGNet      & 17.0 & 46.0 & 24.0 & 58.0 & 25.0 & 56.0 & 28.5 & 61.0 & 17.0 & 47.0 & 28.0 & 60.5 & 22.5 & 54.0 & 31.5 & 64.5 & 25.5 & 57.0 & 29.0 & 62.5 & 24.8 & 56.7 \\
    w/ ShrinkA  & 18.0 & 38.5 & 27.5 & 56.0 & 20.0 & 54.0 & 29.0 & 59.5 & 19.0 & 45.5 & 32.0 & 62.5 & 23.0 & 51.0 & 41.0 & 72.0 & 30.5 & 61.0 & 36.0 & 73.0 & 27.6 & 57.3 \\ \midrule
    TSConv      & 30.0 & 61.0 & 31.5 & 66.0 & 31.5 & 71.0 & 36.0 & 73.0 & 24.0 & 58.5 & 46.0 & 74.0 & 35.5 & 70.0 & 43.0 & 79.0 & 30.0 & 67.0 & 37.0 & 76.0 & 34.5 & 69.6 \\
    w/ ShrinkA  & 33.0 & 66.0 & 36.0 & 69.5 & 38.0 & 72.5 & 46.0 & 77.5 & 29.5 & 60.0 & 49.5 & 79.0 & 35.0 & 69.0 & 45.0 & 81.0 & 39.5 & 74.0 & 43.5 & 83.0 & 39.5 & 73.2 \\ \midrule
    EEGProject  & 35.0 & 66.5 & 42.0 & 75.5 & 43.5 & 77.0 & 42.0 & 77.5 & 31.5 & 64.0 & 55.5 & 82.5 & 42.5 & 74.5 & 48.0 & 79.5 & 45.0 & 75.0 & 53.0 & 86.0 & 43.8 & 75.8 \\
    w/ ShrinkA  & 35.5 & 70.5 & 45.5 & 78.5 & 46.5 & 78.0 & 44.5 & 80.0 & 39.0 & 68.0 & 54.0 & 81.5 & 44.5 & 76.5 & 55.5 & 84.5 & 42.5 & 73.0 & 56.0 & 88.0 & 46.4 & 77.9 \\ \midrule
    STAE (Ours) & 35.0 & 65.5 & 43.5 & 78.0 & 44.5 & 77.5 & 40.5 & 76.0 & 35.0 & 67.0 & 57.0 & 84.0 & 41.5 & 76.0 & 45.5 & 82.5 & 47.0 & 77.5 & 52.5 & 85.5 & 44.2 & 77.0 \\
    w/ ShrinkA  & 38.0 & 68.0 & 48.5 & 81.0 & 47.0 & 78.5 & 49.0 & 82.0 & 40.0 & 73.5 & 55.5 & 83.5 & 43.0 & 76.0 & 54.0 & 86.5 & 44.5 & 81.0 & 56.5 & 88.0 & 47.6 & 79.8 \\
    \bottomrule
  \end{tabular}
  \caption{Performance comparison on THINGS-EEG using OpenCLIP ViT-bigG/14 as the vision encoder (latent dimension: 1280).}
  \label{tab:compare_vit-bigg-14}
\end{table*}

\begin{table*}[t!]
  \centering
  \small
  \setlength{\tabcolsep}{0.8pt} 
  
  \begin{tabular}{lcccccccccccccccccccccc}
    \toprule
    \multirow{2}{*}{Backbone}& \multicolumn{2}{c}{Subject 1} & \multicolumn{2}{c}{Subject 2} & \multicolumn{2}{c}{Subject 3} & \multicolumn{2}{c}{Subject 4}  & \multicolumn{2}{c}{Subject 5}  & \multicolumn{2}{c}{Subject 6}  & \multicolumn{2}{c}{Subject 7}  & \multicolumn{2}{c}{Subject 8} & \multicolumn{2}{c}{Subject 9} & \multicolumn{2}{c}{Subject 10} & \multicolumn{2}{c}{Avg} \\
    \cmidrule(r){2-3} \cmidrule(r){4-5} \cmidrule(r){6-7} \cmidrule(r){8-9} \cmidrule(r){10-11} \cmidrule(r){12-13} \cmidrule(r){14-15} \cmidrule(r){16-17} \cmidrule(r){18-19} \cmidrule(r){20-21} \cmidrule(r){22-23}
    & top-1 & top-5 & top-1 & top-5 & top-1 & top-5 & top-1 & top-5 & top-1 & top-5 & top-1 & top-5 & top-1 & top-5 & top-1 & top-5 & top-1 & top-5 & top-1 & top-5 & top-1 & top-5\\
    \midrule
    ShallowNet  & 9.5  & 25.5 & 18.5 & 39.0 & 14.0 & 39.5 & 16.5 & 45.5 & 9.0  & 25.5 & 15.0 & 41.5 & 12.0 & 36.0 & 14.5 & 39.0 & 11.5 & 31.0 & 15.5 & 43.0 & 13.6 & 36.6 \\
    w/ ShrinkA  & 8.5  & 25.0 & 14.0 & 38.0 & 16.0 & 42.0 & 21.0 & 45.0 & 10.5 & 30.5 & 15.5 & 44.0 & 20.0 & 40.5 & 20.5 & 45.0 & 14.0 & 33.0 & 17.5 & 47.5 & 15.8 & 39.1 \\ \midrule
    DeepNet     & 4.0  & 14.0 & 3.5  & 13.0 & 4.5  & 16.5 & 5.0  & 19.5 & 1.5  & 14.0 & 6.5  & 18.0 & 5.5  & 16.5 & 5.0  & 18.5 & 7.5  & 18.0 & 10.5 & 28.0 & 5.4  & 17.6 \\
    w/ ShrinkA  & 1.0  & 6.0  & 1.0  & 5.5  & 1.5  & 8.5  & 6.0  & 22.0 & 1.0  & 7.0  & 1.5  & 7.0  & 2.5  & 13.5 & 5.0  & 14.0 & 1.0  & 5.0  & 9.0  & 30.5 & 3.0  & 11.9 \\ \midrule
    EEGNet      & 4.0  & 13.5 & 9.0  & 29.0 & 9.0  & 25.0 & 8.5  & 32.0 & 5.0  & 22.0 & 7.0  & 28.5 & 9.5  & 28.5 & 10.0 & 32.5 & 6.5  & 25.0 & 12.0 & 31.5 & 8.1  & 26.8 \\
    w/ ShrinkA  & 5.0  & 20.0 & 12.5 & 40.5 & 12.5 & 35.0 & 18.5 & 34.5 & 9.0  & 26.0 & 15.0 & 35.5 & 13.5 & 35.5 & 16.0 & 45.5 & 12.0 & 38.0 & 18.5 & 42.5 & 13.3 & 35.3 \\ \midrule
    TSConv      & 12.5 & 34.0 & 12.5 & 33.5 & 14.0 & 35.0 & 16.5 & 48.0 & 12.0 & 33.0 & 16.5 & 45.0 & 12.5 & 38.5 & 16.0 & 43.5 & 16.0 & 34.0 & 22.5 & 43.5 & 15.1 & 38.8 \\
    w/ ShrinkA  & 13.0 & 36.5 & 12.5 & 32.0 & 11.0 & 32.5 & 17.5 & 44.0 & 12.0 & 30.0 & 18.0 & 40.0 & 17.0 & 41.0 & 18.0 & 45.0 & 12.0 & 34.0 & 23.5 & 47.0 & 15.5 & 38.2 \\ \midrule
    EEGProject  & 10.5 & 29.5 & 21.0 & 46.0 & 16.5 & 43.0 & 22.0 & 51.5 & 12.0 & 38.0 & 23.0 & 50.0 & 24.0 & 47.5 & 26.5 & 50.5 & 17.0 & 44.5 & 24.5 & 51.5 & 19.7 & 45.2 \\
    w/ ShrinkA  & 16.0 & 31.5 & 22.0 & 46.5 & 20.0 & 44.5 & 23.0 & 52.0 & 11.5 & 37.0 & 25.0 & 50.5 & 19.5 & 44.5 & 25.0 & 54.0 & 21.0 & 45.5 & 24.5 & 58.5 & 20.8 & 46.5 \\ \midrule
    STAE (Ours) & 13.0 & 36.5 & 27.5 & 49.5 & 20.5 & 48.0 & 24.0 & 59.0 & 14.5 & 41.0 & 23.5 & 52.0 & 21.5 & 53.5 & 31.5 & 59.0 & 21.0 & 50.0 & 32.5 & 56.0 & 23.0 & 50.5 \\
    w/ ShrinkA  & 14.5 & 31.5 & 24.5 & 53.0 & 23.5 & 45.5 & 22.0 & 52.0 & 15.0 & 38.5 & 27.0 & 55.0 & 22.0 & 47.5 & 26.5 & 55.5 & 22.0 & 54.0 & 29.5 & 60.5 & 22.7 & 49.3 \\
    \bottomrule
  \end{tabular}
  \caption{Performance comparison on THINGS-EEG using DINOv2 ViT-b/14 as the vision encoder (latent dimension: 768). }
  \label{tab:compare_dinov2-b14}
\end{table*}

\begin{table*}[t!]
  \centering
  \small
  \setlength{\tabcolsep}{0.8pt} 
  
  \begin{tabular}{lcccccccccccccccccccccc}
    \toprule
    \multirow{2}{*}{Backbone}& \multicolumn{2}{c}{Subject 1} & \multicolumn{2}{c}{Subject 2} & \multicolumn{2}{c}{Subject 3} & \multicolumn{2}{c}{Subject 4}  & \multicolumn{2}{c}{Subject 5}  & \multicolumn{2}{c}{Subject 6}  & \multicolumn{2}{c}{Subject 7}  & \multicolumn{2}{c}{Subject 8} & \multicolumn{2}{c}{Subject 9} & \multicolumn{2}{c}{Subject 10} & \multicolumn{2}{c}{Avg} \\
    \cmidrule(r){2-3} \cmidrule(r){4-5} \cmidrule(r){6-7} \cmidrule(r){8-9} \cmidrule(r){10-11} \cmidrule(r){12-13} \cmidrule(r){14-15} \cmidrule(r){16-17} \cmidrule(r){18-19} \cmidrule(r){20-21} \cmidrule(r){22-23}
    & top-1 & top-5 & top-1 & top-5 & top-1 & top-5 & top-1 & top-5 & top-1 & top-5 & top-1 & top-5 & top-1 & top-5 & top-1 & top-5 & top-1 & top-5 & top-1 & top-5 & top-1 & top-5\\
    \midrule
    ShallowNet  & 12.5 & 37.0 & 16.0 & 42.5 & 20.0 & 45.5 & 17.5 & 51.5 & 15.0 & 36.5 & 21.5 & 50.5 & 15.5 & 44.5 & 25.5 & 51.5 & 17.5 & 44.5 & 20.5 & 55.0 & 18.2 & 45.9 \\
    w/ ShrinkA  & 14.5 & 37.0 & 20.5 & 42.5 & 24.5 & 50.0 & 22.5 & 52.5 & 13.5 & 35.5 & 19.0 & 47.5 & 22.0 & 49.5 & 28.0 & 58.0 & 21.0 & 45.5 & 25.0 & 58.0 & 21.1 & 47.6 \\ \midrule
    DeepNet     & 5.5  & 16.5 & 6.5  & 22.0 & 5.5  & 23.5 & 8.5  & 23.0 & 5.5  & 17.5 & 7.0  & 23.5 & 7.0  & 23.5 & 7.0  & 26.5 & 10.5 & 29.5 & 7.5  & 29.5 & 7.1  & 23.5 \\
    w/ ShrinkA  & 1.0  & 5.5  & 1.0  & 5.5  & 1.0  & 8.0  & 6.0  & 20.5 & 0.5  & 3.0  & 3.0  & 7.0  & 1.5  & 9.5  & 3.0  & 10.5 & 3.0  & 10.0 & 4.0  & 25.0 & 2.4  & 10.5 \\ \midrule
    EEGNet      & 6.5  & 21.0 & 9.5  & 25.5 & 9.0  & 26.0 & 11.5 & 36.5 & 6.0  & 19.5 & 14.0 & 34.0 & 8.5  & 28.5 & 13.0 & 40.0 & 12.5 & 31.5 & 15.5 & 42.0 & 10.6 & 30.5 \\
    w/ ShrinkA  & 8.0  & 21.5 & 10.5 & 31.5 & 10.0 & 25.0 & 16.5 & 40.5 & 6.5  & 22.0 & 15.0 & 37.5 & 9.0  & 28.0 & 17.0 & 45.5 & 11.0 & 30.0 & 20.0 & 48.5 & 12.4 & 33.0 \\ \midrule
    TSConv      & 18.0 & 40.0 & 17.0 & 38.5 & 18.5 & 48.5 & 24.5 & 50.5 & 14.5 & 35.0 & 28.0 & 53.0 & 24.5 & 51.5 & 27.0 & 55.5 & 21.0 & 49.0 & 22.5 & 55.5 & 21.6 & 47.7 \\
    w/ ShrinkA  & 20.5 & 46.5 & 16.5 & 42.0 & 23.0 & 48.0 & 24.0 & 56.0 & 17.5 & 37.0 & 28.5 & 58.0 & 25.5 & 53.5 & 29.5 & 61.5 & 21.0 & 47.5 & 29.0 & 62.0 & 23.5 & 51.2 \\ \midrule
    EEGProject  & 20.0 & 41.5 & 22.5 & 55.5 & 21.5 & 59.0 & 25.5 & 56.0 & 23.0 & 50.5 & 28.0 & 59.5 & 22.5 & 57.5 & 33.0 & 63.5 & 30.0 & 60.5 & 36.5 & 65.0 & 26.3 & 56.9 \\
    w/ ShrinkA  & 24.0 & 46.5 & 29.5 & 56.0 & 27.0 & 56.5 & 30.0 & 59.5 & 23.5 & 50.5 & 35.0 & 66.0 & 28.5 & 62.0 & 37.5 & 71.0 & 24.0 & 56.5 & 41.5 & 65.5 & 30.1 & 59.0 \\ \midrule
    STAE (Ours) & 20.0 & 46.0 & 29.0 & 57.5 & 25.5 & 64.0 & 28.0 & 62.5 & 24.0 & 55.0 & 33.0 & 64.5 & 26.5 & 60.0 & 38.5 & 70.5 & 33.0 & 63.0 & 35.5 & 69.5 & 29.3 & 61.3 \\
    w/ ShrinkA  & 25.5 & 49.5 & 30.5 & 59.5 & 32.0 & 59.0 & 32.5 & 64.0 & 26.0 & 51.5 & 36.0 & 69.5 & 28.5 & 63.0 & 38.0 & 69.5 & 27.0 & 60.5 & 36.0 & 67.5 & 31.2 & 61.4 \\
    \bottomrule
  \end{tabular}
  \caption{Performance comparison on THINGS-EEG using DINOv2 ViT-l/14 as the vision encoder (latent dimension: 1024).}
  \label{tab:compare_dinov2-l14}
\end{table*}

\begin{table*}[t!]
  \centering
  \small
  \setlength{\tabcolsep}{0.8pt} 
  
  \begin{tabular}{lcccccccccccccccccccccc}
    \toprule
    \multirow{2}{*}{Backbone}& \multicolumn{2}{c}{Subject 1} & \multicolumn{2}{c}{Subject 2} & \multicolumn{2}{c}{Subject 3} & \multicolumn{2}{c}{Subject 4}  & \multicolumn{2}{c}{Subject 5}  & \multicolumn{2}{c}{Subject 6}  & \multicolumn{2}{c}{Subject 7}  & \multicolumn{2}{c}{Subject 8} & \multicolumn{2}{c}{Subject 9} & \multicolumn{2}{c}{Subject 10} & \multicolumn{2}{c}{Avg} \\
    \cmidrule(r){2-3} \cmidrule(r){4-5} \cmidrule(r){6-7} \cmidrule(r){8-9} \cmidrule(r){10-11} \cmidrule(r){12-13} \cmidrule(r){14-15} \cmidrule(r){16-17} \cmidrule(r){18-19} \cmidrule(r){20-21} \cmidrule(r){22-23}
    & top-1 & top-5 & top-1 & top-5 & top-1 & top-5 & top-1 & top-5 & top-1 & top-5 & top-1 & top-5 & top-1 & top-5 & top-1 & top-5 & top-1 & top-5 & top-1 & top-5 & top-1 & top-5\\
    \midrule
    ShallowNet  & 8.0  & 26.5 & 13.0 & 34.0 & 14.5 & 39.5 & 17.5 & 42.5 & 8.5  & 29.0 & 15.5 & 44.5 & 19.0 & 42.0 & 19.5 & 43.0 & 15.0 & 39.0 & 23.0 & 49.0 & 15.4 & 38.9 \\
    w/ ShrinkA  & 6.0  & 27.0 & 13.5 & 29.0 & 14.0 & 37.5 & 14.0 & 40.0 & 5.0  & 20.5 & 8.5  & 33.5 & 15.0 & 38.5 & 19.0 & 46.0 & 16.5 & 39.5 & 23.0 & 50.0 & 13.5 & 36.2 \\ \midrule
    DeepNet     & 3.5  & 13.5 & 4.0  & 14.0 & 5.5  & 14.5 & 5.5  & 21.5 & 2.0  & 10.5 & 5.0  & 19.0 & 8.0  & 21.0 & 5.0  & 18.5 & 8.0  & 24.5 & 8.0  & 24.5 & 5.5  & 18.2 \\
    w/ ShrinkA  & 1.0  & 6.0  & 0.5  & 5.0  & 1.0  & 7.5  & 3.5  & 16.0 & 1.5  & 6.0  & 1.5  & 5.0  & 1.5  & 7.5  & 4.0  & 14.0 & 1.5  & 5.5  & 5.0  & 14.0 & 2.1  & 8.7  \\ \midrule
    EEGNet      & 7.5  & 24.0 & 9.5  & 29.0 & 9.5  & 27.5 & 14.5 & 36.5 & 7.5  & 24.5 & 11.5 & 34.0 & 14.5 & 29.5 & 11.5 & 35.0 & 13.0 & 34.0 & 18.0 & 38.5 & 11.7 & 31.3 \\
    w/ ShrinkA  & 5.0  & 23.0 & 9.0  & 28.0 & 10.5 & 23.5 & 12.5 & 34.0 & 6.5  & 21.5 & 9.0  & 28.0 & 11.0 & 31.0 & 16.5 & 39.5 & 10.0 & 28.0 & 18.0 & 39.0 & 10.8 & 29.6 \\ \midrule
    TSConv      & 11.0 & 31.0 & 13.0 & 35.5 & 15.0 & 38.5 & 20.5 & 44.0 & 14.0 & 36.0 & 18.0 & 42.5 & 13.5 & 41.0 & 18.0 & 45.5 & 19.0 & 45.5 & 20.0 & 50.0 & 16.2 & 41.0 \\
    w/ ShrinkA  & 10.5 & 27.5 & 10.5 & 31.5 & 13.5 & 41.0 & 16.5 & 43.0 & 10.5 & 24.0 & 20.0 & 40.5 & 14.0 & 39.0 & 18.0 & 44.5 & 12.5 & 38.5 & 19.5 & 46.5 & 14.6 & 37.6 \\ \midrule
    EEGProject  & 13.0 & 32.5 & 21.0 & 42.5 & 25.5 & 51.5 & 18.0 & 52.5 & 19.0 & 38.0 & 21.0 & 48.0 & 16.5 & 46.5 & 26.0 & 50.0 & 16.0 & 46.5 & 26.5 & 55.0 & 20.3 & 46.3 \\
    w/ ShrinkA  & 18.0 & 34.0 & 19.5 & 42.0 & 26.5 & 53.0 & 24.5 & 47.0 & 15.0 & 40.0 & 22.0 & 50.5 & 22.5 & 51.0 & 27.5 & 56.5 & 20.5 & 50.0 & 34.0 & 59.0 & 23.0 & 48.3 \\ \midrule
    STAE (Ours) & 15.5 & 38.5 & 23.5 & 46.0 & 24.0 & 54.0 & 23.0 & 53.0 & 18.0 & 45.5 & 24.0 & 51.0 & 16.5 & 52.5 & 29.5 & 55.0 & 20.5 & 49.0 & 28.0 & 61.0 & 22.3 & 50.6 \\
    w/ ShrinkA  & 17.5 & 36.0 & 21.0 & 45.5 & 25.0 & 54.5 & 25.5 & 47.0 & 17.0 & 42.5 & 23.5 & 54.0 & 23.0 & 47.5 & 29.0 & 58.5 & 22.5 & 49.0 & 30.5 & 56.5 & 23.5 & 49.1 \\
    \bottomrule
  \end{tabular}
  \caption{Performance comparison on THINGS-EEG using DINOv2 ViT-g/14 as the vision encoder (latent dimension: 1536).}
  \label{tab:compare_dinov2-g14}
\end{table*}
\setlength{\abovecaptionskip}{4pt}
\setlength{\belowcaptionskip}{4pt}
\setlength{\floatsep}{6pt}    
\setlength{\textfloatsep}{6pt} 
\setlength{\intextsep}{6pt}    
\renewcommand{\arraystretch}{0.85} 

\begin{table}[h]
  \centering
  \small
  \setlength{\tabcolsep}{0.8pt} 
  \begin{tabular}{lcccccccccc}
    \toprule
    \multirow{2}{*}{Method}& \multicolumn{2}{c}{Subject 1} & \multicolumn{2}{c}{Subject 2} & \multicolumn{2}{c}{Subject 3} & \multicolumn{2}{c}{Subject 4}& \multicolumn{2}{c}{Avg} \\
    \cmidrule(r){2-3} \cmidrule(r){4-5} \cmidrule(r){6-7} \cmidrule(r){8-9} \cmidrule(r){10-11}
    & top-1 & top-5 & top-1 & top-5 & top-1 & top-5 & top-1 & top-5 & top-1 & top-5 \\

    \midrule
ShallowNet  & 8.0  & 25.0 & 22.5 & 54.0 & 16.0 & 43.5 & 8.0  & 28.5 & 13.6 & 37.8 \\
w/ ShrinkA  & 3.0  & 13.0 & 19.5 & 51.0 & 7.0  & 27.5 & 5.5  & 17.5 & 8.8  & 27.3 \\ \midrule
DeepNet     & 3.5  & 13.0 & 17.0 & 40.5 & 11.0 & 30.0 & 4.0  & 17.5 & 8.9  & 25.3 \\
w/ ShrinkA  & 1.5  & 5.5  & 8.0  & 23.0 & 2.0  & 6.5  & 0.5  & 3.0  & 3.0  & 9.5  \\ \midrule
EEGNet      & 4.0  & 14.5 & 20.0 & 56.0 & 14.5 & 37.0 & 5.0  & 20.5 & 10.9 & 32.0 \\
w/ ShrinkA  & 1.0  & 7.0  & 15.0 & 43.0 & 4.5  & 18.5 & 1.5  & 9.5  & 5.5  & 19.5 \\ \midrule
TSConv      & 11.0 & 35.0 & 28.0 & 67.5 & 22.5 & 56.5 & 13.0 & 37.0 & 18.6 & 49.0 \\
w/ ShrinkA  & 5.5  & 17.0 & 31.0 & 61.5 & 15.5 & 44.0 & 5.5  & 20.5 & 14.4 & 35.8 \\ \midrule
EEGProject  & 9.0  & 30.5 & 36.0 & 70.5 & 25.0 & 52.0 & 12.5 & 38.0 & 20.6 & 47.8 \\
w/ ShrinkA  & 8.0  & 32.0 & 46.5 & 82.5 & 34.0 & 62.5 & 14.5 & 38.0 & 25.8 & 53.8 \\ \midrule
STAE (Ours) & 8.5  & 26.0 & 35.0 & 69.5 & 22.5 & 48.5 & 10.5 & 30.0 & 19.1 & 43.5 \\
w/ ShrinkA  & 14.0 & 40.0 & 55.5 & 84.0 & 29.5 & 63.0 & 15.5 & 44.5 & 28.6 & 57.9 \\ \bottomrule
  \end{tabular}
  \caption{Performance comparison on THINGS-MEG using OpenCLIP RN50 as the vision encoder (latent dimension: 1024). }
  \label{tab:compare_RN50_MEG}
\end{table}

\begin{table}[h]
  \centering
  \small
  \setlength{\tabcolsep}{0.8pt} 
  \begin{tabular}{lcccccccccc}
    \toprule
    \multirow{2}{*}{Method}& \multicolumn{2}{c}{Subject 1} & \multicolumn{2}{c}{Subject 2} & \multicolumn{2}{c}{Subject 3} & \multicolumn{2}{c}{Subject 4}& \multicolumn{2}{c}{Avg} \\
    \cmidrule(r){2-3} \cmidrule(r){4-5} \cmidrule(r){6-7} \cmidrule(r){8-9} \cmidrule(r){10-11}
    & top-1 & top-5 & top-1 & top-5 & top-1 & top-5 & top-1 & top-5 & top-1 & top-5 \\

    \midrule
ShallowNet  & 9.0  & 29.5 & 25.5 & 62.0 & 18.5 & 49.5 & 9.5  & 31.0 & 15.6 & 43.0 \\
w/ ShrinkA  & 8.5  & 23.0 & 15.5 & 44.0 & 11.0 & 35.0 & 4.0  & 15.5 & 9.8  & 29.4 \\ \midrule
DeepNet     & 7.5  & 20.5 & 16.5 & 44.5 & 14.5 & 38.0 & 5.5  & 20.0 & 11.0 & 30.8 \\
w/ ShrinkA  & 1.0  & 6.0  & 5.0  & 26.0 & 2.5  & 12.5 & 1.0  & 5.5  & 2.4  & 12.5 \\ \midrule
EEGNet      & 7.5  & 26.0 & 22.0 & 54.5 & 19.5 & 45.5 & 5.5  & 21.0 & 13.6 & 36.8 \\
w/ ShrinkA  & 3.0  & 16.0 & 11.5 & 31.0 & 6.5  & 19.0 & 4.0  & 10.0 & 6.3  & 19.0 \\ \midrule
TSConv      & 13.5 & 37.0 & 38.5 & 71.5 & 25.5 & 60.0 & 15.5 & 44.0 & 23.3 & 53.1 \\
w/ ShrinkA  & 10.0 & 32.0 & 27.0 & 61.0 & 13.5 & 46.0 & 5.0  & 19.5 & 13.9 & 39.6 \\ \midrule
EEGProject  & 13.0 & 35.5 & 44.5 & 79.0 & 27.0 & 60.0 & 16.0 & 43.0 & 25.1 & 54.4 \\
w/ ShrinkA  & 13.0 & 31.5 & 46.0 & 75.0 & 27.5 & 61.0 & 20.0 & 37.5 & 26.6 & 51.3 \\ \midrule
STAE (Ours) & 13.5 & 35.5 & 47.0 & 78.5 & 25.5 & 61.5 & 19.0 & 42.0 & 26.3 & 54.4 \\
w/ ShrinkA  & 12.0 & 39.5 & 48.5 & 79.5 & 28.0 & 59.5 & 16.5 & 39.0 & 26.3 & 54.4 \\ \bottomrule
  \end{tabular}
  \caption{Performance comparison on THINGS-MEG using OpenCLIP ViT-L/14 as the vision encoder (latent dimension: 768). }
  \label{tab:compare_ViT-L/14_MEG}
\end{table}

\begin{table}[h]
  \centering
  \small
  \setlength{\tabcolsep}{0.8pt} 
  \begin{tabular}{lcccccccccc}
    \toprule
    \multirow{2}{*}{Method}& \multicolumn{2}{c}{Subject 1} & \multicolumn{2}{c}{Subject 2} & \multicolumn{2}{c}{Subject 3} & \multicolumn{2}{c}{Subject 4}& \multicolumn{2}{c}{Avg} \\
    \cmidrule(r){2-3} \cmidrule(r){4-5} \cmidrule(r){6-7} \cmidrule(r){8-9} \cmidrule(r){10-11}
    & top-1 & top-5 & top-1 & top-5 & top-1 & top-5 & top-1 & top-5 & top-1 & top-5 \\

    \midrule
ShallowNet  & 7.0  & 25.0 & 20.5 & 50.0 & 15.0 & 43.5 & 9.0  & 32.0 & 12.9 & 37.6 \\
w/ ShrinkA  & 2.0  & 10.0 & 6.0  & 25.5 & 5.0  & 15.0 & 3.0  & 9.0  & 4.0  & 14.9 \\ \midrule
DeepNet     & 4.5  & 12.0 & 14.0 & 36.0 & 8.5  & 26.0 & 5.0  & 18.5 & 8.0  & 23.1 \\
w/ ShrinkA  & 0.5  & 4.0  & 2.5  & 14.0 & 1.5  & 3.5  & 0.5  & 4.0  & 1.3  & 6.4  \\ \midrule
EEGNet      & 6.5  & 19.5 & 19.0 & 46.0 & 10.5 & 27.5 & 5.5  & 16.5 & 10.4 & 27.4 \\
w/ ShrinkA  & 0.0  & 3.0  & 4.0  & 19.5 & 1.5  & 5.5  & 1.0  & 5.5  & 1.6  & 8.4  \\ \midrule
TSConv      & 9.0  & 32.0 & 26.0 & 62.0 & 20.5 & 52.0 & 12.0 & 37.5 & 16.9 & 45.9 \\
w/ ShrinkA  & 2.5  & 10.0 & 9.0  & 36.0 & 5.0  & 14.0 & 2.5  & 8.0  & 4.8  & 17.0 \\ \midrule
EEGProject  & 11.0 & 36.5 & 32.0 & 66.5 & 21.0 & 54.5 & 12.0 & 35.0 & 19.0 & 48.1 \\
w/ ShrinkA  & 10.5 & 31.0 & 43.0 & 75.0 & 29.0 & 62.5 & 12.0 & 37.5 & 23.6 & 51.5 \\ \midrule
STAE (Ours) & 12.0 & 33.5 & 35.0 & 63.5 & 21.0 & 49.5 & 10.0 & 32.0 & 19.5 & 44.6 \\
w/ ShrinkA  & 10.0 & 37.5 & 46.5 & 80.5 & 32.5 & 67.5 & 12.5 & 37.0 & 25.4 & 55.6 \\ \bottomrule
  \end{tabular}
  \caption{Performance comparison on THINGS-MEG using OpenCLIP RN101 as the vision encoder (latent dimension: 512). }
  \label{tab:compare_RN101_MEG}
\end{table}

\begin{table}[h]
  \centering
  \small
  \setlength{\tabcolsep}{0.8pt} 
  \begin{tabular}{lcccccccccc}
    \toprule
    \multirow{2}{*}{Method}& \multicolumn{2}{c}{Subject 1} & \multicolumn{2}{c}{Subject 2} & \multicolumn{2}{c}{Subject 3} & \multicolumn{2}{c}{Subject 4}& \multicolumn{2}{c}{Avg} \\
    \cmidrule(r){2-3} \cmidrule(r){4-5} \cmidrule(r){6-7} \cmidrule(r){8-9} \cmidrule(r){10-11}
    & top-1 & top-5 & top-1 & top-5 & top-1 & top-5 & top-1 & top-5 & top-1 & top-5 \\

    \midrule
ShallowNet  & 9.0  & 28.0 & 25.5 & 58.5 & 11.5 & 40.0 & 11.5 & 30.0 & 14.4 & 39.1 \\
w/ ShrinkA  & 4.5  & 15.0 & 9.5  & 30.5 & 5.5  & 23.5 & 5.0  & 12.0 & 6.1  & 20.3 \\ \midrule
DeepNet     & 5.0  & 21.5 & 14.5 & 39.5 & 9.5  & 34.0 & 5.5  & 21.0 & 8.6  & 29.0 \\
w/ ShrinkA  & 0.0  & 5.0  & 5.5  & 17.0 & 3.0  & 9.0  & 1.0  & 4.5  & 2.4  & 8.9  \\ \midrule
EEGNet      & 5.0  & 20.0 & 22.5 & 59.5 & 12.5 & 36.5 & 4.5  & 18.0 & 11.1 & 33.5 \\
w/ ShrinkA  & 0.5  & 5.0  & 7.5  & 23.5 & 2.0  & 10.5 & 1.5  & 5.0  & 2.9  & 11.0 \\ \midrule
TSConv      & 13.0 & 39.0 & 35.5 & 70.5 & 21.5 & 54.5 & 14.0 & 43.5 & 21.0 & 51.9 \\
w/ ShrinkA  & 4.5  & 16.0 & 23.5 & 47.0 & 6.0  & 28.0 & 3.0  & 16.0 & 9.3  & 26.8 \\ \midrule
EEGProject  & 12.5 & 31.5 & 41.0 & 78.5 & 26.0 & 54.5 & 13.0 & 41.0 & 23.1 & 51.4 \\
w/ ShrinkA  & 15.0 & 38.0 & 46.0 & 82.5 & 29.0 & 61.0 & 14.0 & 43.5 & 26.0 & 56.3 \\ \midrule
STAE (Ours) & 10.0 & 32.0 & 39.0 & 73.0 & 22.5 & 51.5 & 8.5  & 34.0 & 20.0 & 47.6 \\
w/ ShrinkA  & 14.5 & 42.0 & 55.5 & 84.0 & 31.5 & 63.0 & 17.0 & 46.5 & 29.6 & 58.9 \\ \bottomrule
  \end{tabular}
  \caption{Performance comparison on THINGS-MEG using OpenCLIP ViT-B/16 as the vision encoder (latent dimension: 512). }
  \label{tab:compare_ViT-B/16_MEG}
\end{table}

\begin{table}[h]
  \centering
  \small
  \setlength{\tabcolsep}{0.8pt} 
  \begin{tabular}{lcccccccccc}
    \toprule
    \multirow{2}{*}{Method}& \multicolumn{2}{c}{Subject 1} & \multicolumn{2}{c}{Subject 2} & \multicolumn{2}{c}{Subject 3} & \multicolumn{2}{c}{Subject 4}& \multicolumn{2}{c}{Avg} \\
    \cmidrule(r){2-3} \cmidrule(r){4-5} \cmidrule(r){6-7} \cmidrule(r){8-9} \cmidrule(r){10-11}
    & top-1 & top-5 & top-1 & top-5 & top-1 & top-5 & top-1 & top-5 & top-1 & top-5 \\

    \midrule
ShallowNet  & 8.0  & 22.0 & 23.0 & 56.0 & 16.0 & 51.0 & 7.5  & 26.0 & 13.6 & 38.8 \\
w/ ShrinkA  & 2.0  & 14.5 & 13.5 & 35.5 & 8.0  & 27.0 & 3.0  & 16.0 & 6.6  & 23.3 \\ \midrule
DeepNet     & 3.5  & 17.0 & 15.0 & 39.0 & 10.5 & 34.0 & 6.0  & 20.0 & 8.8  & 27.5 \\
w/ ShrinkA  & 0.5  & 4.0  & 6.0  & 22.0 & 2.0  & 10.5 & 0.5  & 3.0  & 2.3  & 9.9  \\ \midrule
EEGNet      & 4.0  & 16.5 & 26.0 & 60.0 & 16.0 & 39.0 & 4.0  & 19.5 & 12.5 & 33.8 \\
w/ ShrinkA  & 1.0  & 4.5  & 8.0  & 28.0 & 2.5  & 12.0 & 1.5  & 5.0  & 3.3  & 12.4 \\ \midrule
TSConv      & 11.0 & 34.0 & 34.5 & 70.0 & 28.5 & 60.0 & 11.5 & 41.0 & 21.4 & 51.3 \\
w/ ShrinkA  & 2.5  & 17.5 & 20.5 & 48.5 & 10.5 & 31.0 & 1.5  & 11.0 & 8.8  & 27.0 \\ \midrule
EEGProject  & 7.0  & 32.0 & 44.5 & 73.5 & 29.5 & 59.0 & 15.5 & 40.0 & 24.1 & 51.1 \\
w/ ShrinkA  & 12.0 & 35.5 & 43.5 & 79.5 & 27.0 & 60.5 & 12.5 & 36.5 & 23.8 & 53.0 \\ \midrule
STAE (Ours) & 8.5  & 29.5 & 44.0 & 72.0 & 31.0 & 60.0 & 14.0 & 37.0 & 24.4 & 49.6 \\
w/ ShrinkA  & 13.0 & 40.0 & 48.0 & 83.0 & 30.5 & 66.5 & 14.0 & 38.5 & 26.4 & 57.0 \\ \bottomrule
  \end{tabular}
  \caption{Performance comparison on THINGS-MEG using OpenCLIP ViT-B/32 as the vision encoder (latent dimension: 512). }
  \label{tab:compare_ViT-B/32_MEG}
\end{table}

\begin{table}[h]
  \centering
  \small
  \setlength{\tabcolsep}{0.8pt} 
  \begin{tabular}{lcccccccccc}
    \toprule
    \multirow{2}{*}{Method}& \multicolumn{2}{c}{Subject 1} & \multicolumn{2}{c}{Subject 2} & \multicolumn{2}{c}{Subject 3} & \multicolumn{2}{c}{Subject 4}& \multicolumn{2}{c}{Avg} \\
    \cmidrule(r){2-3} \cmidrule(r){4-5} \cmidrule(r){6-7} \cmidrule(r){8-9} \cmidrule(r){10-11}
    & top-1 & top-5 & top-1 & top-5 & top-1 & top-5 & top-1 & top-5 & top-1 & top-5 \\

    \midrule
ShallowNet  & 9.0  & 31.0 & 27.0 & 56.5 & 20.0 & 51.0 & 13.0 & 36.0 & 17.3 & 43.6 \\
w/ ShrinkA  & 8.5  & 26.5 & 27.0 & 58.5 & 14.5 & 40.5 & 5.0  & 21.5 & 13.8 & 36.8 \\ \midrule
DeepNet     & 4.0  & 20.0 & 20.0 & 44.5 & 16.0 & 35.5 & 8.5  & 21.5 & 12.1 & 30.4 \\
w/ ShrinkA  & 0.5  & 2.5  & 11.5 & 33.0 & 4.5  & 14.0 & 1.0  & 3.0  & 4.4  & 13.1 \\ \midrule
EEGNet      & 7.0  & 26.0 & 27.0 & 60.5 & 18.0 & 46.5 & 8.5  & 27.0 & 15.1 & 40.0 \\
w/ ShrinkA  & 2.0  & 10.0 & 27.5 & 58.5 & 15.0 & 32.0 & 5.0  & 16.5 & 12.4 & 29.3 \\ \midrule
TSConv      & 14.5 & 43.5 & 36.0 & 72.5 & 27.0 & 63.0 & 14.0 & 39.0 & 22.9 & 54.5 \\
w/ ShrinkA  & 13.5 & 32.5 & 41.5 & 71.0 & 25.0 & 59.0 & 5.5  & 26.0 & 21.4 & 47.1 \\ \midrule
EEGProject  & 14.5 & 42.5 & 47.5 & 76.0 & 34.5 & 65.0 & 16.5 & 43.0 & 28.3 & 56.6 \\
w/ ShrinkA  & 10.0 & 35.5 & 46.5 & 73.5 & 30.5 & 62.0 & 12.0 & 39.0 & 24.8 & 52.5 \\ \midrule
STAE (Ours) & 13.5 & 39.0 & 47.0 & 75.0 & 36.0 & 63.0 & 16.0 & 42.5 & 28.1 & 54.9 \\
w/ ShrinkA  & 16.0 & 34.5 & 48.0 & 78.5 & 31.5 & 63.0 & 15.5 & 40.5 & 27.8 & 54.1 \\ \bottomrule
  \end{tabular}
  \caption{Performance comparison on THINGS-MEG using OpenCLIP ViT-H/14 as the vision encoder (latent dimension: 1024). }
  \label{tab:compare_ViT-H/14_MEG}
\end{table}

\begin{table}[h]
  \centering
  \small
  \setlength{\tabcolsep}{0.8pt} 
  \begin{tabular}{lcccccccccc}
    \toprule
    \multirow{2}{*}{Method}& \multicolumn{2}{c}{Subject 1} & \multicolumn{2}{c}{Subject 2} & \multicolumn{2}{c}{Subject 3} & \multicolumn{2}{c}{Subject 4}& \multicolumn{2}{c}{Avg} \\
    \cmidrule(r){2-3} \cmidrule(r){4-5} \cmidrule(r){6-7} \cmidrule(r){8-9} \cmidrule(r){10-11}
    & top-1 & top-5 & top-1 & top-5 & top-1 & top-5 & top-1 & top-5 & top-1 & top-5 \\

    \midrule
ShallowNet  & 7.5  & 26.0 & 29.5 & 57.0 & 19.5 & 47.0 & 10.0 & 27.0 & 16.6 & 39.3 \\
w/ ShrinkA  & 6.5  & 31.0 & 32.0 & 64.5 & 18.5 & 54.0 & 9.0  & 32.0 & 16.5 & 45.4 \\ \midrule
DeepNet     & 5.0  & 16.0 & 14.5 & 43.0 & 11.0 & 33.5 & 2.5  & 22.0 & 8.3  & 28.6 \\
w/ ShrinkA  & 1.5  & 7.0  & 17.0 & 43.5 & 6.5  & 29.0 & 4.0  & 13.5 & 7.3  & 23.3 \\ \midrule
EEGNet      & 6.5  & 19.5 & 23.0 & 58.5 & 15.0 & 40.0 & 6.0  & 25.5 & 12.6 & 35.9 \\
w/ ShrinkA  & 6.0  & 19.0 & 33.0 & 64.5 & 10.0 & 38.5 & 6.0  & 21.5 & 13.8 & 35.9 \\ \midrule
TSConv      & 10.5 & 37.5 & 33.0 & 73.0 & 23.0 & 56.0 & 11.5 & 35.0 & 19.5 & 50.4 \\
w/ ShrinkA  & 12.5 & 37.0 & 40.5 & 73.5 & 28.5 & 58.5 & 9.0  & 38.0 & 22.6 & 51.8 \\ \midrule
EEGProject  & 10.0 & 35.0 & 40.5 & 76.0 & 23.0 & 56.5 & 14.5 & 37.0 & 22.0 & 51.1 \\
w/ ShrinkA  & 8.0  & 29.5 & 42.5 & 74.0 & 23.5 & 52.5 & 12.5 & 36.5 & 21.6 & 48.1 \\ \midrule
STAE (Ours) & 13.0 & 36.0 & 42.5 & 76.5 & 26.5 & 55.5 & 17.5 & 34.5 & 24.9 & 50.6 \\
w/ ShrinkA  & 7.0  & 31.0 & 40.5 & 74.5 & 24.5 & 54.5 & 15.5 & 34.0 & 21.9 & 48.5 \\ \bottomrule
  \end{tabular}
  \caption{Performance comparison on THINGS-MEG using OpenCLIP ViT-g/14 as the vision encoder (latent dimension: 1024). }
  \label{tab:compare_ViT-g/14_MEG}
\end{table}

\begin{table}[h]
  \centering
  \small
  \setlength{\tabcolsep}{0.8pt} 
  \begin{tabular}{lcccccccccc}
    \toprule
    \multirow{2}{*}{Method}& \multicolumn{2}{c}{Subject 1} & \multicolumn{2}{c}{Subject 2} & \multicolumn{2}{c}{Subject 3} & \multicolumn{2}{c}{Subject 4}& \multicolumn{2}{c}{Avg} \\
    \cmidrule(r){2-3} \cmidrule(r){4-5} \cmidrule(r){6-7} \cmidrule(r){8-9} \cmidrule(r){10-11}
    & top-1 & top-5 & top-1 & top-5 & top-1 & top-5 & top-1 & top-5 & top-1 & top-5 \\

    \midrule
ShallowNet  & 8.5  & 29.0 & 28.5 & 55.5 & 19.0 & 54.5 & 10.5 & 30.0 & 16.6 & 42.3 \\
w/ ShrinkA  & 8.0  & 28.0 & 33.5 & 64.0 & 22.5 & 53.5 & 8.0  & 30.0 & 18.0 & 43.9 \\ \midrule
DeepNet     & 4.0  & 17.0 & 21.5 & 47.5 & 14.5 & 36.5 & 5.0  & 22.5 & 11.3 & 30.9 \\
w/ ShrinkA  & 1.5  & 8.5  & 17.0 & 43.5 & 9.0  & 32.0 & 3.0  & 12.0 & 7.6  & 24.0 \\ \midrule
EEGNet      & 7.5  & 24.5 & 28.5 & 59.5 & 22.0 & 53.5 & 6.0  & 26.5 & 16.0 & 41.0 \\
w/ ShrinkA  & 6.0  & 24.0 & 32.0 & 62.5 & 21.0 & 49.0 & 6.0  & 23.5 & 16.3 & 39.8 \\ \midrule
TSConv      & 12.0 & 38.0 & 42.5 & 71.5 & 30.0 & 66.5 & 18.0 & 40.0 & 25.6 & 54.0 \\
w/ ShrinkA  & 10.0 & 31.5 & 40.5 & 75.0 & 29.5 & 61.0 & 16.0 & 40.5 & 24.0 & 52.0 \\ \midrule
EEGProject  & 13.5 & 33.0 & 47.5 & 74.5 & 35.0 & 66.5 & 14.5 & 43.5 & 27.6 & 54.4 \\
w/ ShrinkA  & 14.5 & 30.0 & 43.0 & 75.0 & 29.5 & 63.0 & 12.5 & 38.0 & 24.9 & 51.5 \\ \midrule
STAE (Ours) & 17.5 & 34.5 & 49.0 & 76.5 & 36.0 & 67.0 & 15.5 & 43.0 & 29.5 & 55.3 \\
w/ ShrinkA  & 14.5 & 31.5 & 48.0 & 80.5 & 31.0 & 63.5 & 10.5 & 43.0 & 26.0 & 54.6 \\ \bottomrule
  \end{tabular}
  \caption{Performance comparison on THINGS-MEG using OpenCLIP ViT-bigG/14 as the vision encoder (latent dimension: 1280). }
  \label{tab:compare_ViT-bigG/14_MEG}
\end{table}

\begin{table}[h]
  \centering
  \small
  \setlength{\tabcolsep}{0.8pt} 
  \begin{tabular}{lcccccccccc}
    \toprule
    \multirow{2}{*}{Method}& \multicolumn{2}{c}{Subject 1} & \multicolumn{2}{c}{Subject 2} & \multicolumn{2}{c}{Subject 3} & \multicolumn{2}{c}{Subject 4}& \multicolumn{2}{c}{Avg} \\
    \cmidrule(r){2-3} \cmidrule(r){4-5} \cmidrule(r){6-7} \cmidrule(r){8-9} \cmidrule(r){10-11}
    & top-1 & top-5 & top-1 & top-5 & top-1 & top-5 & top-1 & top-5 & top-1 & top-5 \\

    \midrule
ShallowNet  & 7.0 & 23.0 & 17.5 & 44.0 & 11.5 & 30.5 & 6.5 & 24.0 & 10.6 & 30.4 \\
w/ ShrinkA  & 7.0 & 21.5 & 16.5 & 46.0 & 11.5 & 34.0 & 9.5 & 20.5 & 11.1 & 30.5 \\ \midrule
DeepNet     & 3.5 & 12.0 & 9.5  & 22.5 & 7.0  & 20.5 & 3.0 & 10.5 & 5.8  & 16.4 \\
w/ ShrinkA  & 1.0 & 5.5  & 6.0  & 21.5 & 5.5  & 17.0 & 1.0 & 6.5  & 3.4  & 12.6 \\ \midrule
EEGNet      & 3.0 & 10.5 & 12.0 & 33.0 & 6.5  & 20.5 & 3.0 & 11.0 & 6.1  & 18.8 \\
w/ ShrinkA  & 2.5 & 12.5 & 13.5 & 36.0 & 6.0  & 24.5 & 1.0 & 11.5 & 5.8  & 21.1 \\ \midrule
TSConv      & 7.5 & 23.0 & 23.0 & 52.0 & 17.5 & 38.0 & 9.0 & 29.0 & 14.3 & 35.5 \\
w/ ShrinkA  & 7.0 & 30.5 & 27.0 & 60.0 & 15.0 & 42.0 & 7.5 & 23.5 & 14.1 & 39.0 \\ \midrule
EEGProject  & 7.5 & 21.5 & 21.5 & 53.0 & 19.5 & 42.5 & 8.5 & 23.0 & 14.3 & 35.0 \\
w/ ShrinkA  & 6.0 & 22.5 & 27.0 & 59.5 & 16.5 & 38.0 & 7.5 & 22.0 & 14.3 & 35.5 \\ \midrule
STAE (Ours) & 8.5 & 21.0 & 26.0 & 52.5 & 20.5 & 42.0 & 9.0 & 24.0 & 16.0 & 34.9 \\
w/ ShrinkA  & 7.5 & 22.0 & 29.0 & 65.5 & 17.5 & 44.5 & 8.5 & 23.0 & 15.6 & 38.8 \\ \bottomrule
  \end{tabular}
  \caption{Performance comparison on THINGS-MEG using DINOv2 ViT-b/14 as the vision encoder (latent dimension: 768). }
  \label{tab:compare_dinov2-b14_MEG}
\end{table}

\begin{table}[htbp]
  \centering
  \small
  \setlength{\tabcolsep}{0.8pt} 
  \begin{tabular}{lcccccccccc}
    \toprule
    \multirow{2}{*}{Method}& \multicolumn{2}{c}{Subject 1} & \multicolumn{2}{c}{Subject 2} & \multicolumn{2}{c}{Subject 3} & \multicolumn{2}{c}{Subject 4}& \multicolumn{2}{c}{Avg} \\
    \cmidrule(r){2-3} \cmidrule(r){4-5} \cmidrule(r){6-7} \cmidrule(r){8-9} \cmidrule(r){10-11}
    & top-1 & top-5 & top-1 & top-5 & top-1 & top-5 & top-1 & top-5 & top-1 & top-5 \\

    \midrule
ShallowNet  & 4.0 & 17.0 & 9.5  & 35.5 & 5.5  & 22.0 & 5.0 & 14.5 & 6.0  & 22.3 \\
w/ ShrinkA  & 5.5 & 16.0 & 15.0 & 36.5 & 7.0  & 22.5 & 6.5 & 20.0 & 8.5  & 23.8 \\ \midrule
DeepNet     & 3.5 & 8.5  & 5.0  & 19.5 & 5.5  & 15.0 & 2.0 & 6.0  & 4.0  & 12.3 \\
w/ ShrinkA  & 1.0 & 3.5  & 7.5  & 21.0 & 2.5  & 12.5 & 1.0 & 4.5  & 3.0  & 10.4 \\ \midrule
EEGNet      & 2.0 & 9.0  & 9.0  & 26.5 & 6.0  & 16.0 & 2.5 & 6.5  & 4.9  & 14.5 \\
w/ ShrinkA  & 2.5 & 15.5 & 13.0 & 33.5 & 7.5  & 21.0 & 5.5 & 15.5 & 7.1  & 21.4 \\ \midrule
TSConv      & 6.0 & 19.0 & 18.0 & 46.0 & 10.0 & 27.5 & 5.5 & 18.5 & 9.9  & 27.8 \\
w/ ShrinkA  & 5.0 & 16.0 & 20.5 & 50.0 & 14.0 & 35.0 & 6.5 & 21.5 & 11.5 & 30.6 \\ \midrule
EEGProject  & 3.5 & 15.0 & 16.5 & 37.0 & 10.5 & 26.0 & 8.0 & 19.5 & 9.6  & 24.4 \\
w/ ShrinkA  & 7.0 & 16.0 & 16.5 & 42.5 & 9.0  & 29.5 & 8.0 & 21.0 & 10.1 & 27.3 \\ \midrule
STAE (Ours) & 6.5 & 17.0 & 15.0 & 39.0 & 7.0  & 27.0 & 6.5 & 20.0 & 8.8  & 25.8 \\
w/ ShrinkA  & 6.5 & 18.0 & 16.0 & 44.0 & 9.0  & 29.5 & 9.0 & 23.5 & 10.1 & 28.8 \\ \bottomrule
  \end{tabular}
  \caption{Performance comparison on THINGS-MEG using DINOv2 ViT-l/14 as the vision encoder (latent dimension: 1024). }
  \label{tab:compare_dinov2-l14_MEG}
\end{table}

\begin{table}[htbp]
  \centering
  \small
  \setlength{\tabcolsep}{0.8pt} 
  \begin{tabular}{lcccccccccc}
    \toprule
    \multirow{2}{*}{Method}& \multicolumn{2}{c}{Subject 1} & \multicolumn{2}{c}{Subject 2} & \multicolumn{2}{c}{Subject 3} & \multicolumn{2}{c}{Subject 4}& \multicolumn{2}{c}{Avg} \\
    \cmidrule(r){2-3} \cmidrule(r){4-5} \cmidrule(r){6-7} \cmidrule(r){8-9} \cmidrule(r){10-11}
    & top-1 & top-5 & top-1 & top-5 & top-1 & top-5 & top-1 & top-5 & top-1 & top-5 \\

    \midrule
ShallowNet  & 2.5 & 12.0 & 8.5  & 24.0 & 8.5  & 24.0 & 2.5 & 11.0 & 5.5  & 17.8 \\
w/ ShrinkA  & 5.5 & 19.0 & 10.5 & 34.0 & 8.5  & 22.5 & 2.0 & 10.5 & 6.6  & 21.5 \\ \midrule
DeepNet     & 2.5 & 7.0  & 6.0  & 21.0 & 4.0  & 13.5 & 0.5 & 6.0  & 3.3  & 11.9 \\
w/ ShrinkA  & 1.0 & 5.5  & 6.5  & 19.5 & 2.0  & 8.0  & 1.5 & 3.5  & 2.8  & 9.1  \\ \midrule
EEGNet      & 0.5 & 10.0 & 10.5 & 30.0 & 3.5  & 14.5 & 1.5 & 8.0  & 4.0  & 15.6 \\
w/ ShrinkA  & 4.5 & 9.0  & 10.0 & 31.5 & 6.5  & 22.0 & 4.0 & 14.5 & 6.3  & 19.3 \\ \midrule
TSConv      & 3.0 & 15.5 & 18.0 & 42.0 & 9.5  & 27.5 & 4.5 & 18.0 & 8.8  & 25.8 \\
w/ ShrinkA  & 5.5 & 21.5 & 14.5 & 40.0 & 11.5 & 32.5 & 3.5 & 14.0 & 8.8  & 27.0 \\ \midrule
EEGProject  & 4.5 & 16.0 & 10.5 & 36.5 & 9.5  & 29.0 & 7.0 & 17.0 & 7.9  & 24.6 \\
w/ ShrinkA  & 7.0 & 20.0 & 21.0 & 40.0 & 13.5 & 31.0 & 4.0 & 14.5 & 11.4 & 26.4 \\ \midrule
STAE (Ours) & 6.5 & 17.5 & 13.5 & 43.0 & 10.0 & 27.0 & 7.0 & 17.0 & 9.3  & 26.1 \\
w/ ShrinkA  & 8.0 & 24.0 & 21.0 & 42.5 & 15.5 & 32.5 & 6.5 & 15.5 & 12.8 & 28.6 \\ \bottomrule
  \end{tabular}
  \caption{Performance comparison on THINGS-MEG using DINOv2 ViT-g/14 as the vision encoder (latent dimension: 1536). }
  \label{tab:compare_dinov2-g14_MEG}
\end{table}

\end{document}